Università degli Studi di Milano-Bicocca
Dipartimento di Informatica, Sistemistica e Comunicazione
Ph.D. in Computer Science

# Towards Responsible AI in Banking: Addressing Bias for Fair Decision-Making

**Supervisor**: Prof. Fabio Mercorio
**Co-supervisor**: Prof. Mario Mezzanzanica
**Industrial Supervisor**: Dr. Andrea Cosentini
**Ph.D. Coordinator**: Prof. Leonardo Mariani

**Ph.D. Candidate:**
Dr. Alessandro Castelnovo, 736101
XXXVI Cycle

Academic Year 2022/2023

*"People often claim to hunger for truth, but seldom like the taste when it’s served up."*

Tyrion Lannister, A Clash of Kings

# Abstract

In an era characterized by the pervasive integration of artificial intelligence into decision-making processes across diverse industries, the demand for trust has never been more pronounced. This thesis embarks on a comprehensive exploration of bias and fairness, with a particular emphasis on their ramifications within the banking sector, where AI-driven decisions bear substantial societal consequences.

In this context, the seamless integration of fairness, explainability, and human oversight is of utmost importance, culminating in the establishment of what is commonly referred to as "Responsible AI". This emphasizes the critical nature of addressing biases within the development of a corporate culture that aligns seamlessly with both AI regulations and universal human rights standards, particularly in the realm of automated decision-making systems.

Nowadays, embedding ethical principles into the development, training, and deployment of AI models is crucial for compliance with forthcoming European regulations and for promoting societal good. This thesis is structured around three fundamental pillars: *understanding bias*, *mitigating* bias, and *accounting for* bias.

Within the realm of understanding bias, we introduce `Bias On Demand`, a model framework that enables the generation of synthetic data to illustrate various types of bias. Additionally, we analyze the intricate landscape of fairness metrics, shedding light on their interconnectedness and nuances.

Transitioning to the domain of mitigating bias, we present `BeFair`, a versatile toolkit designed to enable the practical implementation of fairness in real-world scenarios. Moreover, we developed `FFTree`, a transparent and flexible fairness-aware decision tree.

In the sphere of accounting for bias, we propose a structured roadmap for addressing fairness in the banking sector, underscoring the importance of interdisciplinary collaboration to holistically address contextual and societal implications. We also introduce `FairView`, a novel tool that supports users to select ethical frameworks when addressing fairness. Our investigation into the dynamic nature of fairness over time has culminated in the development of `FairX`, a strategy based on eXplainable AI adept at monitoring fairness trends.

These contributions are validated through their practical application in real-world scenarios, in collaboration with Intesa Sanpaolo. This collaborative effort not only contributes to our understanding of fairness but also provides practical tools for the responsible implementation of AI-based decision-making systems. In line with open-source principles, we have released `Bias On Demand` and `FairView` as accessible Python packages, further promoting progress in the field of AI fairness.

# Contents

# List of Figures

# List of Tables

# Acronyms

# 1
# Introduction

The global shift towards digitization, coupled with the growing availability of big data and significant advancements in Machine Learning (ML), and computational power, has prompted businesses worldwide to embrace Artificial Intelligence (AI) as a strategic goal [86]. Consequently, AI-based decision-making systems are progressively being employed across a wide range of industries, including the banking sector. Examples of applications encompass credit management, fraud detection, recruiting, and marketing automation [138, 129, 130]. The motivation behind adopting automated learning models is evident: we expect that ML algorithms will outperform humans in several tasks. ML algorithms have the ability to process vast amounts of data beyond human capacity, enabling them to consider multiple factors. Furthermore, ML algorithms excel at performing complex computations at significantly faster speeds. Additionally, while human decisions can be influenced by subjectivity, algorithms offer the potential for more objective decision-making [147].

Hence, individuals may unconsciously assume that the use of automated algorithms leads to more objective and fair decisions. However, this belief is often misguided, as ML algorithms can fall short of the desired objectivity. The misconception that ML algorithms are inherently bias-free stems from the flawed assumption that the training data used is unbiased. In reality, bias is deeply rooted in human society and, consequently, it is reflected in data [141]. For this reason, algorithms, like humans, are susceptible to biases that might lead to unfair outcomes [9].

Despite the numerous achievements and persistent enthusiasm surrounding AI, recent studies have revealed that AI can inadvertently cause harm to humans [86]. It is precisely the widespread implementation and extensive utilization of AI technologies that carry the immense and unforeseen potential for new types of threats that have yet to be fully understood or anticipated [87]. Considering the profound implications that crucial decisions, such as hiring or loan allocation, can have on people's lives, it is paramount to prioritize the evaluation and enhancement of the ethical dimensions that underpin the integration of automated systems within business decision-making processes [147]. The realization of AI systems' potential relies on their ethical and responsible utilization, creating a foundation of trust for both businesses and individuals.

Promoting trust in AI-based decision-making requires the integration of three key elements: fairness, eXplainable Artificial Intelligence (XAI), and human oversight, which must be combined harmoniously [196] (see figure 1.1). The adoption of AI by a company should be contingent upon widespread understanding, not only among data scientists and developers but also within governance and compliance structures. It is crucial to establish comprehensive guidelines and promote ethical values that align with both contingent AI regulations and human rights standards throughout the corporate culture. Any AI application must facilitate the identification and mitigation of biases in data or in the system outcome. Ultimately, the final decision-making authority should rest consciously with domain experts who are supported by AI. In this environment, fairness enables the measurement and mitigation of undesired biases, ensuring that AI systems exhibit desirable ethical characteristics. On the other hand, explanations for an AI system provide human-understandable interpretations of its inner workings and outcomes [52]. Both fairness and explanation are important components for building "Responsible AI" [196].

The importance of this direction is unequivocally demonstrated by the European Union's proactive efforts to regulate AI, aiming to create a more favorable environment for the development and deployment of this innovative technology. In April 2021, the European Commission introduced the First Proposal for a "Regulation laying down harmonised rules on artificial intelligence" (AI Act) [169]. On December 8, 2023, the European Parliament and Council reached a provisional agreement on the AI Act. The agreed text will now have to be formally adopted by both Parliament and Council to become EU law[1]. An important aspect of the AI Act's provisions targets high-risk AI systems involved in credit assessments and hiring practices, significantly impacting banks by imposing stringent requirements for ensuring fairness, and accountability, particularly in guarding against potential undesired bias. These requirements aren't meant to limit AI adoption in these processes; instead, they serve as crucial guidelines for deploying AI-driven solutions. Rooted in EU values and fundamental rights, the AI Act aims to cultivate trust in AI-driven solutions while encouraging responsible progress by businesses.

[1]https://www.europarl.europa.eu/news/en/press-room/20231206IPR15699/artificial-intelligence-act-deal-on-comprehensive-rules-for-trustworthy-ai.

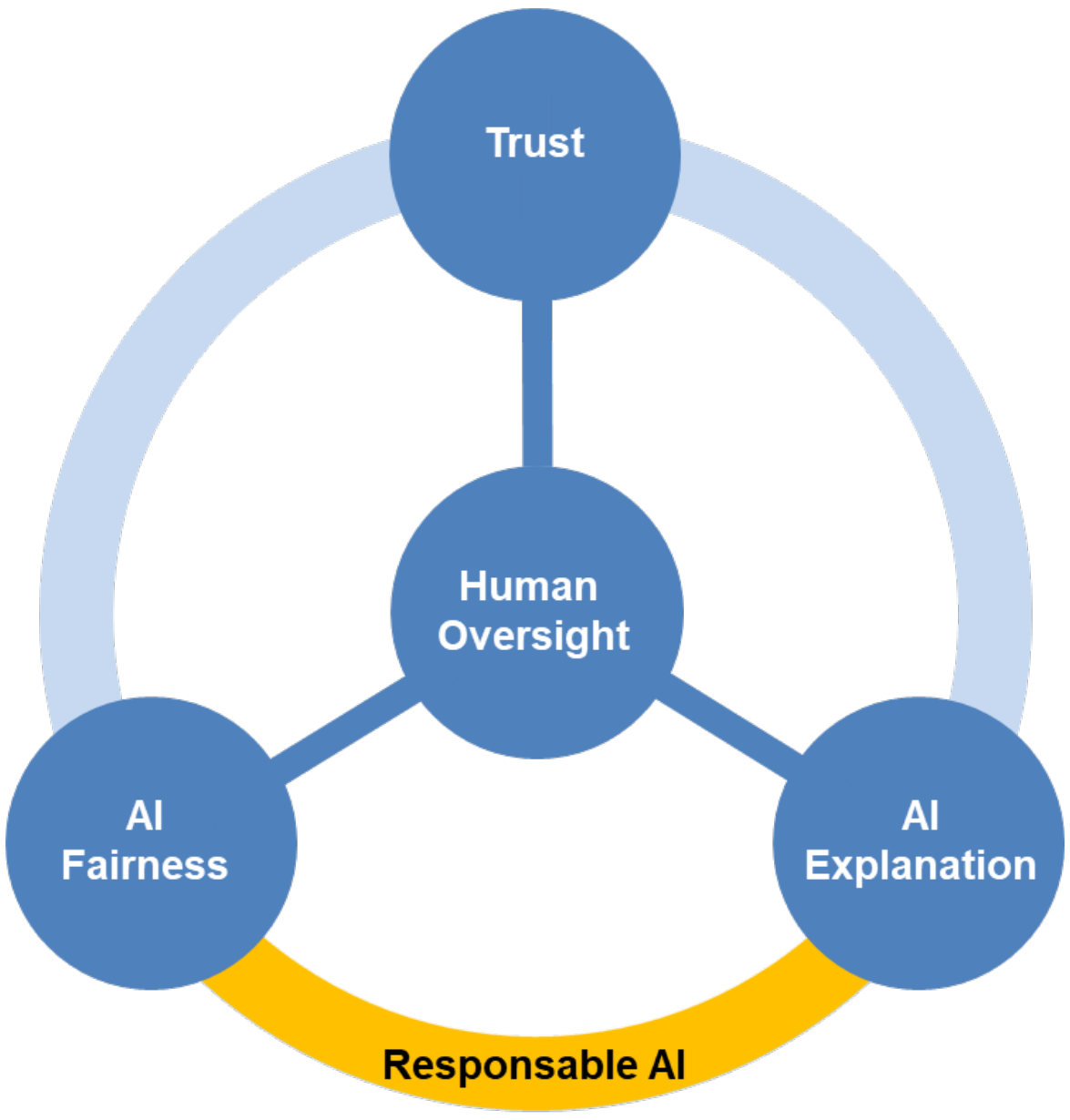

Figure 1.1 Relations among AI fairness, AI explanation, human oversight, and trust, as discussed by Zhou et al. [196].

This thesis contributes to the literature in the field of fairness by addressing bias within the banking sector. According with Ntoutsi et al. [141], the work is divided into three broad categories:

- *Understanding bias*. These approaches aim to deepen our understanding of how bias is generated, how it manifests in the data, and how it impacts the outcomes of AI systems.
- *Mitigating bias*. These approaches focus on addressing bias at different stages of AI decision-making, such as pre-processing, in-processing, and post-processing. They aim to mitigate bias by carefully handling data inputs, optimizing learning algorithms, and refining model outputs.
- *Accounting for bias*. These approaches proactively account for bias by incorporating bias-aware decision-making mechanisms. They also prioritize human involvement, allowing for human intervention and oversight, while ensuring that understandable explanations of AI outcomes are provided.

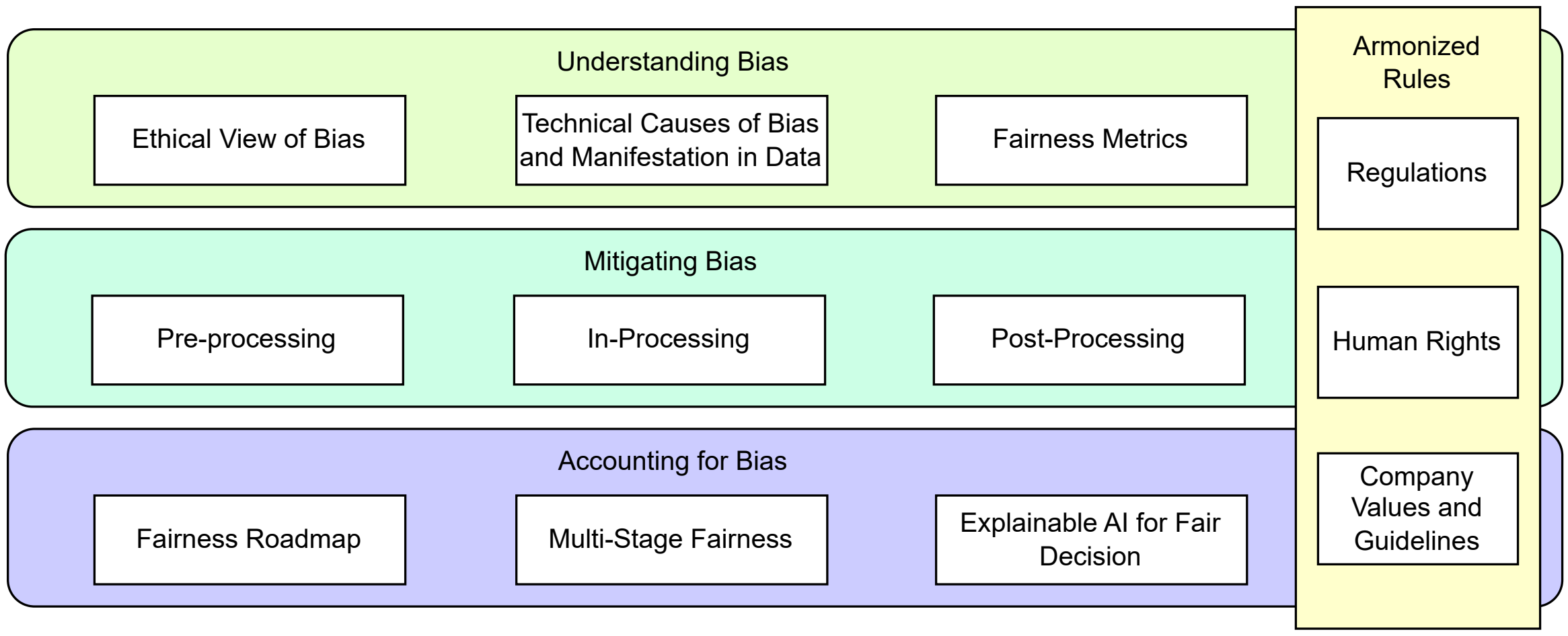

Figure 1.2 Overview of topics discussed in this thesis. The categories are inspired by the survey on bias conducted by Ntoutsi et al. [141].

By providing real-world applications in Intesa Sanpaolo, this thesis makes contributions to all these three categories. Figure 1.2 visually represents the comprehensive coverage of these topics within the thesis.

## 1.1 Contributions

This thesis makes contributions to *(1) understanding, (2) mitigating*, and *(3) accounting for bias*. The aim is to provide tools, frameworks, and real-world applications to foster trust and promote the adoption of responsible AI-based decision-making systems in the banking industry. In particular:

**1** in the field of *understanding Bias*:

1a) we introduce `Bias On Demand` [15], a model framework for generating synthetic data with specific types of bias. The formalisation of these various types of bias is based on the theoretical classifications present in the relevant literature, such as the surveys on bias in ML by Mehrabi et al. [128], Ntoutsi et al. [141], and Suresh and Guttag [167]. We provide an explicit mathematical representation of the fundamental types of bias and link it with the stream of literature that investigates their relation with moral worldviews;

1b) we provide a **clarification in the fairness metrics landscape** [36]. More in detail, we contribute to the understanding of fairness metrics and their interplay, specifically exploring the relationship between individual and group fairness as well as observational and causality-based criteria. We highlight the complexity and intertwined nature of fairness notions. Our analysis suggests that fairness notions and metrics are interconnected, and the amount of sensitive information accepted in non-sensitive features influences the trade-off between fairness metrics.

**2** In the field of *Mitigating Bias*:

2a) we present `BeFair` [35], a toolkit that offers a range of functionalities designed to implement fairness in real-world use cases. Its capabilities encompass bias detection within input datasets and model outcomes, bias mitigation utilizing various strategies from existing literature, performance evaluation using pertinent metrics, performance/fairness trade-off comparison across different strategies, and interpretation of feature relationships through causal graphs;

2b) we present `FFTree` [34], a transparent, flexible, and fairness-aware decision tree. `FFTree` enhances the classical approach introduced in [29] with a method to find "fair" splits designed to work with different fairness criteria. This method aims to support the user in the creation of responsible decision systems through the following peculiarity:

- *flexibility*. Depending on the application domain, the user can select the proper fairness criteria from a large set of possible fairness metrics. Moreover, it permits the implementation of multiple fairness definitions or mitigates on multiple sensitive groups simultaneously. The user can also set the required level of fairness as an input parameter to meet different business needs or regulatory requirements.
- *transparency*. Since it is based on a decision tree, `FFTree` is transparent by design. Furthermore, the constraints introduced in the system are directly derived from the chosen fairness criteria and explicitly added to the model.

**3** In the field of *Accounting for Bias*:

3a) we propose a **roadmap for addressing fairness in the banking sector**. In particular, we provide a general guideline to address fairness in ML projects and focus on the fact that different expertise should work together along the process in order to properly take into account the context and the social impact of the technological service/product being developed;

3b) we introduce `FairView` [38], a tool that aims to support the choice from the user of which ethical framework to assume when addressing fairness in the development of a responsible AI system. This approach is based on evaluative AI, which generates evidence that supports human judgments by explaining trade-offs between any set of possibilities. `FairView` represents one of the first attempts to link XAI with moral frameworks with which the ethical-philosophical literature formalises the matter of fairness;

3c) we investigate the **effects of fairness through time** [40]. In particular, this work:

- highlights that classical group fairness mitigation strategies can fail when used over time;
- investigates a strategy to retrain the group fairness mitigation model assuming the absence of the final human decision;
- proposes `FairX`, a XAI strategy based on a SHAP [121] extension to monitor fairness over time.

We provide applications of `FFTree`, `BeFair`, and `FairX` using real data on credit lending owned by Intesa Sanpaolo. Moreover, we released `Bias On Demand` and `FairView` as **open source** Python packages.

## 1.2 Thesis Outline

The thesis is divided into four main parts, organized as follows:

**Part I** focuses on to the *understanding of bias*. Chapter 2 delves into the fundamental types of bias, exploring their ethical considerations and how they can influence and emerge throughout the machine learning lifecycle. In Chapter 3, a concise mathematical representation of prevalent biases in machine learning is provided, alongside a framework for generating synthetic data that emulates these biases (*contribution 1a*). In Chapter 4, fairness metrics are presented and their interconnections are comprehensively discussed (*contribution 1b*).

**Part II** describes strategies for *mitigating bias*. In Chapter 5, we explore prevalent fairness mitigation techniques and their implementation in both real-world and synthetic datasets (*contribution 2a*). Chapter 6 presents `FFTree`, our transparent and flexible mitigation approach, and compares it with other related state-of-the-art methods (*contribution 2b*).

**Part III** presents our approaches for proactively *account for bias*. In Chapter 7, we present our roadmap to address fairness in banking ML projects (*contribution 3a*). In Chapter 8, we introduce `FairView`, a tool that leverages evaluative AI and contrastive explanation to aid decisions about fairness metrics to be enforced in a given use case (*contribution 3b*). Chapter 9 shifts the focus to the long-term effects of fairness, proposing strategies to ensure fairness over time and monitor it through explainable AI (*contribution 3c*).

**Part IV** concludes.

## 1.3 Prior Work

All research in this thesis either has already been published as conference or journal papers.

*Contribution 1a* resulted from joint work with Joachim Baumann, Riccardo Crupi, Nicole Inverardi and Daniele Regoli titled "Bias on Demand: A Modelling Framework That Generates Synthetic Data With Bias", published at *Proceedings of the 2023 ACM Conference on Fairness, Accountability, and Transparency. 2023*.

*Contribution 1b* resulted from joint work with Riccardo Crupi, Greta Greco, Daniele Regoli, Ilaria Penco and Andrea Cosentini titled "A clarification of the nuances in the fairness metrics landscape", published at *Scientific Reports*.

*Contributions 2a and 3a* resulted from joint work with Riccardo Crupi, Giulia Del Gamba, Greta Greco, Aisha Naseer, Daniele Regoli and Beatriz San Miguel Gonzalez titled "Be-Fair: Addressing Fairness in the Banking Sector", published at *2020 IEEE International Conference on Big Data*.

*Contribution 2b* resulted from joint work with Lorenzo Malandri, Fabio Mercorio, Mario Mezzanzanica and Andrea Cosentini titled "FFTree: A flexible tree to handle multiple fairness criteria", published at *Information Processing and Management*.

*Contribution 3b* resulted from joint work with Nicole Inverardi, Lorenzo Malandri, Fabio Mercorio, Mario Mezzanzanica and Andrea Seveso titled "Leveraging Group Contrastive Explanations for Handling Fairness", published at Conference proceedings *Explainable Artificial Intelligence First World Conference, xAI 2023*.

*Contribution 3c* resulted from joint work with Andrea Cosentini, Lorenzo Malandri, Fabio Mercorio and Mario Mezzanzanica titled "Towards Fairness Through Time", published at *ECML PKDD 2021: Machine Learning and Principles and Practice of Knowledge Discovery in Databases*.

# Part I

# Understanding Bias

# 2

# Bias and Moral Framework in AI-based Decision Making

## 2.1 Ethical View of Bias

There is little consensus in the literature regarding bias classification and taxonomy. Indeed, the very notion of bias depends on deep philosophical considerations, and ethical issues are rarely resolved in a definitive and univocal way. Different understandings of bias and fairness depend on the assumption of a belief system beforehand. Friedler et al. [71] and Hertweck et al. [84] talk about *worldviews*. In particular, Friedler et al. [71] outline two extreme cases, referred to as What You See Is What You Get (WYSIWYG) and We Are All Equal (WAE). The first one consists in taking as "good" the situation as it is, while the second assumes that different groups of people identified by some sensitive characteristic are equal *at some level*, and thus they *should* be treated equally, given that level. Notice that, potentially, the WAE worldview embodies a very rich set of possibilities, depending on the level at which equality is assumed.

More in detail, starting from the definition of three different metric spaces, WYSIWYG and WAE differ because of the way they consider the relations in between. The first space is the Construct Space (CS) and represents all the unobservable realised characteristics of an individual, such as intelligence, skills, determination, or commitment. The second space is the Observable Space (OS) and contains all the measurable properties that aim to quantify the unobservable features, think e.g. of IQ or aptitude tests. The last space is the Decision Space (DS), representing the set of choices made by the algorithm on the basis of the measurements available in OS. Note that shades of ambiguity are already detectable at this level, because the mappings between spaces are susceptible to distortions. Moreover, CS is by definition unobservable, thus we can only make *assumptions* on it.

According to WYSIWYG, CS, and OS are essentially equal, and any distortion between the two is altogether irrelevant to the fairness of the decision resulting in DS. Contrarily, WAE does not make assumptions about the similarity of OS and CS, and moreover, assumes that we are all equal in CS, i.e. that any difference between CS and OS is due to a biased observation process that results in an unfair mapping between CS and OS. With the distance between worldviews in mind, the notion of fairness inspired by [63] affirming that individuals that are close in CS shall also be close in DS (commonly known as individual fairness) appears diversified and differently achievable. If WYSIWYG is assumed, non-discrimination is guaranteed as soon as the mapping between OS and DS is fair, since CS $\approx$ OS. On the other hand, according to WAE the mapping between CS and OS is distorted by some bias whenever a difference among individuals emerges (this difference is named *measurement bias* in [84]); therefore, to obtain a fair mapping between CS and DS those biases should be mitigated properly. Note that the differences visible in OS could be a “true” measure of what the algorithm aims to evaluate, for example, a CV may be a trustworthy translation of personal talents. Nonetheless, according to WAE, a bias could still be present, since the worldview promotes the perspective of considering all individuals equal *in principle*.

Building on [71], Hertweck et al. [84] describe a more nuanced scenario by introducing the notion of Potential Space (PS): individuals belonging to different groups may indeed have different realised talents (i.e. they actually differ in CS), and these may be accurately measured by resumes (i.e. CS $\approx$ OS), but, if we assume that these groups have the same *potential* talents (i.e. they are equal in PS), then the realised difference must be due to some form of unfair treatment of one group, that is referred to as *life bias*. Hertweck et al. [84] call this view We Are All Equal in Potential Space (WAEPS). Actually, as argued in [84], we can effectively think of the WAE assumption as a family of assumptions, depending on the point in time in which the equality is assumed to hold: the more we go back in time in presuming equality between individuals, the more the consideration of life circumstances becomes strong, and thus the less discrimination between individuals is considered legitimate.

These extreme worldviews amount on the one hand to accept the situation as it is observed (WYSIWIG), on the other to infer some form of unfairness whenever there is some observed disparity (WAE). To avoid such extreme scenarios, philosophical theories around Equality of Opportunity (EO) offer some suggestions and interpretive tools for approaching biases in different situations [67, 158, 152]. In this sense, western political philosophy and algorithmic fairness literature encounter themselves in the formulation of fairness around the concept of equal opportunities for all members of society. Heidari et al. [83] list three different EO conceptions, going from more permissive to more stringent: *Libertarian EO*, by which individuals are held accountable for any characterizing feature, sensitive one included; *Formal EO*, by which individuals are not held accountable for differences in sensitive features only; *Substantive EO*, by which there is a set of individual characteristics which are due to *circumstances* and others that are a consequence of individual *effort*, and people should be held accountable only on the basis of the latter. The choice of which characteristics fall in the level of circumstances and which can be considered as individual effort is far from obvious. Depending on the EO framework that one is willing to embrace, observed disparities may be seen as “just” or “unjust” forms of bias.

In the following sections, we shall describe the most common biases, explaining how they relate to these fundamental concepts.

## 2.2 Bias Throughout the ML Life Cycle

In the previous section, we discussed bias from an ethical and philosophical perspective. With a different perspective, Suresh and Guttag [167] argue that bias can also be seen as a source of harm that arises during different stages in the ML life cycle. Indeed, the entire ML life cycle, from data collection to model deployment, involves a series of decisions and actions that can lead to unintended consequences. It is challenging for an ML practitioner working on a novel system to identify whether and how problems may occur. Even if detected, it is difficult to establish the proper mitigation method for dealing with biases. A first step in this direction is to understand the different types of bias, their sources and consequences. Figures 2.1 and 2.2 exemplify the representation of the fundamental biases from a philosophical (Figure 2.1) and technical (Figure 2.2) point of view. The main difference between these two representations does not lie in the notations but in what they describe: the philosophical vision depicts individuals, whereas the technical one describes the population. With the individual standpoint in mind, it is possible to comprehend that philosophically reasoning the historical bias is responsible for the transition between potential and construct space, whereas the measurement bias renders reason of the distortion between construct and observed space. On the other hand, the technical approach by [167] does not define the potential space, but rather begins by representing the world with the historical bias entrenched in it. To pursue analogies, the world can be considered the equivalent of the construct space, while the dataset coincides with the observed space. Since the technical perspective focuses on all phases of the ML system's life cycle, the description of bias is much more granular: between the world and the dataset plays a significant role the sample used to collect the data, which may be affected by representation bias. Furthermore, direct discrimination, which in philosophy refers to the transition between OS and DS, can be introduced across a ML pipeline by a variety of biases arising from algorithm development (algorithmic bias) and production implementation (deployment bias).

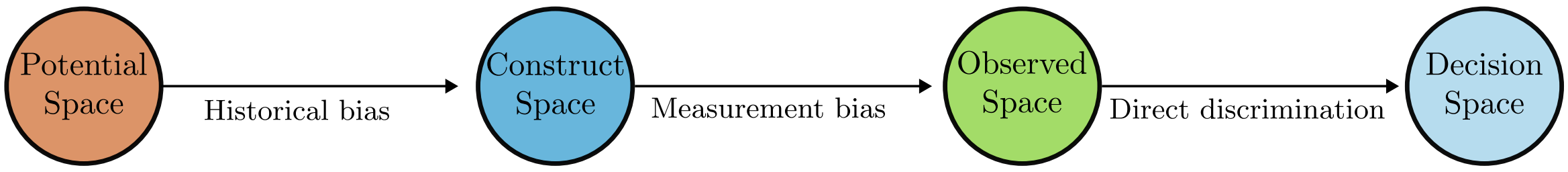

Figure 2.1 Schematic representation of biases in terms of abstract spaces, as introduced in [71] and extended in [84].

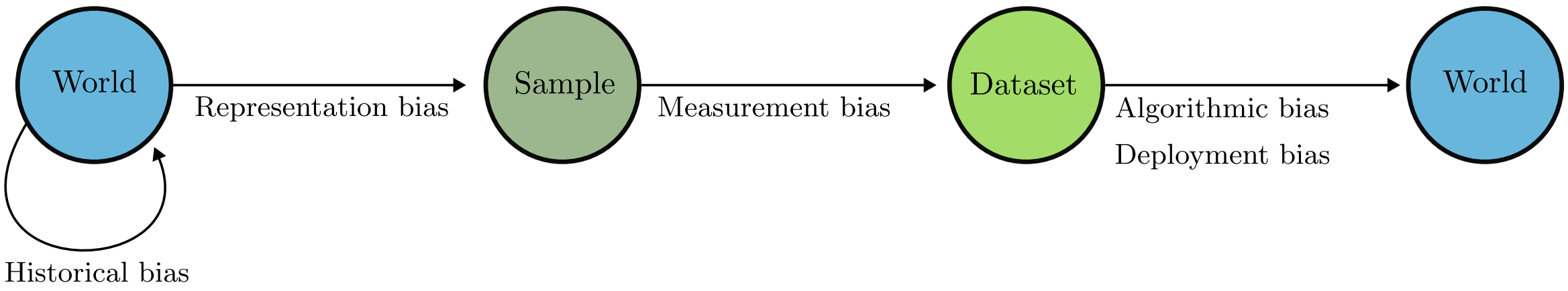

Figure 2.2 Schematic representation of biases in the ML modelling pipeline, as introduced in [167].

It is important to distinguish between biases that arise during the data collection (affecting the data generation) and biases that arise during the development and deployment of the model (affecting the system's outcome). Namely, because in real cases, the former typically depend on context and are inherent in the data without the user being able to eliminate them during data collection, whereas the latter depend on user's decisions in handling the data. The proper mitigation strategy depends on the comprehension of the biases that affect the data generation and should be determined through both technical and philosophical considerations. In the following section, we will discuss in more detail the most common biases.

## 2.3 Fundamental Types of Bias

In what follows we focus on what we consider the core building blocks of most types of bias. Following [128], we divide the fundamental types on bias in *from user to data*, *from data to algorithm*, *from algorithm to user* and *from user to world*.

### 2.3.1 Bias From User to Data

Biases going *from user to data* impact the phenomenon to be studied and thus the dataset [128].

**Historical bias** — sometimes referred to as *social bias*, *life bias*, or *structural bias* [128, 141, 84]— occurs whenever a variable of the dataset relevant to some specific goal or task is dependent on some sensitive characteristic of individuals, *but in principle it should not*. An example is the different average income among men and women, due to long-lasting social barriers and not reflecting intrinsic differences among genders.

A similar situation may arise when a dependence on sensitive individual characteristics is present with respect to the variable that we are trying to estimate or predict. These are the cases in which the target of model estimation is itself prone to some form of bias, e.g. because it is the outcome of some human decision. Consider e.g. trying to build a data-driven decision process that determines whether to grant a loan on the basis of past loan officers' decisions rather than actual repayments.

Note that the actual presence of historical bias is conditioned by the previous assumption of the WAEPS worldview. Indeed, arguing that, in principle, there should be no dependence on some sensitive features only makes sense if a moral belief of substantial equity is required in the first place. Otherwise, according to WYSIWYG, CS is fairly reported in OS, and therefore structural differences between individuals are legitimate sources of inequality.

### 2.3.2 Bias Form Data to Algorithm

Biases going *from data to algorithm* impact the dataset but not the phenomenon itself [128].

**Measurement bias** — occurs when a proxy of some variable relevant to a specific goal or target is employed, and that proxy happens to be dependent on some sensitive characteristics. For instance, one may argue that IQ is not a "fair" approximation of actual "intelligence", and it might systematically favour/disfavour specific groups of individuals. Statistically speaking, this type of bias is not very different from historical bias —since it results in employing a variable correlated with sensitive attributes— but the underlying mechanism is nevertheless different, and in this case the bias needs not to be present in the phenomenon itself, but rather it may be a consequence of the means chosen to translate unobservable properties into OS. This is an example of bias *from data to algorithm* in the taxonomy of [128], i.e. a bias due to data availability, choice and collection. Note, incidentally, that this form of bias might as well occur with the target variable (i.e. the label). In this situation, it is the quantity that we are trying to estimate/predict that is somehow "flawed". Further, notice that the WYSIWYG worldview assumes that CS $\approx$ OS, i.e. that there is no measurement bias. On the other hand, the WAE worldview assumes equality among groups only in the CS, which allows for measurement bias (i.e. $CS \neq OS$).

Despite assuming that there is no measurement bias in a WYSIWYG worldview (CS $\approx$ OS), there is still the risk that the construct in question is ill-defined. In this case, on would have a correct measurement, but, in fact, to measure 'the wrong thing'. Alternatively, according to a WAE worldview, even if one measures 'the right thing', is possible that is is measured wrongly, i.e. resulting in measurement bias ($CS \neq OS$).

The fact that measurement bias depends also on a choice component, which is to say the choice of the dataset, can extend its occurrence in the WYSIWYG worldview as well. Indeed, the choice of "what to measure" and "how to measure" determines and influences what is made observable. The eventuality of awareness of biased measurements would probably require mitigation also in WYSIWYG. Alternatively, according to WAE, measurement bias may lie in the mapping between CS and OS, which means between a "real" ability of an individual and an observable quantity that aims to measure it.

**Representation bias** — occurs when, for some reason, data are not representative of the world population. For example, one subgroup of individuals, identified by a sensitive characteristic such as ethnicity, age, etc., may be heavily underrepresented. This may occur in different ways. It may be at random, i.e. the subgroup is less numerous than it should be, but without any particular skewness in the other characteristics: in this scenario, this single mechanism is not sufficient to create disparities, but it may exacerbate existing ones. Alternatively, the under-represented subgroup might contain individuals with disproportionate characteristics with respect to their corresponding world population, e.g. only low-income individuals or only low-education individuals. In the latter case, representation bias may be sufficient to create inequalities in decision-making processes based on that data.

The mechanism underlying the representation bias should be analysed on the basis of the assumed worldview: e.g. if the data has an under-represented ethnic minority one should investigate *why* it is so. If the target population is itself different from world population (i.e it is not merely a poor data collection), then one should consider the reasons by which this ethnic minority is under-represented in the target population, e.g. in the Substantive EO framework one should understand if these reasons have to be regarded as *circumstance* or as consequence of individual *effort*. In the latter case, the representation disparities are not to be considered as "unfair" *per se*.

Representation bias is strictly connected to *sampling bias*, in that it embodies problems arising during data collection, e.g. by collecting disproportionately less observations from one subgroup, possibly skewed with respect to some characteristics. Like *measurement bias*, this is a form of bias going *from data to algorithm*.

**Omitted variable bias** —may occur when the collected dataset omits a variable relevant to some specific goal or task. If the variables that are present in the dataset have some dependence on sensitive characteristics of individuals, an ML model trained on such a dataset will learn those dependencies, thus producing outcomes with spurious dependence on sensitive attributes. Notice that the omission of a relevant variable alone cannot, in general, be a source of disparities and bias in the data, but it can amplify and exacerbate other biases already present (e.g. historical biases). Assuming the Formal EO framework, sensitive features are omitted by default. While this may appear fairer because the decision is made solely on the basis of the relevant attributes, on the other hand there may be cases in which it becomes arduous to mitigate on structural biases that affect achievements. Depending on worldview assumptions or on the chosen EO framework, the mechanism through which the residual variables happen to depend on sensitive individual characteristics should be as well analysed to understand/decide whether this dependence is legitimate or if they are themselves a consequence of some bias at work.

In terms of consequences on the data, it may well be that different types of bias result in very similar effects. For example, representation bias may create in the dataset spurious correlations among sensitive characteristics of individuals and other characteristics relevant to the problem at hand, a situation very similar to the correlations present as a consequence of historical bias. This reminds us that, in general, we are not aware of the type of bias (or biases) affecting the data and that their interpretation depends on former assumptions. Furthermore, starting from specific worldview and framework influence the choices of optimization.

The above list of biases should be seen as the set of the most important mechanisms through which unfairness can be introduced to ML-based decision-making systems due to the used dataset. However, biases can also occur during the development of the ML algorithm (*algorithm to user* biases) or when the system is deployed (*user to world* biases).

### 2.3.3 Bias From Algorithm to User

Biases going *from algorithm to user* impact the resulting predictor, which is then used to inform decisions [44].

**Algorithmic bias** — may occur whenever the algorithmic outcomes affect the behaviour of users. i.e. the bias is generated purely by the algorithm using unbiased data. There are different specific aspects of the ML pipeline that can result in *algorithmic bias* [54]: *Aggregation bias*[1] arises when just one ML model is used for everyone even though there are subgroups for which a different model would be better suited due to heterogeneity w.r.t. the mapping of the features to the labels, i.e. the probability of having a label given some features. *Learning bias* occurs when the algorithmic design choices (such as the specified learning objective function or regularisation techniques) are not equally suited for all subgroups. *Evaluation bias* bias occurs if the benchmark dataset or the metrics used to assess the performance do not appropriately capture the relevant target population for which the system is ultimately used. In line with [128], we combine *aggregation bias*, *learning bias*, and *evaluation bias* using the umbrella term *algorithmic bias*: what these three variations of *algorithmic bias* have in common is that they all exacerbate performance disparities on underrepresented groups. Thus, they are going *from algorithm to user* in that they affect the learned predictor in a way that it results in unintended harmful consequences despite using an unbiased dataset.

### 2.3.4 Bias From User to World

Biases going *from user to world* impact the way final decisions are made and may thereby end up causing harmful consequences.

**Deployment bias** — arises if the process followed to take decisions based on the algorithm's prediction results in harmful downstream consequences. *Deployment bias* often occurs when predictions are used to inform human decision-makers, whereas the system has been created as if the decision would be taken fully automated based on the algorithmic predictions [167]. It is difficult to model all possible ways in which human decision-makers may act on the predictions, as this can be the result of complex processes in reality. Extending Mehrabi et al. [128]'s classification, *deployment bias* can be seen as a type of bias that is going *from user to world*, as it affects the way users of algorithmic decision-making systems derive (potentially consequential) decisions on individuals in the world.

[1] Notice that Mehrabi et al. [128] describe *aggregation bias* as a type of bias that goes *from data to algorithm*, whereas Suresh and Guttag [167] count it towards the biases arising during the model building stage as it represents a limitation on the learned predictor. Here, we follow Suresh and Guttag [167]'s interpretation.

# 3

# Bias On Demand: A Framework for Generating Synthetic Data with Bias

For fostering the understanding of bias we introduced Bias On Demand [15], a model framework for generating synthetic data with specific types of bias. We provide an explicit mathematical representation of the various types of bias based on the theoretical classifications discussed in the previous chapter.

The benefits of this strategy include the possibility of examining circumstances not available with real-world data but that may occur, and – even when real-world data is available – to precisely control and understand the data generation mechanism. Moreover, it is indisputable that making data and related challenges accessible to the research community for analysis could contribute to sound policy decisions that benefit society [151].

With Bias On Demand, we aim to draw attention to the issue of bias in AI systems and its potential impact on fundamental rights and legal compliance. The objective of this chapter is to raise awareness and promote the development of responsible AI systems, *free of bias*.

## 3.1 Related Works on Synthetic Data Generation

Synthetic data generation is a relevant practice for both businesses and the scientific community. As a result, the literature has given it a lot of attention. Two main directions in the research on synthetic data are the *emulation* of certain key information in the real dataset while preserving privacy [151, 10], and the *generation* of different testing scenarios for evaluating phenomena not covered by available data [113]. Assefa et al. [10] presented basic use cases with specific examples in the financial domain like: internal data use restrictions, data sharing, tackling class imbalance, lack of historical data, and training ML models. According to Assefa et al. [10], synthetic representations should possess several desirable properties, including *human readability*, *compactness*, and *privacy preservation*. Notice that synthetic data generation may also be a valid alternative to data anonymisation as a means of preserving privacy in data to be published or shared [125]. Indeed, synthetic data are typically newly generated data (thus different, by design, from real observations), subject to constraints to protect sensitive personal information while still allowing valid inferences [151].

Synthetic data are generally classified into: *fully synthetic data*, *partially synthetic data*, and *hybrid synthetic data* (see [5] for further details). Our model framework belongs to the first of these cases.

Synthetic data generation can be approached in several ways, depending mainly on the objective —see [64] for a detailed overview of the techniques for generating synthetic data. When there is (enough) real data available and the main goal is to emulate the "structure" of that data, synthetic samples can be drawn from a probability distribution learned from the real data. This is achieved through distribution fitting approaches, such as Gaussian Mixture Models or Hidden Markov Models, as well as modern Deep Learning-based approaches, ranging from Autoencoders to Generative Adversarial Networks, Diffusion models, and Language Models, which are collectively referred to as Generative AI. If the objective is to create benchmark scenarios that comply with specific properties, a possible strategy is to simulate instances using a set of (stochastic) equations that represent the desired relationships among variables. This approach is aligned with the method we propose in the following. We refer to [64] for a detailed overview of the techniques for generating synthetic data.

Researchers in the field of algorithmic fairness acknowledge the difficulty in finding suitable datasets for their experiments, relying heavily on a handful of benchmark datasets, e.g. for studying and developing bias mitigation strategies [59]. To overcome this limitation, it is not uncommon to use synthetic datasets to demonstrate specific properties of a novel discrimination-aware method, as highlighted in algorithmic fairness dataset surveys, such as [113, 65]. They show that some works, such as [50], use well-known benchmark synthetic datasets to validate fair representation learning, whereas other studies, such as [54, 119, 154, 186, 36, 16], generate *ad hoc* toy datasets for their testing scenarios. Reddy et al. [154], e.g. evaluate different fairness methods trained with deep neural networks on synthetic dataset: different imbalances and correlations are embedded in the data to verify the limits of the current models and better understand under which setups they are subject to failure.

As introduced in the previous section, our aim is to introduce a model framework for generating synthetic data that replicates common biases, allowing us and the research community to investigate fairness-related issues that arise when bias is introduced at different points in the ML pipeline during the development of ML-based decision-making systems.

## 3.2 How `Bias On Demand` Generates Synthetic Dataset

We propose a simple modelling framework able to simulate the bias-generating mechanisms described in Section 2.3.

The rationale behind the model is that of being at the same time sufficiently flexible to accommodate all the main forms of bias *while* maintaining a structure as simple and intuitive as possible to facilitate *human readability* and ensure *compactness* avoiding unnecessary complexities that might hide the relevant patterns.

As noted in Section 2.3, following [128], we can distinguish between *from user to data* (impacting the phenomenon to be studied and thus the dataset), *from data to algorithm* (impacting directly the dataset but not the phenomenon itself), *From algorithm to user* (impacting the predictor) and *from user to world* biases (impacting the final decisions).

Formally, we model the relevant quantities describing a phenomenon as random variables, in particular, we label $Y$ the *target* variable, namely the quantity to be estimated or predicted on the basis of other *feature* variables, that we collectively call $X$. We assume that the underlying phenomenon is described by the formula

$$Y = f(X) + \varepsilon, \tag{3.1}$$

where $f$ represents the actual relationship between features and target variables, modulated by some idiosyncratic noise $\varepsilon$. Oftentimes, what we observe in the OS is not equivalent to the construct we would like to grasp (in the CS). Formally, this refers to how features and labels are generated and collected:

$$\widetilde{X} = g(X), \quad \widetilde{Y} = h(Y); \tag{3.2}$$

where $g$ and $h$ represent the collection and measurement of relevant individual attributes and outcomes. The use of $(\widetilde{X}, \widetilde{Y})$ rather than $(X, Y)$ describes the fact that the set of variables and outcomes employed to make inferences about a phenomenon may not coincide with the actual variables that play a role in that phenomenon. This is precisely what happens in some forms of biases, such as measurement bias or omitted variable bias, but also in case of representation issues.

A data-driven decision-maker infers from a (training) set of samples $\{(\widetilde{x}_i, \widetilde{y}_i)\}_{i=1}^{N}$, an estimate for $f$ that we label $\hat{f}$, thus producing its best estimate for $Y$, namely

$$\hat{Y} = \hat{f}(\widetilde{X}). \tag{3.3}$$

The prediction $\hat{Y}$ is then used to inform a final decision $D$. Thereby, the decision rule, which we denote by $r$, specifies how a decision is taken based on the individual prediction.

$$D = r(\hat{Y}). \tag{3.4}$$

In its simplest, fully automated, form without any fairness constraints, optimal decision rules $r^*$ usually take the form of a uniform threshold, i.e. all individuals with a prediction that lies above a certain value $\tau$ (i.e. $\hat{Y} > \tau$) are assigned a positive decision $D = 1$, all others are assigned a negative decision $D = 0$. However, in many real-world scenarios, decisions are not fully automated but are taken by human decision-makers who take a decision (potentially) based on the predicted outcome. In this case, decisions are not necessarily just based on $\hat{Y}$, i.e. it can depend on any other environmental information $Z$ ($r : \hat{Y}, A, Z \to D$). If the decision rule $r$ applied by (human or machine) decision-makers introduces unexpected behaviour resulting in disparities between the decision received by individuals from different groups *deployment bias* can arise [167].

Notice that *user to data* types of bias impact directly Equation (3.1), *data to algorithm* biases affect the data observation process described in Equation (3.2), *algorithm to user* biases (i.e. *algorithmic bias*) occur at the level of Equation (3.3), and *user to world* biases (i.e. *deployment bias*) is linked to the decision rule formalised in Equation (3.4).

Our framework is very much in line with the discussions outlined by Suresh and Guttag [167]. In particular, we refer to Figure 2 in [167] and the corresponding discussion. Incidentally, notice that while Suresh and Guttag [167] make explicit reference to the sampling process, i.e. the act of drawing specific observations from the target population, we embed this aspect directly in the measurement Equations (3.2). What we propose in the following is a simple and explicit mathematical formalisation of the framework.

First, it is useful to illustratively represents the building blocks of biases as discussed in Section 2.3 via Directed Acyclic Graphs[1] (similar to [143, 145, 148]). In general, in order to provide an intuitive grasp on interesting mechanisms and patterns, we shall use the following notation: $R$ are variables representing *resources* of individuals — be them economic resources, or personal talents and skills — which are relevant for the problem, i.e. they directly impact the target $Y$; $A$ denote variables indicating sensitive attributes, such as ethnicity, gender, etc.; $P_R$ stand for proxy variables that we have access to instead of the original variable $R$; $Q$ denote additional variables, that may or may not be relevant for the problem (i.e. impacting $Y$) and that may or may not be impacted either by $R$ or $A$, e.g. the neighborhood one lives in.

In particular, Figure 3.1 shows four minimal graph representations of historical, omitted variable, and measurement biases that make use of the notation just introduced. Historical bias occurs when the relevant variable $R$ is somehow impacted by sensitive feature $A$. Omitted variable bias occurs when, for some reason, we omit the relevant variable $R$ from our dataset and we employ another variable which happens to be impacted by $A$. Measurement bias occurs when the relevant variable $R$ is, in general, free of bias, but we cannot access it. Therefore, we employ a proxy $P_R$ (which is typically strongly dependent on $R$) that *is* impacted by sensitive characteristic $A$. Measurement bias can also occur on the target variable $Y$ when we only have access to a (biased) proxy $P_Y$ of the phenomenon we want to predict.

The system of Equations (3.5) formalises the relationships between variables used to simulate specific forms of biases. Notice that the variables $N_.$ and $B_.$ denote independent random variables, either continuous-valued ($N_.$) or integer-valued ($B_.$). Intuitively, they represent the sources of variability in the generated dataset, while the structure of the equations imposes the (desired) dependence among the relevant variables. The continuous variable $R$ could represent, e.g., salary, and the discrete variable $Q$ —which can take $K+1$ different values— could represent a zone in a city. Indeed, $Q$ is distributed as a binomial variable in $\{0,\dots,K\}$, with Bernoulli marginal probability $p_Q$ dependent on $R$ and $A$ via a simple logistic function. The binary sensitive variable ($A$) is distributed as a Bernoulli $\{0,1\}$ variable, with $p_A$ proportion. Variable $S$ is an auxiliary variable used to effectively generate a binary target $Y$ by thresholding $S$.

[1] We will provide additional details about DAG in Section 4.5.1.

$$A = B_A, \quad B_A \sim \mathscr{B}er(p_A); \tag{3.5a}$$
$$R = -\beta_h^R A + N_R, \quad N_R \sim Gamma(k_R, \theta_R); \tag{3.5b}$$
$$Q = B_Q, \quad B_Q \mid (R,A) \sim \mathscr{B}in(K, p_Q(R,A)), \tag{3.5c}$$
$$p_Q(R,A) = \text{sigmoid}\left(-(\alpha_{RQ} R - \beta_h^Q A)\right);$$
$$S = \alpha_R R - \alpha_Q Q - \beta_h^Y A + N_S, \quad N_S \sim \mathscr{N}(0, \sigma_S^2); \tag{3.5d}$$
$$Y = \mathbf{1}_{\{S > \overline{P_S}\}}. \tag{3.5e}$$

When simulating measurement bias, either on resources $R$ or on target $Y$,[2] we are going to use the following *proxies* as noisy (and biased) substitutes for the actual variables:

$$P_R = R - \beta_m^R A + N_{P_R}, \quad N_{P_R} \sim \mathscr{N}(0, \sigma_{P_R}^2); \tag{3.6a}$$
$$P_S = S - \beta_m^Y A + N_{P_S}, \quad N_{P_S} \sim \mathscr{N}(0, \sigma_{P_S}^2); \tag{3.6b}$$
$$P_Y = \mathbf{1}_{\{P_S > \overline{P_S}\}}. \tag{3.6c}$$

We denote with $\beta$'s the parameters governing the presence and strength of each form of bias, while we use $\alpha$'s for parameters that regulate the relationships among variables not directly involving bias introduction. By varying the values of the parameters, we are able to generate different aspects of biases as follows:

- $\beta_h^j$ determines the presence and amplitude of the *historical bias* on the variable $j \in \{R, Q, Y\}$;
- $\beta_m^j$, when the proxy $P_j$ is used instead of the original variable $j$, governs the intensity of *measurement bias* on $j \in \{R, Y\}$;
- $\alpha_R$, $\alpha_Q$ control the linear impact on ($S$ and thus) $Y$ of $R$ and $Q$, respectively; $\alpha_{RQ}$ represents the intensity of the dependence of $Q$ on $R$.

Additionally, in order to account for *representation bias*, we undersample the group $A = 1$. The amount of undersampling is governed by the parameter $p_u$ defined as the proportion of the under-represented group $A = 1$ with respect to the majority group $A = 0$. We draw the undersampling *conditioned* on $R$ by selecting the $A = 1$ individuals with lower values for $R$. Finally, simulating *omission bias* is as simple as dropping the variable $R$ from the set of features the model uses to estimate $Y$.

[2] We use the distribution mean of $P_S$, denoted by $\overline{P_S}$, to derive binary values for $Y$ and its proxy $P_Y$ to avoid predominantly positive or negative labels for one of the groups in the dataset.

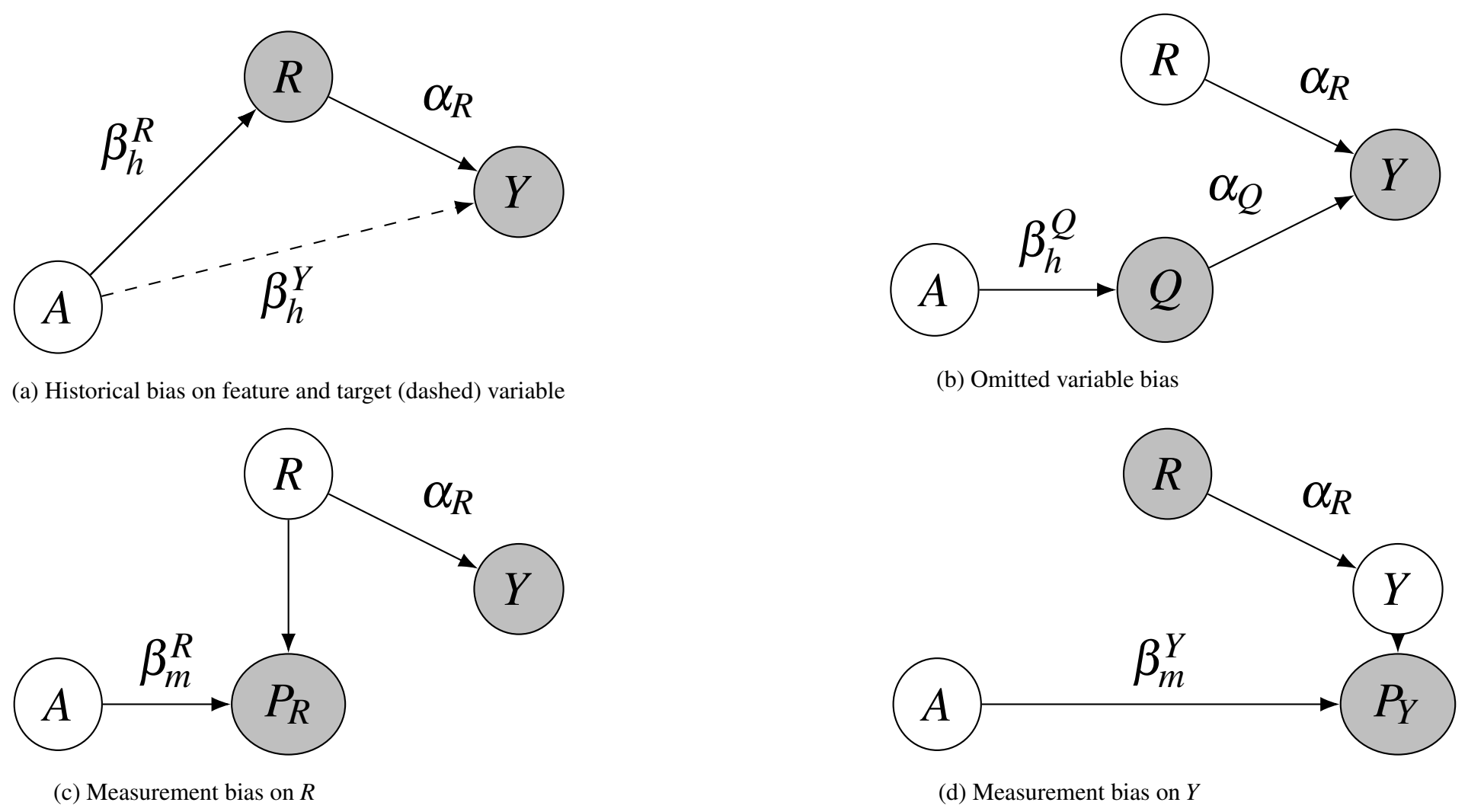

Figure 3.1 Illustrative representation of biases. Grey-filled circles represent variables employed by the model $\hat{f}$.

Figure 3.2a shows an illustrative representation of biases that can arise during the algorithmic development, i.e. downstream with respect to the dataset generation. In this work, our main focus lies in the generation of a biased dataset. Nevertheless, one could easily simulate this type of bias as well, e.g. by translating the graph into the following equation:

$$\hat{Y} = \hat{f}(\widetilde{X}) - \beta^{\hat{Y}} A, \tag{3.7}$$

where $\beta^{\hat{Y}}$ denotes the magnitude of *algorithmic bias*.

Figure 3.2b shows an illustrative representation of biases that can arise during the algorithmic deployment, i.e. downstream with respect to the dataset generation. Similar to the *algorithmic bias* described above, one could easily simulate this type of bias as well, e.g. by translating the graph into the following equation:

$$D = r(\hat{Y}) - \beta^D A, \tag{3.8}$$

where $\beta^D$ and represents the magnitude of *deployment bias*.

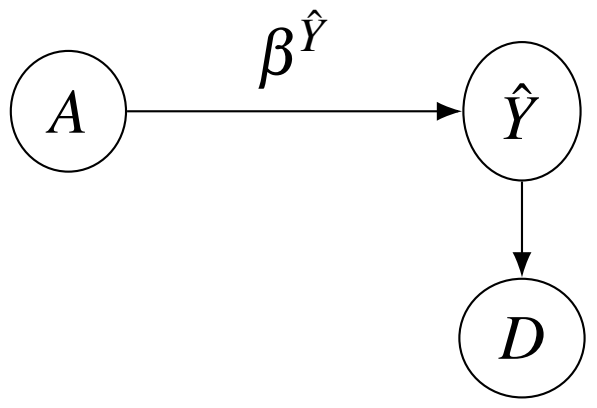

(a) Algorithmic bias on $\hat{Y}$

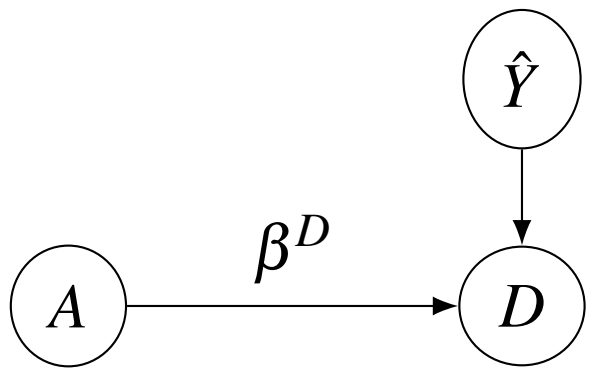

(b) Deployment bias on $D$

Figure 3.2 Illustrative representation of biases involving the algorithm and its deployment – see also Equations (3.7) and (3.8).

We want to make clear that what we propose is by no means the more general modelling framework to generate any form of bias: we just propose one possibility to formalise different types of bias mathematically, guided by two principles: *simplicity* and *exhaustiveness* with respect to bias types. One can easily think of many variations (some of which are also included in the code implementation) of the modeling framework generating the same bias types in a different way. For example, one could use other distributions for $N_R$, $N_{P_R}$, $N_S$, and $N_{P_S}$. Other alternatives lie in the functional forms relating the variables, which are here assumed mostly linear for sake of simplicity. Moreover, in some cases, even the mechanism underlying the biases can be more complicated than the simple shift in the expected values: e.g. historical bias could be due to a different variance of $R$ among sensitive groups, or, in general, to the fact that the distribution of $R \mid A$ varies with the specific value of $A$. Further, note that we understand bias as systematic differences across groups, which is in line with [14]. Thus, as can be seen in Equations (3.5) and (3.6), we multiply the bias parameters $\beta$ with the sensitive attribute $A$ and do not make any explicit assumptions on the underlying causal paths.

## 3.3 `Bias on Demand` as a Open Source Tool

A large set of experiments, as well as the code to create new ones, is publicly available at `https://github.com/rcrupiISP/BiasOnDemand`. The package can be installed via `pip` and used to generate synthetic datasets with various types of bias in just a few lines of code. We lay the groundwork for developing novel tools and strategies, such as systems for detecting and identifying different types of bias, as well as implementing specific bias mitigation techniques. We hope that our toolkit will encourage the research community to undertake further studies using synthetic datasets where real-world datasets are lacking.

We will illustrate some applications of `Bias On Demand` in Sections 5.7 and 8.3. However, preceding these demonstrations, we must first introduce fairness metrics (Chapter 4) for quantifying fairness and explore Fairness Mitigation approaches (Chapter 5).

# 4

# Fairness Metrics Landscape in Machine Learning

The complex nature of biases, as well as the corresponding moral and technical perspectives, results in a large number of possible fairness metrics [164, 168]. In the last few years an incredible number of definitions have been proposed, formalizing different perspectives from which to assess and monitor fairness in decision-making processes. A popular tutorial presented at the Conference on Fairness, Accountability, and Transparency in 2018 was titled "21 Fairness Definitions and their Politics" [140]. The number has grown since then. The proliferation of fairness definitions is not *per se* an issue: it reflects the evidence that fairness is a multi-faceted concept, and concentrates on different meanings and nuances, in turn depending in complex ways on the specific situation considered. As it is often the case with moral and ethical issues, conflicts are present between different and typically equally reasonable positions [88, 30, 172].

Researchers, in proposing definitions, have focused on different intuitive notions of "unfair decisions", often considered as the ones impacting people in different ways on the basis of some personal characteristics, such as gender, ethnicity, age, sexual or political, or religious orientations, considered to be *protected*, or *sensitive*. Relationships and interdependence of these sensitive variables with other features useful for making decisions lie at the very heart of these ambiguities and nuances, and entangled in complex ways the ultimate aim of any model, i.e. making efficient decisions, and the desired goal of not allowing unfair discrimination to impact people's lives.

Fairness notions proposed in the literature are usually classified in broad areas, such as: definitions based on parity of statistical metrics across groups identified by different values in protected attributes (e.g. male and female individuals, or people in different age groups); definitions focusing on preventing different treatment for individuals considered similar with respect to a specific task; definitions advocating the necessity of finding and employing causality among variables in order to really disentangle unfair impacts on decisions.

These three broad classes can be further seen as the result of two important distinctions:

(a) observational vs. causality-based criteria;
(b) group (or statistical) vs. individual (or similarity-based) criteria.

Distinction (a) discriminates criteria that are purely based on the observational distribution of the data from criteria that try to first unveil causal relationships among the variables at play in a specific situation (mainly through a mixture of domain knowledge and opportune inference techniques) and then assess fairness. Distinction (b) discriminates criteria that focus on equality of treatment among groups of people from criteria requiring equality of treatment among couples of similar individuals.

This chapter deals with the aforementioned fact that the number of different metrics for fairness introduced in the literature has boomed. The researcher or practitioner first approaching this facet of ML may easily feel confused and somehow lost in this zoo of definitions. These multiple definitions capture different aspects of the concept of fairness but, to the best of our knowledge, there is still no clear understanding of the overall landscape where these metrics live. We aim to take a step in the direction of analyzing the relationship among the metrics and trying to put order in the fairness landscape. Table 4.1 provides a rough schematic list of fairness metrics discussed throughout this thesis. As general references for the definition of fairness in ML we refer to the book by Barocas, Hardt and Narayanan [13] and to more compact surveys [176, 135, 142, 46, 47, 128, 126].

We shall mainly deal with the most common problem in supervised ML, the binary classification with a single sensitive attribute, and we shall only make a brief reference to the additional subtleties arising when multiple sensitive attributes are present. Despite this huge simplifications, the landscape of fairness definitions is nevertheless extremely rich and complex.

In the literature, some attempts to survey existing fairness definitions in ML are present [135, 176, 128, 13], but, to the best of our knowledge, none has focused solely on metrics trying to disentangle and analyze in depth the interdependence and relationships among them. Our contribution don't lie in the exhaustiveness of the taxonomy as in the attempt to clarify the nuances in the fairness metrics landscape with respect to the twofold dimensions listed above, namely group vs. individual notions and observational vs. causality-based notions. We focus mainly on general and qualitative description of the fairness metrics landscape: building upon rigorous definitions we want to highlight incompatibilities and links between apparently different concepts of fairness.

Table 4.1 **Fairness metrics.** Qualitative schema of the most important fairness metrics discussed throughout the paper.

| | | notion | use of $Y$ | condition |
|---|---|---|---|---|
| group fairness | | Demographic Parity | - | equal acceptance rate across groups |
| | | Conditional Demographic Parity | -* | equal acceptance rate across groups in any strata |
| | error parity | Equal Accuracy | ✓ | equal accuracy across groups |
| | | Equality of Odds | ✓ | equal FPR and FNR across groups |
| | | Predictive Parity | ✓ | equal precision across groups |
| individual fairness | | FTU/Blindness | - | no explicit use of sensitive attributes |
| | | Fairness Through Awareness | -* | similar people are given similar decisions |
| causality-based fairness | | Counterfactual Fairness | - | an individual would have been given the same decision if she had had different values in sensitive attributes |
| | | path-specific Counterfactual Fairness | - | same as above, but keeping fixed some specific attributes |

* there are exceptions to these cases where $Y$ is actually employed, e.g. CDP conditioning on $Y$ becomes Equality of Odds, and there are notions of individual fairness that use a similarity metric defined on the target space [20].

## 4.1 Toy Examples and Notation

For the sake of clarity, we will refer to two illustrative examples within the banking sector: credit lending decisions, and job recruiting.

As in the previous chapter, we call $A$ the categorical random variable representing the protected attribute, which for reference we shall take to be gender, we label with $X$ all the other (non-sensitive) random variables that the AI-based decision-making system is going to use to provide its yes-or-not decisions $\hat{Y} = f(X,A) \in \{0,1\}$; while we label $Y \in \{0,1\}$ the ground truth target variable that needs to be estimated/predicted – typically by minimizing some loss function $\mathscr{L}(Y,\hat{Y})$. $\widetilde{X} = (\widetilde{X}_1,\ldots,\widetilde{X}_d) = (X,A)$ collectively represent all the $d$ features involved in the problem. We denote with lowercase letters the specific realizations of random variables, e.g. $\{(x_1,y_1),\ldots,(x_n,y_n)\}$ represent a dataset of $n$ independent realizations of $(X,Y)$. We employ calligraphic symbols to refer to domain spaces, namely $\mathscr{X}$ denotes the space where features $X$ live. (Technically, since we employ uppercase letters to denote random variables, it would be more proper to say that $\mathscr{X}$ is the image space of $X : \Omega \to \mathscr{X}$, where $\Omega$ is the event space, and that $x \in \mathscr{X}$. We shall nevertheless use this slight abuse of notation for the sake of simplicity).

While we shall for simplicity refer to a binary sensitive attribute $A$, what follows in principle holds for any categorical sensitive attribute. Instead, things change when the target variable is multi-class, ordered multi-class (i.e. ranking problems), or real-valued (i.e. regression problems). Most of the literature on fairness in ML deals with binary classification problems, and we shall stick to this scenario for the sake of clarity. While most metrics defined for binary targets can be generalized in the multi-class case, for ranking and regression different criteria are needed. We refer the interested reader to the literature devoted to fairness in multi-class classification [57], ranking [187], and regression [20, 3], and to other surveys on fairness in ML [128, 41].

## 4.2 Individual Fairness

### 4.2.1 Similarity-Based Criteria

Individual fairness is embodied in the following principle: *similar individuals should be given similar decisions*. This principle deals with the comparison of single individuals rather than focusing on groups of people sharing some characteristics. On the other hand, group fairness starts from the idea that there are *groups of people* potentially suffering biases and unfair decisions, and thus tries to reach equality of treatment for groups instead of individuals.

The first attempt to deal with a form of individual fairness was presented in [63], where this concept is introduced as a Lipschitz condition on the map $f$ from the feature space to the model space:

$$dist_Y(\hat{y}_i, \hat{y}_j) < L \times dist_{\widetilde{X}}(\widetilde{x}_i, \widetilde{x}_j), \tag{4.1}$$

where $dist_Y$ and $dist_{\widetilde{X}}$ denote suitable distances in the target space and feature space, respectively, and $L$ is a constant. Loosely speaking, *a small distance in feature space (i.e. similar individuals) must correspond to a small distance in decision space (i.e. similar outcomes).*

The concept of individual fairness is straightforward, and it certainly resonates with our intuitive notion of "equality". Formula (4.1) provides also an easy way to assess whether the decision model $\hat{Y} = f(\widetilde{X})$ satisfies such concept.

However, one drawback of this type of definition lies in the ambiguous concept of "similar individuals". Indeed, defining a suitable distance metric $dist_{\widetilde{X}}$ on feature space to embody the concept of similarity on ethical grounds alone is almost as difficult as defining fairness in the first place. In addition, assigning equal importance to all the features in $X$ space is inconsistent with the objective function of the classification model. The point is that one should come up with a distance that captures the essential features for determining that target, and that does not mix with sensitive attributes. But, again, this is not very different from defining what is fairness in a specific situation. Take, e.g., the job recruiting framework: what identifies the couples of individuals that should be considered similar and thus given the same chance of being recruited? Maybe the ones with the same level of skills and experience irrespective of anything else?

One possibility is to define similar individuals as couples belonging to different groups with respect to sensitive features (e.g. male and female) but with the same values for all the other features. With this choice, what we are requiring is that its outcome should be unchanged if we take an observation and we only change its protected attribute $A$. This concept is usually referred to as Fairness Through Unawareness (FTU)) or *blindness* [176], and it is expressed as the *requirement of not explicitly employing protected attributes in making decisions*.

Notice that this idea is very likely one of the first that one may think of when asked for, e.g., a decision-making process that does not discriminate against gender: not to explicitly use gender to make decisions, i.e. $\hat{Y} = f(X)$. Indeed, this concept is also referred to as *disparate treatment*, i.e. there is disparate treatment whenever two individuals sharing the same values of non-sensitive features but differing on the sensitive ones are treated differently [14, 184].

Unfortunately, despite its compelling simplicity, fairness through unawareness comes not without flaws. Firstly, if it is true that reaching FTU is straightforward, it is not that easy to assess. The problem of *bias assessment* in a model consists in measuring whether the model decisions are biased given a set of realizations of $(A, X, \hat{Y})$, the corresponding values of $Y$ are needed as well for some criteria. In this setting, it is tricky to measure whether FTU holds, the main reason being that it is more a request on how the model works rather than a request on properties of the output decisions. One possible candidate metric is the following [188]:

$$\text{consistency} = 1 - \frac{1}{n}\left(\sum_{i=1}^{n}\left|\hat{y}_i - \frac{1}{k}\sum_{x_j \in kNN(x_i)} \hat{y}_j\right|\right). \tag{4.2}$$

Namely, for each observation $(x_i, \hat{y}_i)$ it measures how much the decision $\hat{y}_i$ is close to the decisions given to its $k$ nearest neighbors $kNN(x_i)$ in the $\mathscr{X}$ space (notice that also in the computation of $kNN$ one has to choose a distance function on $\mathscr{X}$) [122]. It may happen that the $k$ neighbors of, say, a male individual are all males: in this case consistency (4.2) would in fact be equal to 1, but this does not prevent the model from explicitly using $A$ in making decisions. Another possibility would be to compute the following metric [20]:

$$\frac{1}{n_1 n_2}\sum_{\substack{a_i=1,\\ a_j=0}} e^{-dist(x_i,x_j)}|\hat{y}_i - \hat{y}_j|, \tag{4.3}$$

which measures the difference in decisions among men and women weighted by their similarity in feature space: the higher its value the higher the difference in treatment for couples of similar males and females (the term $e^{-dist(x_i,x_j)}$ can be substituted with any measure of similarity of the points $x_i$ and $x_j$). On the other hand, it is much easier to assess FTU if we also have access to the model: we could create a synthetic dataset by flipping $A$ (this holds if $A$ is binary, but it is easy to come up with generalizations to the multiclass case): $\{(a'_1, x_1), \ldots, (a'_n, x_n)\}$, feeding it to the model to get the corresponding outcomes $\hat{y}'_1, \ldots, \hat{y}'_n$, and then compute the average $\frac{1}{n}\sum_{i=1}^{n}|\hat{y}_i - \hat{y}'_i|$.

However, the main drawback of FTU is the following: it does not take into account the possible interdependence between $A$ and $X$. Other features may contain information on the sensitive attribute, thus explicitly removing the sensitive attribute is not sufficient to remove its information from the dataset. Namely, it may be that in the actual dataset there is a very low chance that a male and a female have similar values in all the (other) features, since gender is correlated with some of them. One may or may not decide that (some of) these correlations are legitimate (e.g. gender may be correlated with income, but, depending on the problem, one may decide that the use of income, even if correlated with gender, is not a source of unfair discrimination; in Section 4.5 we will dwell more on this issue, namely that there may be some information correlated to the sensitive attribute but still considered "fair"). This is one of the reasons why the definition of a issue-specific distance is crucial for a more refined notion of individual fairness.

One straightforward way to deal with correlations is to develop a model that is blind not only with respect to the sensitive attribute but also to all the other variables with sufficiently high correlation with it. This is a method known as *suppression* [101]. Apart from the obvious issue in defining a good threshold for correlation above which a predictor should be removed, this approach has the main drawback in the potentially huge loss of legitimate information that may reside in features correlated with the sensitive attribute.

Counterfactual frameworks [111, 44], that will be more thoroughly discussed in Section 4.5, provide a clear way in which this similarity should be thought of: a male individual is similar to himself in the counterfactual world where he is a woman. Notice that this is crucially different from fairness through unawareness approach: a male individual transported in the counterfactual world where he is a woman will have differences in other features as well, and such differences are precisely due to the causal structure among the variables (very roughly speaking, this is the "causal way" to account for correlations). As an example, in the now popular Judea Pearl's assessment of the 1973 UC Berkeley admission case [24, 144, 145] where there was different admission rate between men and women, being a woman "causes" the choice of higher demanding departments, thus impacting the admission rate (in causal theory jargon, department choice is said to be a *mediator* from gender to admission) [13]. In this case, a male transported in the counterfactual world where he is a woman would have himself chosen higher demanding departments, thus this feature would change as well. In this case, a simple gender-blind model would not guarantee individual fairness. Indeed attribute flipping in general does not produce valid counterfactuals [28]. More details on this will be given in Section 4.5.

Turning to the general similarity-based definition (4.1), some work has been done to address the problem of finding a suitable similarity metrics in feature space. E.g. [63] and [98, 92] introduce the possibility to learn a issue-specific distance from data and from the contributions of domain experts. Indeed, the simple idea of using standard similarities, e.g. related to the euclidean distance on feature space, does not take into account the trivial fact that some feature are more important than other in determining the relationship of an individual to specific target. Namely, for two applicants for a loan, the difference in income is much more important than the difference in, say, age, or even profession. Thus, judging what does it mean to be similar *with respect to a specific task* is not that simple, and is in some sense connected also to the ground truth target variable $Y$.

Indeed, some have proposed [20] a notion of individual fairness as a penalty function – to be added to the risk minimization fitting – which shifts the concept of similarity from the feature space to the target space: namely two individuals are deemed similar if they have a similar value of the target variable. Thus, this notion of individual fairness relies directly on the target attribute to define a task-specific distance. Of course, this definition is prone to biases possibly present in the target.

Broadly speaking, the Lipschitz notion (4.1) is sometimes referred to as Fairness Through Awareness (FTA), as opposed to Fairness Through Unawareness: in fact, even if they share the same principle of treating equally similar individuals, FTA is generally meant to use similarity metrics that are problem and target specific, i.e. that derive from an "awareness" of the possible impact, while FTU is a simple recipe that does not depend on the actual scenario.

## 4.3 Group Fairness

Group fairness criteria are typically expressed as conditional independence statements among the relevant variables in the problem. There are three main broad notions of observational group fairness, *independence*, *separation*, *sufficiency* [13]. Independence is strictly linked to what is known as *Demographic Parity (DP)* or *Statistical Parity*, separation is related to *Equality of Odds (EOdd)* and its relaxed versions, while sufficiency is connected to the concept of *calibration* and *Predictive Parity (PP)*.

There is a crucial aspect that discriminates independence criteria from the others: *independence criteria rely only on the distribution of features and decisions, namely on $(A, X, \hat{Y})$, while separation and sufficiency criteria deal with error rate parities, thus making use of the target variable $Y$ as well.* This is an important thing to have in mind when trying to find your way in the zoo of fairness criteria in a specific case study: when using separation-based or sufficiency-based criteria one must be careful to check whether the target variable $Y$ is itself prone to sources of bias and unfairness.

In the example of credit lending, if $Y$ represents the decision of a loan officer, than separation-based and sufficiency-based criteria must be used with particular care, since $Y$ can be itself biased against some groups of people. Actually, even if $Y$ stands for the repayment (or lack of it) of the loan, a form of *representation bias* is very likely at work: we have that information only on applicants that received the loan in the first place, and these are almost surely not representative of the whole population of applicants.

Sometimes the term *Disparate Mistreatment (DM)* is used to refer to all group fairness criteria relying on the disparity of errors, thus, in general, all group metrics dealing with comparisons among decisions $\hat{Y}$ and true outcomes $Y$ [184].

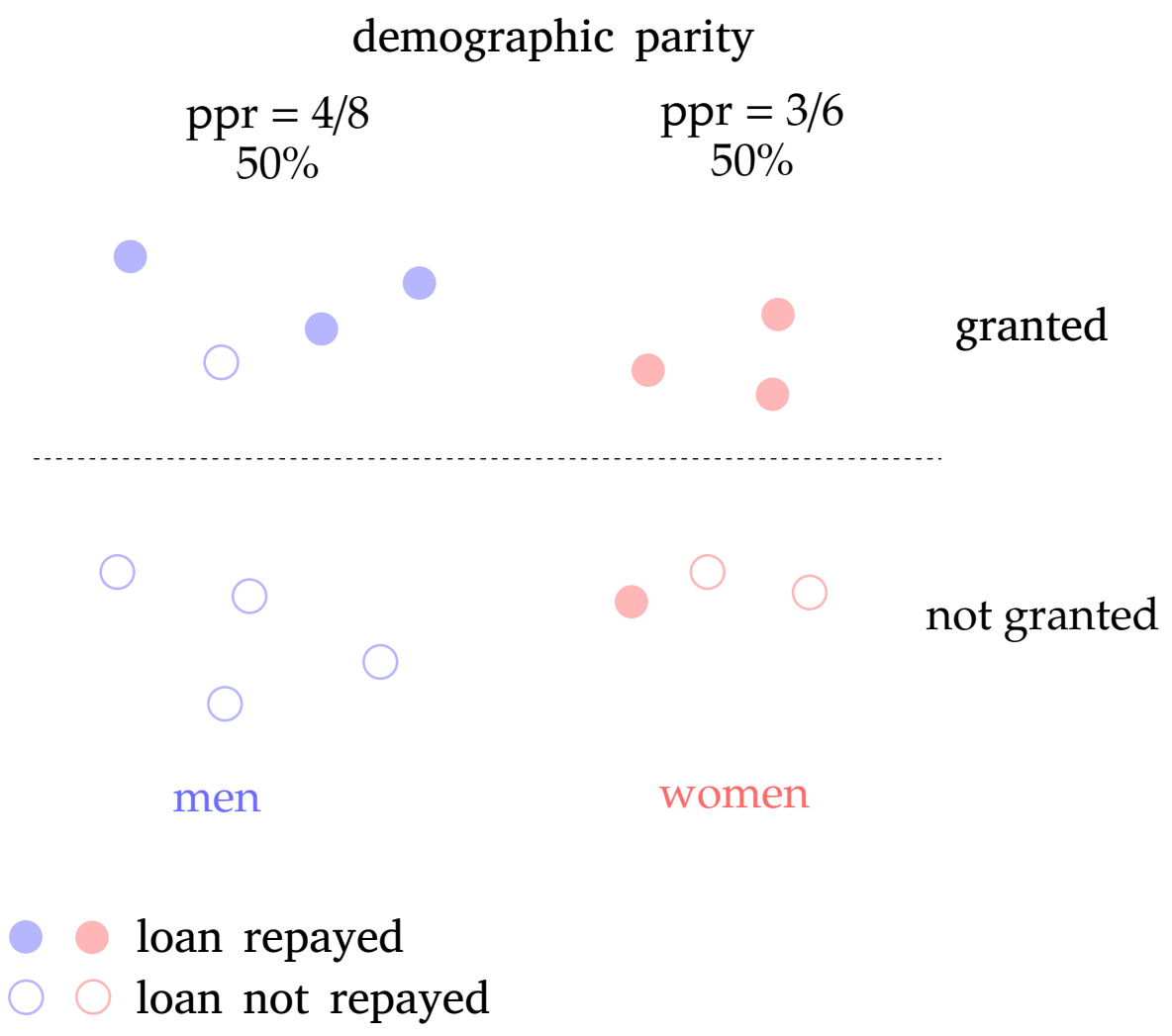

Figure 4.1 example of demographic parity in gender in credit lending toy model.

Moreover, independence is said to be a *non-conservative* measure of fairness [153] — meaning that it forces to change the *status quo* — since it is in general not satisfied by the perfect classifier $\hat{Y} = Y$ (apart from the trivial case when $Y \perp\!\!\!\perp A$). Error rate parities, on the other hand, are *conservative*, since they trivially hold for the perfect predictor.

### 4.3.1 Independence

The criterion of independence [13] states that the *decisions should be independent of any sensitive attribute*:

$$\hat{Y} \perp\!\!\!\perp A. \tag{4.4}$$

This can be also expressed as follows — in the binary classification assumption:

$$P(\hat{Y} = 1 \mid A = a) = P(\hat{Y} = 1 \mid A = b), \quad \forall a, b \in \mathscr{A}, \tag{4.5}$$

i.e. the ratio of loans granted to men should be equal to the ratio of loans granted to women.

The ratio of favorite outcomes is sometimes known as Positive Prediction Ratio (PPR), thus independence is equivalent to requiring the same positive prediction ratio across groups identified by the sensitive features. This form of independence is usually known as *demographic parity*, *statistical parity*, or sometimes as *group fairness*. DP is a very common concept in the literature on fairness in ML [13, 45], and it was actually employed even before the now common nomenclature was introduced [100]. Figure 4.1 shows a very simple visualization of a model reaching demographic parity among men and women.

If some group has a significantly lower positive prediction ratio with respect to others, we say that demographic parity is not satisfied by the model decisions $\hat{Y}$. In order to have a single number summarizing the amount of disparity, it is common to use either the maximum possible difference or the minimum possible ratio of positive prediction ratios: a difference close to 0 or a ratio close to 1 indicates a decision system *fair* with respect to $A$ in the sense of DP. Typically, some tolerance is considered by employing a threshold below (or above) which the decisions are still considered acceptable.

In the literature, it is common to cite the "$4/5$ rule" or "80% rule": the selection rate of any group should be not less than $4/5$ than one of the groups with the highest selection rate, i.e DP ratio greater than 80%. This is a reference to the guidelines by the US Equal Employment Opportunity Commission (EEOC) [171, 66], and it is often cited as one of the few examples of a legal framework relying on a specific definition of fairness, in particular on the notion of Demographic Parity. However, it is usually overlooked that the guidelines explicitly state that this should be considered only as a rule of thumb, with many possible exceptions, and in particular they prescribe that business or job-related necessities can justify a lower DP ratio, depending on the circumstances. This is actually a step away from pure DP towards more individual notions, such as those of Conditional Demographic Parity (CDP) and error parities, that shall be discussed in sections 4.3.1, 4.3.2, 4.3.3.

**Subtleties of Demographic Parity**

The meaning of DP is intuitive only to a superficial analysis. For example, one may at first think that removing the sensitive attribute from the decision making process is enough to guarantee independence and thus demographic parity. In general, this is not the case. Take the credit lending example and assume that, for whatever reason, women tend to actually pay back their loans with higher probability with respect to men. If this is the case, it is reasonable to assume that a rating variable that we call $R$ will be higher for women than for men, on average. In this scenario, the sensitive attribute gender ($A$) is correlated with rating. If the model uses only rating (but not gender) to compute its decision, there will be a higher rate of loans granted to women than to men, resulting in a demographic *disparity* among these groups. In this case, if you want to reach DP, the model needs to *favor men over women*, granting loans to men with a lower rating threshold with respect to women.

Thus, because of interdependence of $X$ and $A$, not only it is not enough to remove the sensitive attribute from the decision making process, but if you want to have DP you need, in general, to *treat different groups in different ways*, precisely in order to compensate for the (unwanted) effect of this dependence. This is somehow the opposite of an intuitive notion of fairness!

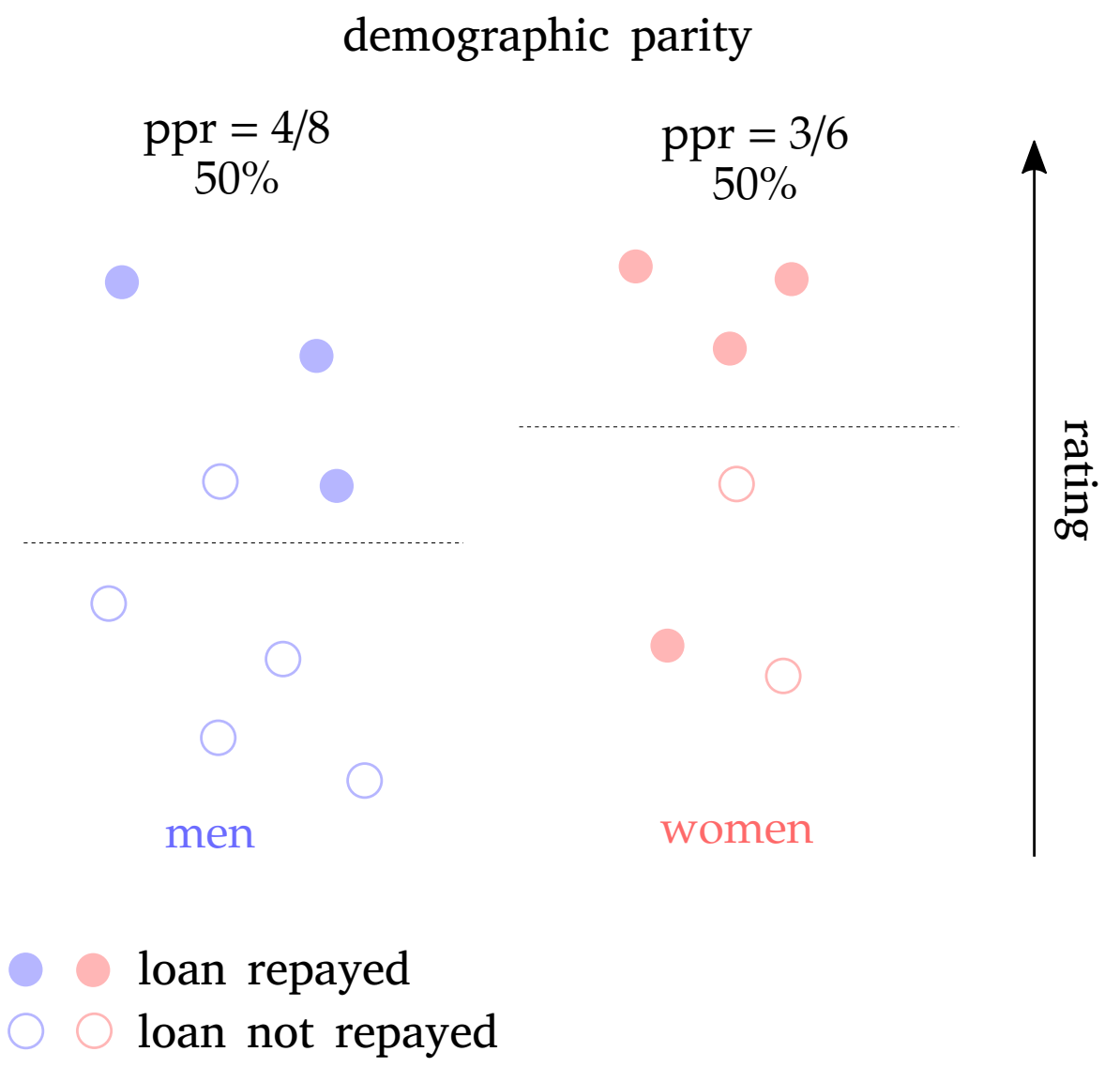

Figure 4.2 example of a subtlety of demographic parity: in order to reach demographic parity between men and women and still using rating as fundamental feature, one must use a different threshold between the groups, thus manifestly treating differently men and women.

Figure 4.2 displays this example: to reach equal ppr, one must use a different rating threshold for each group.

This, in turn, reveals another subtlety: even if your dataset and your setting does not apparently contain any sensitive features, discrimination could lurk in via correlations to sensitive features that you are not even collecting.

We want to stress here what we think is a crucial aspect: another possibility to reach DP would be to use *neither gender nor rating in making decisions*, i.e. trying to remove all gender information from the dataset. Notice that this could be problematic for the accuracy of your decisions, since it's plausible that, by removing all variables correlated to $A$, information useful to estimate the target is lost as well. This is called *suppression*, and was discussed already in Section 4.2, together with the concept of *fair representation*, by which one tries to remove all sensitive information from the dataset while keeping as much useful information as possible. As we shall argue in Section 4.4, this approach allows to satisfy DP, while not favouring any subgroups with respect to $A$.

Another possible flaw is the following: if it is true that women repay their loans with higher probability, is it really fair to have demographic parity between men and women? Why the bank should agree to grant loans with the same rate to groups that actually pay back their loans with different probabilities? Should we stick to actual repayment rates or we should ask *why* these probabilities are different and if this is possibly due to gender discrimination in the first place?

We try to summarize, in a non-exhaustive list, a set of scenarios in which it might be reasonable to take into account demographic parity, among other metrics:

- when you want to actively *enforce* some form of equality between groups, irrespective of all other information. Indeed there are some characteristics that are widely recognized to be independent *in principle* of sensitive attributes, e.g. intelligence and talent, and there may be the need to enforce independence in problems where the decisions are fundamentally linked to those characteristics; more in general, there may be reasons to consider unfair any relation among $A$ and $Y$, even if the data (objective data as well) is telling differently;
- (intertwined with the previous) when you deem that in your specific problem, there are hidden historical biases that impact in a complex way the entire dataset;
- when you cannot trust the objectivity of the target variable $Y$, then demographic parity still makes sense, while many other metrics don't (e.g. separation);
- for strategic reasoning when aiming to generate long-term advantages for underprivileged population subgroups. In Chapter 9, we will explore this concept further.

Justifying the use of DP as a metric with the idea of enforcing an equality that should hold in principle has been discussed in the literature [70, 84, 153]. We would like to spend here a word of care in this respect, namely on the fact that favouring a group does indeed guarantee that the equality is reached in that specific use case, but it may trigger mechanisms that actually *amplify* the bias that we intended to break. In the example of job recruiting, imposing the decision maker to increase the acceptance rate on a group of people with (relatively) low skill levels, will increase the likelihood that applicants hired from this group will perform poorly relative to the average, effectively perpetuating the bias in the way in which this group is perceived by others. In general, assessing the future impact of imposing a fairness condition, i.e. considering fairness as a dynamical process rather than a one-shot condition, is a subtle issue by itself [94, 117, 89].

#### Conditional Demographic Parity

Another version of the independence criteria is that of *Conditional Demographic Parity* [103]. In the example given above, we may think that a fairer thing to do, with respect to full independence, is to require independence of the decision on gender only for men and women with the same level of rating. In other words, if a man and a woman have both a certain level of rating, we want them to have the same chance of getting the loan. This goes somewhat in the direction of being an individual form of fairness requirement, since parity is assessed into smaller groups with respect to the entire sample.

Formally, this results in requiring $\hat{Y}$ *independent of A given R*,

$$\hat{Y} \perp\!\!\!\perp A \mid R, \tag{4.6}$$

or, in other terms:

$$P(\hat{Y} = 1 \mid A = a, R = r) = P(\hat{Y} = 1 \mid A = b, R = r), \quad \forall a, b \in \mathscr{A}, \forall r. \tag{4.7}$$

This seems a very reasonable requirement in many real-life scenarios. For example, if you think of a recruitment setting where you don't want to bias women against men, but still you want to recruit the most skilled candidates, you may require your decision to be independent of gender but conditional on a score based on the curriculum and past work experiences: among people with an "equivalent" set of skills, you want to recruit men and women with the same rate.
In other words, the only disparities that you are willing to accept between male and female candidates are those justified by curriculum and experience.

Unfortunately, also in this case one must be really careful, because the variable that you are conditioning on *might itself be a source of unfair discrimination*. For example, it might well be that rating is higher for women not because they actually pay back their debts more likely than men, but because the rating system is biased against men. And this may be due to a self-fulfilling prophecy: if men have lower ratings they may receive loans with higher interest rates and thus have a higher probability of not paying them back, in a self-reinforcing loop.

Moreover, it may not be so straightforward to select the variables to condition on: why condition on rating and not, e.g., on level of income, or profession, etc...?

Finally, in line with what has been outlined above for demographic parity, one may argue that women's curricula and work experiences tend to be on average different than those of male candidates for historical reasons and for a long-lasting (and die-hard) man-centered society. This may suggest that plain demographic parity could be more appropriately enforced in this case to reach "true" equality.

Incidentally, notice that pushing to the extreme the notion of conditional demographic parity, i.e. conditioning on *all the (non-sensitive) variables*, one has

$$\hat{Y} \perp\!\!\!\perp A \mid X, \tag{4.8}$$

i.e.

$$P(\hat{Y}=1 \mid A=a, X=x) = P(\hat{Y}=1 \mid A=b, X=x), \quad \forall a,b \in \mathscr{A}, \forall x \in \mathscr{X}; \tag{4.9}$$

meaning that a male and a female *with the same value for all the other features* must be given the same outcome. This criterion is reachable by a gender-blind model, thus is strictly connected to the notion of FTU. As mentioned above – and in more details in Section 4.4 – conditioning is equivalent to restrict the groups of people among which we require parity, thus is a way to go in the direction of obtaining an individual criterion.

### 4.3.2 Separation

Independence and conditional independence do not make use of the true target $Y$. What if, instead of conditioning over rating $R$ we condition on the target $Y$? This is equivalent to requiring the independence of the decision $\hat{Y}$ and gender $A$ separately for individuals that actually repay their debt and for individuals that don't. Namely, among people that repay their debt (or don't), we want to have the same rate of loan granting for men and women.

This concept has been called separation [13]:

$$\hat{Y} \perp\!\!\!\perp A \mid Y. \tag{4.10}$$

In other terms

$$P(\hat{Y}=1 \mid A=a, Y=y) = P(\hat{Y}=1 \mid A=b, Y=y), \quad \forall a,b \in \mathscr{A},\ y \in \{0,1\}. \tag{4.11}$$

Equivalently, *disparities in groups with different values of* $A$ (male and female) *should be completely justified by the value of* $Y$ (repayment or not).

As in the conditional independence case, this seems a very reasonable fairness requirement, *provided that you can completely trust the target variable*. Namely, one should be extremely careful to check whether the target $Y$ is not itself a source of bias.

For example, if $Y$ instead of reflecting true repayment was the outcome of loan officers' decision on whether to grant the loans, it could incorporate bias, thus we it would be risky to assess fairness with direct comparisons with $Y$. Moreover, as said above, even in the objective case where $Y$ is the actual repayment, a form of selection bias would likely distort the rate of repayment.

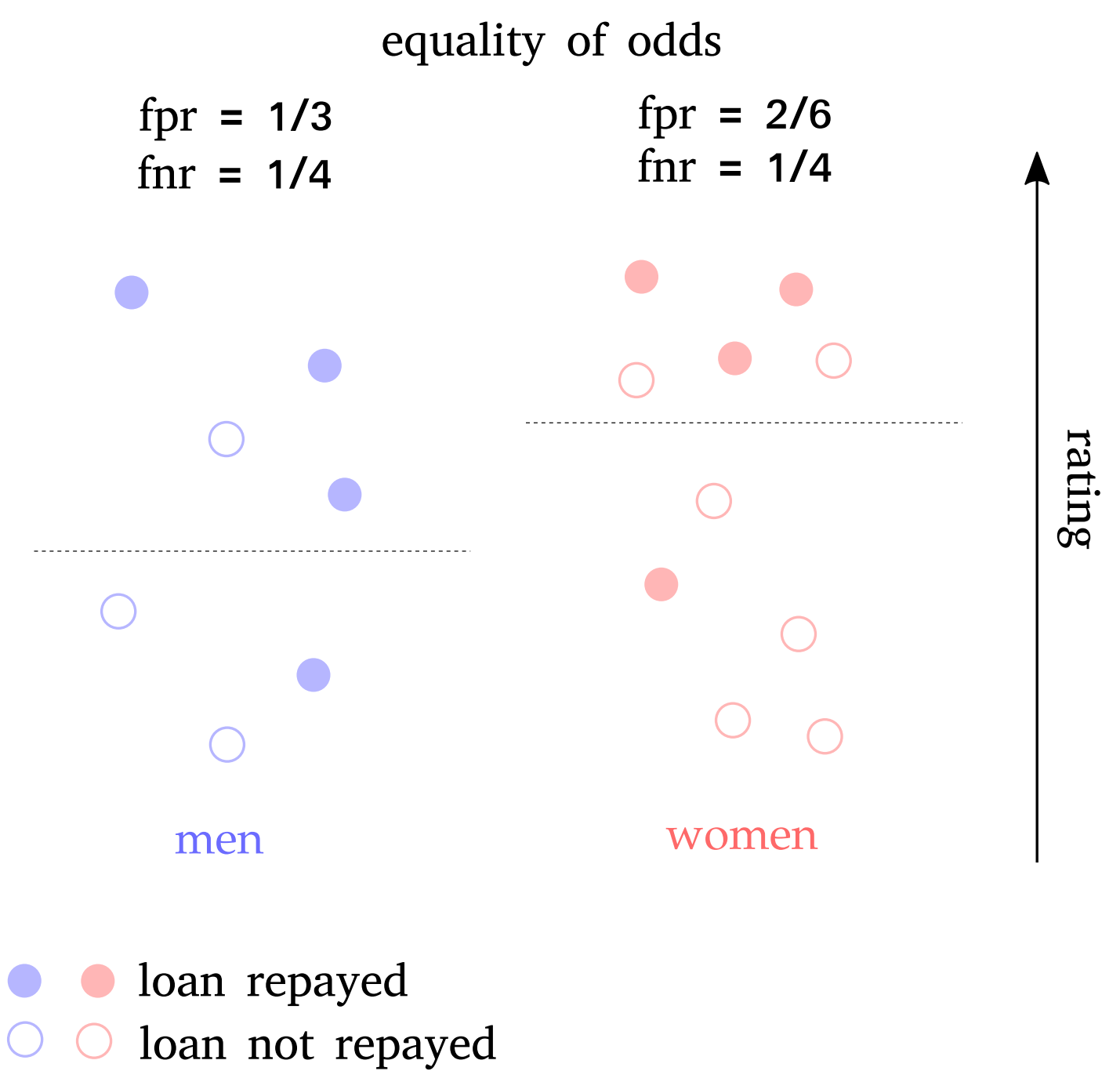

Figure 4.3 Example of Equality of Odds between men and women: false negative and false positive rates must be equal across groups.

We can express separation in terms of what are known in statistics as *type I* and *type II* errors. Indeed, it is easy to see that the two conditions in equation (4.11) (one for $y = 1$ and one for $y = 0$) are equivalent to requiring that the model has *the same false positive rate and false negative rate across groups identified via* $A$. False positives and false negatives are precisely type I and type II errors, respectively. Namely, individuals that are granted loans but are not able to repay, and individuals that are able to repay but are not granted loans. This is known as *Equality of Odds* [81], and is thus the requirement of having the same type I and type II error rates across relevant groups, as displayed in Figure 4.3.

There are two relaxed version of this criterion:

- *Predictive Equality*: equality of false positive rate across groups,

$$P(\hat{Y} = 1 \mid A = a, Y = 0) = P(\hat{Y} = 1 \mid A = b, Y = 0), \quad \forall a, b \in A,$$

- *Equality of Opportunity*: equality of false negative rate across groups,

$$P(\hat{Y} = 0 \mid A = a, Y = 1) = P(\hat{Y} = 0 \mid A = b, Y = 1), \quad a, b \in \mathscr{A}.$$

While demographic parity, and independence in general, focuses on equality in terms of acceptance rate (loan grating rate), Equality of Odds, and separation in general, focuses on equality in terms of error rate: the model is fair if it is as efficient in one group as it is in the other.

The difference between Predictive Equality and Equality of Opportunity is the perspective from which equality is required: Predictive Equality takes the perspective of people that won't repay the loan, while Equality of Opportunity takes the one of people that will repay. Depending on the problem at hand, one may consider either of these two perspectives as more important. For example, Predictive Equality may be considered when we want to minimize the risk of innocent people from being erroneously arrested: in this case it may be reasonable to focus on the parity of among innocents. Both in the credit lending and job listing examples, on the other hand, it may be reasonable to focus on Equality of Opportunity, i.e. on the parity among people that are indeed deserving.

Here follows a non exhaustive set of situations in which separation criteria may be suitable:

- when your target variable $Y$ is an objective ground truth;
- when you are willing to make discrimination as long as they are justified by actual trustable data;
- when you do not want to actively enforce an "ideal" form of equality, and you want to be as equal as possible given the data.

We can summarize by saying that separation, being a concept of parity given the ground truth outcome, is a notion that takes the point of view of people that are subject to the model decisions, rather than that of the decision maker. In the next subsection, instead, we shall take into account the other side of the coin, i.e. parity given the model decision.

### 4.3.3 Sufficiency

Sufficiency [13] takes the perspective of people that are given the same model decision, and requires parity among them irrespective of sensitive features.

While separation deals with error rates in terms of fraction of errors over the ground truth, e.g. the number of individuals whose loan request is denied among those who would have repaid, sufficiency takes into account the number of individuals who won't repay among those who are given the loan.

Mathematically speaking, this is the same distinction you have between recall (or true positive rate) and precision, i.e. $P(\hat{Y} = 1 \mid Y = 1)$ and $P(Y = 1 \mid \hat{Y} = 1)$, respectively.

A fairness criterion that focuses on this type of error rate is called *Predictive Parity* [45], also referred to as *outcome test* [176, 135]:

$$P(Y=1 \mid A=a, \hat{Y}=1) = P(Y=1 \mid A=b, \hat{Y}=1), \quad \forall a,\, b \in \mathscr{A}, \tag{4.12}$$

i.e. the model should have the same precision across sensitive groups. If we require condition (4.12) to hold for the case $Y=0$ as well, then we get the following conditional independence statement:

$$Y \perp\!\!\!\perp A \mid \hat{Y},$$

which is referred to as *sufficiency* [13].

Predictive Parity, and its more general form of sufficiency, focuses on error parity among people who are given the same decision. In this respect, Predictive Parity takes the perspective of the decision maker, since they group people with respect to the decisions rather than the true outcomes. Taking the credit lending example, the decision maker is indeed more in control of sufficiency rather than separation, since parity given decision is something directly accessible, while parity given truth is known only in retrospect. Moreover, as we have discussed above, the group of people who are given the loan ($\hat{Y}=1$) is less prone to selection bias than the group of people who repay the loan ($Y=1$): indeed we can only have the information of repayment for the $\hat{Y}=1$ group, but we know nothing about all the others ($\hat{Y}=0$).

As you may notice, going along a similar reasoning, one can define other group metrics, such as Equality of Accuracy across groups: $P(\hat{Y}=Y \mid A=a) = P(\hat{Y}=Y \mid A=b)$, for all $a,b \in \mathscr{A}$, i.e. focusing on unconditional errors, and others [176].

### 4.3.4 Group Fairness on Scores

In most cases, even in classification setting, the actual output of a model is not a binary value, but rather a *score* $S \in \mathbb{R}$, estimating the probability, for each observation, to have the target equal to the favorable outcome (usually labelled 1). Then, the final decision is made by the following $t$-threshold rule:

$$\hat{Y} = \begin{cases} 1 & S \geq t, \\ 0 & S < t. \end{cases} \tag{4.13}$$

Most of the things we have outlined for group fairness metrics regarding $\hat{Y}$ can be formulated for the joint distribution $(A, Y, S)$ as well:

$$\begin{aligned} &\text{independence} \quad S \perp\!\!\!\perp A, \\ &\text{separation} \quad S \perp\!\!\!\perp A \mid Y, \\ &\text{sufficiency} \quad Y \perp\!\!\!\perp A \mid S. \end{aligned} \tag{4.14}$$

These formulations provide stronger constraints on the model with respect to the analogous with $\hat{Y}$, e.g. the condition $S \perp\!\!\!\perp A$ is sometimes referred to as *strong demographic parity* [95, 142].

For instance, all three criteria in their form (4.5), (4.11), (4.12), i.e. with constraints on the joint distribution $(A, Y, \hat{Y})$ can effectively be satisfied by defining group dependent thresholds $t$ on black-box model outcomes $S$ (a technique that goes under the name of *postprocessing* [13]), while this is not the case for (4.14).

Instead of requiring conditions on the full distribution of $S$ as in (4.14), something analogous to the "binary versions" of group fairness criteria have been defined simply by requiring parity of the *average score* [108]. *Balance of the Negative Class*, defined as

$$E(S \mid Y = 0, A = a) = E(S \mid Y = 0, A = b), \quad \forall a, b \in \mathscr{A}, \tag{4.15}$$

corresponds to Predictive Equality (PE), while *Balance of the Positive Class* (same as (4.15) with $Y = 1$) corresponds to Equality of Opportunity. Notice that these two last definitions fall into the one given in Section 4.3.2 when $S = \hat{Y}$. Requiring both balances is of course equivalent to requiring EO.

*AUC parity*, namely the equality of the area under the ROC for different groups identified by $A$, can be seen as the analogous of the equality of accuracy.

Finally, notice that the score formulation of sufficiency is connected to the concept of *calibration*. Calibration holds when

$$P(Y = 1 \mid S = s) = s, \tag{4.16}$$

i.e. if the model assigns a score $s$ to 100 people then, on average, $100 \times s$ of them will actually be positive. Writing down the condition for sufficiency in score:

$$P(Y = 1 \mid S = s, A = a) = P(Y = 1 \mid S = s, A = b), \quad \forall a, b \in \mathscr{A}, \forall s, \tag{4.17}$$

it is clearly related to what can be called *Calibration within Groups* [108]:

$$P(Y = 1 \mid A = a, S = s) = s \tag{4.18}$$

– actually a consequence of it. This condition is in general not so hard to achieve, and it is often satisfied "for free" by most models [13].

### 4.3.5 Incompatibility Statements

It is interesting to analyse the relationships among different criteria of fairness. In Section 4.4 we shall discuss in detail the connections between individual and group notions, while here we focus on differences among various group criteria. We have already seen that each of them highlights one specific aspect of an overall idea of fairness, and we may wonder what happens if we require to satisfy multiple of them at once. The short answer is that it is not possible except in trivial or degenerate scenarios, as stated by the following propositions drawn from the literature [45, 108, 13, 153, 165]:

1. *if $Y$ is binary, $Y \not\perp\!\!\!\perp S$ and $Y \not\perp\!\!\!\perp A$, then* ***separation*** *and* ***independence*** *are incompatible*.
   In other words, to achieve both separation and independence, the only possibility is that either the model is completely useless ($Y \perp\!\!\!\perp S$), or the target is independent of the sensitive attribute ($Y \perp\!\!\!\perp A$), which implies an equal base rate for different sensitive groups. Namely, if there is an imbalance in groups identified by $A$, then you cannot have both EO and DP holding.
2. Analogously: *if $Y \not\perp\!\!\!\perp A$, then* ***sufficiency*** *and* ***independence*** *cannot hold simultaneously*.
   Thus, if there is an imbalance in base rates for groups identified by $A$, then you cannot impose both sufficiency and independence.
3. Finally, *if $Y \not\perp\!\!\!\perp A$ and the distribution $(A, S, Y)$ is strictly positive, then* ***separation*** *and* ***sufficiency*** *are incompatible*.
   Meaning that separation and sufficiency can both hold either when there is no imbalance in sensitive groups (i.e. the target is independent of sensitive attributes), or when the joint probability $(A, S, Y)$ is degenerate, i.e. — for binary targets — when there are some values of $A$ and $S$ for which only $Y = 1$ (or $Y = 0$) holds, in other terms when the score exactly resolves the uncertainty in the target (as an example, the perfect classifier $S = Y$ always trivially satisfies both sufficiency and separation).

Notice that proposition 3 reduces to the following, more intuitive statement if the classifier is also binary (e.g. when $S = \hat{Y}$): *if $Y \not\perp\!\!\!\perp A$, $Y$ and $\hat{Y}$ are binary, and there is at least one false positive prediction, then separation and sufficiency are incompatible.* Moreover, it has be shown [108] that *Balance of Positive Class, Balance of Negative Class and Calibration within Groups can hold together only if either there is no imbalance in groups identified by A or if each individual is given a perfect prediction (i.e.* $S \in \{0,1\}$ *everywhere).*

When dealing with incompatibility of fairness metrics, the literature often focuses on the 2016 COMPAS — a recidivism prediction instrument developed by Northpoint Inc. — recidivism case [9], now become a case study in the fairness literature. Indeed, the debate on this case is a perfect example to highlight the fact that there are *different and non-compatible notions of fairness*, and that this may have concrete consequences on people. While we refer to the literature for a thorough discussion of the COMPAS case [179, 45], we here just point out that in the debate there were two parties, one stating that the model predicting recidivism was *fair* since it satisfied Predictive Parity by ethnicity, while the other claiming it was *unfair* since it had different false positive and false negative rates for black and white individuals. Chouldechova showed that [45], if $Y \not\perp\!\!\!\perp A$, i.e. if the true recidivism rate is different for black and white people, then Predictive Parity and Equality of Odds cannot both hold, thus implying that a reflection on which of the two (in general of the many) notions is more important to be pursued in that specific case must be carefully considered.

Summarizing, apart from trivial or peculiar scenarios, the three families of group criteria above presented are not mutually compatible.

### 4.3.6 Multiple Sensitive Features

Generally speaking, all the definitions and results we gave in previous sections are subject to the fact that the sensitive feature $A$ is represented by a single categorical variable. If, for a given problem, we identify more than one characteristic that we need to take into account as sensitive or protected – say $(A_1,\ldots,A_l)$ – we can easily assess fairness on each of them separately. Unfortunately this approach — sometimes called *independent group fairness* [182] — is in general not enough: even if fairness is achieved (in whatever sense) separately on each sensitive variable $A_i$, it may happen that some subgroups given by the intersection of two or more $A_i$'s undergo unfair discrimination with respect to the general population. This is sometimes referred to as *intersectional bias* [51], or, more specifically, *fairness gerrymandering* [105].

To prevent bias from occurring in *all the possible subgroups identified by all* $A_i$*'s* one can simply identify a new feature $A = (A_1,\ldots,A_l)$, whose values are the collection of values on all the sensitive attributes, and require fairness constraints on $A$ — *intersectional group fairness* [182].

This "trick" indeed solves the problem of intersectional bias, at least theoretically. Still, issues remain at a computational and practical level, whose two main reasons are:

- the exponential increase of the number of subgroups when adding sensitive features,
- the fact that, with finite samples, many of the subgroups will be empty or with very few observations.

These two aspects imply that assessing (group) fairness with respect to multiple sensitive attributes may be unfeasible in most practical cases. Since the presence of many sensitive features is more of a norm than it is an exception, this actually represents a huge problem, that the literature on fairness in ML has barely begun to address [182, 105, 106, 31].

In Chapter 6, we will tackle the challenge of mitigating fairness across multiple sensitive features.

## 4.4 Group vs. Individual fairness

The most common issue with group fairness definitions is the following: since group fairness requires to satisfy conditions only on average among groups, it leaves room to bias discrimination *inside the groups*. As we argued in the example of Section 4.3 referring to Figure 4.2, one way of reaching DP is to use a different rating threshold for men and women: this means that there will be a certain range of ratings for which men will receive the loan, and women won't. More formally, conditionally on rating there is no independence among $A$ and $\hat{Y}$. In general, to reach group fairness one may "fine tune" the interdependence of $A$ and $X$ to reach parity on average, but effectively producing differences in subgroups of $A$.

Notice that this is precisely what individual fairness is about. In the example above, a men and a woman that have the same rating may be treated differently, thus violating the individual notion of fairness.

As already discussed in Section 4.3, this is only one possibility: one may as well reach DP by using neither gender nor rating, and grant loan on the basis of other information, provided it is independent of $A$. In this case, there will be no group discrimination *and* there won't be any subgroup discrimination as well.

However, we can say that, if we want to reach DP by using as much information of $Y$ as possible contained in $(X,A)$, i.e. minimizing the risk $E\mathscr{L}(f(X,A),Y)$, then it is unavoidable to have some form of disparate treatment among people in different groups with respect to $A$ whenever $X \not\perp\!\!\!\perp A$. *Fair affirmative action* [63] is the name given to the process of requiring

DP while trying to keep as low as possible the amount of disparate treatment between people having similar $X$.

To clarify the general picture, we can put the different notions of (observational) fairness in a plane with two *qualitative* dimensions (see Figure 4.4): 1. to what extent a model is fair at the individual level, 2. how much information of $A$ is retained in making decisions. The first dimension represents to what extent two individuals with similar overall features $X$ are given similar decisions: the maximum value is reached by models blind on $A$ (FTU). These are the models that are also using all information in $X$, irrespective of the interdependence with $A$, thus FTU-compliant models will use all information contained in $\widetilde{X}$ apart from the information that is contained in $A$ only. The minimum value in this dimension is reached by models that satisfy DP. Models using suppression methods, being blind to both $A$ and other features with high correlation with $A$, are individually fair in the sense of preventing disparate treatment. In so doing, they can exploit more or less information of $A$ with respect to general DP-compliant models depending on how many correlated variables are discarded. However, the price to pay for discarding variables is in terms of errors in approximating $Y$, which is not highlighted in this plot. Notice that, of course, full suppression – i.e. removing all variables dependent on $A$ – trivially satisfies the condition $\hat{Y} \perp\!\!\!\perp A$, i.e. it is DP-compliant as well. In Figure 4.4, we label with DP a general model that tries to maximize performance while satisfying a DP constraint, without any further consideration.

Models satisfying CDP are somewhat in-between, of course depending on the specific variables considered for conditioning. They guarantee less disparate treatment than unconditional DP, and they use more information of $A$ by controlling for other variables possibly dependent on $A$.

Notice that approaches such as fair representation (see Section 4.2), where you try to remove all information of $A$ from $X$ to get new variables $Z$ which are as close as possible to being independent of $A$, produce decision systems $\hat{Y} = f(Z)$ that are not, in general, individually fair. This is due to the simple fact that, precisely to remove the interdependence of $A$ and $X$ while keeping as much information of $X$ as possible, two individuals with same $X$ and different $A$ will be mapped in two distinct points on $\mathscr{Z}$, thus having, in general, different outcomes. Referring again to the credit lending example, suppose that we have $R = g(A) + U$, with $g$ a complicated function encoding the interdependence of rating and gender, and $U$ some other factor independent of $A$ representing other information in $R$ "orthogonal" to $A$. In this setting, the variable $Z$ that we are looking for is precisely $U$. Notice that $U$ is indeed independent of $A$, thus any decision system $\hat{Y} = f(U)$ satisfies DP, but given two individuals with $R = r$ nothing prevents them from having different values of $U$. In other terms, you

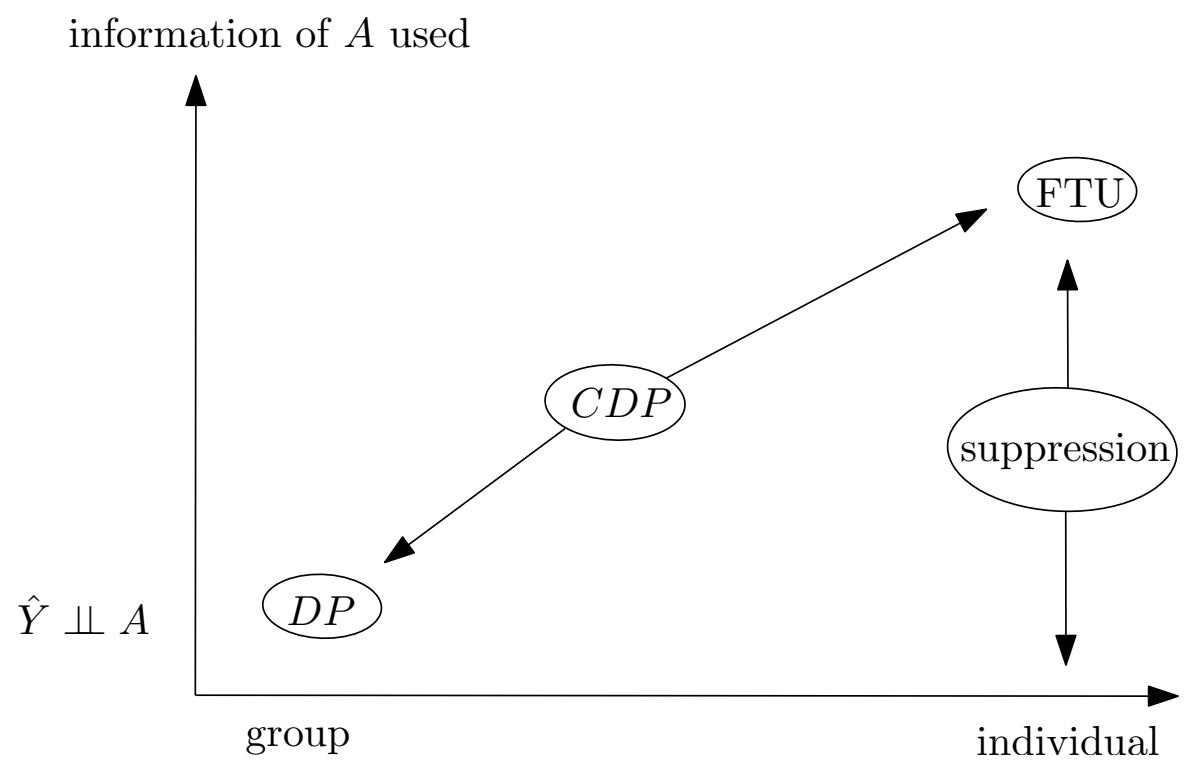

Figure 4.4 Landscape of observational fairness criteria with respect to the group-vs-individual dimension and the amount of information of $A$ used (via $X$).

*need* to have some amount of disparate treatment to guarantee DP *and* employ as much information as possible to estimate $Y$.

Figure 4.5 shows a *qualitative* representation of observational metrics with respect to the amount of information of $A$ (through $X$) that is used by the model, and the predictive performance [21]. Notice that DP can be reached in many ways: e.g. a constant score model, namely a model accepting with the same chance all the individuals irrespective of any feature, is DP-compliant (incidentally, it is also individual), a model in which all the variables dependent on $A$ have been removed (a full suppression), or a model where DP is reached while trying to maximize performance (e.g. through fair representations). All these ways differ, in general, in terms of the overall performance of the DP-compliant model.

FTU-compliant models, on the other hand, by employing all information in $X$ will be, in general, more efficient in terms of model performance.

Incidentally, notice that this discussion is to be taken at a qualitative level, one can come up with scenarios in which, e.g., models satisfying DP have higher performances than models FTU-compliant (think, e.g., of a situation in which $Y \perp\!\!\!\perp A$ and $X \not\!\perp\!\!\!\perp A$).

The (apparent) conflict that holds in principle between individual and group families of fairness notions is debated in the literature [70, 84, 26]. Our analysis provides arguments to the claim that individual and group fairness are not in conflict *in general* [26], since they lie on the same line whose extreme values are separation (i.e. unconditional independence) and FTU (i.e. individuals are given the same decisions on the basis of non-sensitive features only), with conditional metrics ranging between them. Thus, the crucial aspect is assuming and deciding *at the ethical and legal level* what are the variables that are “allowed” in specific scenarios: given those, we place ourselves on a point on this line reachable both by choosing an appropriate distance function on feature space (i.e. employing an individual concept) and by requiring parity among groups conditioned on those variables (a group notion).

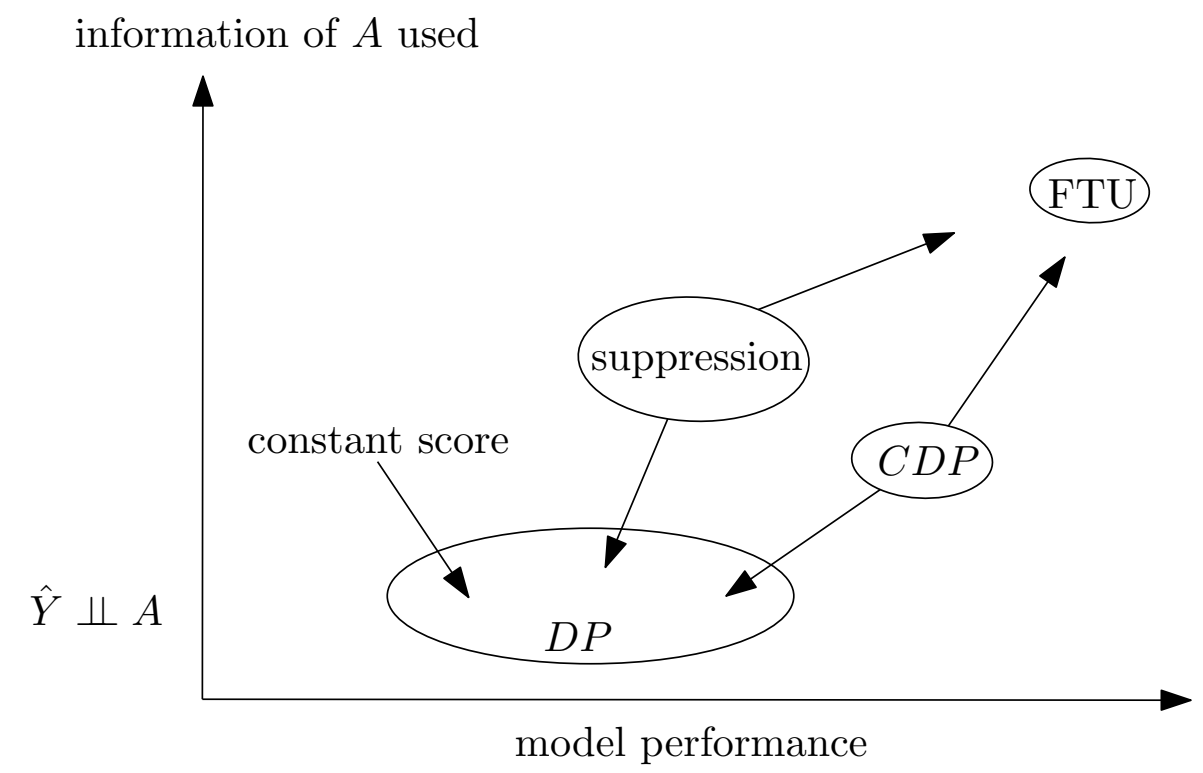

Figure 4.5 Landscape of observational fairness criteria with respect to the model performance dimension and the amount of information of $A$ used (via $X$).

Indeed, when referring to the conflict between individual and group families one is committing a slight "abuse of notation", since the actual clash is rather on the *assumptions* regarding what is to be considered fair in a specific situation, than on metrics *per se*. Namely, one usually considers individual (group) concepts as the ones in which more (less) interdependence of $X$ and $A$ is allowed to be reflected in the final decisions. In the credit lending example, if income is correlated with gender, the issue whether it is fair to allow for a certain gender discrimination as long as justified by income can be seen as an instance of the conflict between individual or group notions, but it is rather a conflict about the underlying ethical and legal assumptions [26].

## 4.5 Causality-Based Criteria

Another important distinction in fairness criteria is the one between *observational* and *causality-based* criteria. As we have seen, observational criteria rely only on observed realizations of the distribution of data and predictions. In fact, they focus on enforcing equal metrics (acceptance rate, error rate, etc. . . ) for different groups of people. In this respect, they don't make further assumptions on the mechanism generating the data and suggest to assess fairness through statistical computation on observed data.

Causality-based criteria, on the other hand, try to employ domain and expert knowledge in order to come up with a casual structure of the problem, through which it becomes possible to answer questions like "what would have been the decision if that individual had a different gender?". While counterfactual questions like this seem in general closer to what one may intuitively think of as "fairness assessment", the observational framework is on the one hand easier to assess and constrain on, and on the other more robust, since counterfactual criteria are subject to strong assumptions about the data and the underlying mechanism generating them [56, 55].

As we argued above (Section 4.2), answering to counterfactual questions is *very different* from taking the feature vector of, e.g., a male individual and just flip the gender label and see the consequences in the outcome. The difference lies precisely in the causal chain of "events" that this flip would trigger. If there are some features related, e.g., to the length of the hair, or the height, then it is pretty obvious that the flip of gender should come together with a change in these two variables as well, with a certain probability. And this may be the case for other — less obvious but more relevant — variables. This also suggests why counterfactual statements involve *causality relationships* among the variables. In general, to answer counterfactual questions, one needs to know the causal links underlying the problem.

However, as major drawback, once given the casual structure there are many counterfactual models compatible with that structure (actually infinite), and the choice of one of them is in general not falsifiable with observations [148, 55], since, by definitions, they deal with claims that have not observable consequences.

Indeed, causality-based criteria can be formulated at least at two different levels, with increasing strength of assumptions: at the level of *interventions* and at the level of *counterfactuals*.

While in principle causality-based criteria are preferable over observational ones, since they assess the causal impact of the sensitive variables on the decision, they are difficult to use in practical cases. Roughly speaking, we can say that inferring causal relationships and validate them is grounded on more solid basis in "hard" disciplines, like physics, chemistry, biology, rather than when complex social behaviors are at play. Unfortunately, fairness issues arise precisely in complex social settings. However, when the number of relevant variables is small and the phenomenon is well understood, it may be feasible to come up with a sound causal model (think e.g. of the widespread use of causality in economics).

### 4.5.1 Causal Models

Before introducing fairness definitions based on the underlying causal structure of a problem, we briefly introduce the necessary theoretical framework of *causal graphs* and Structural Causal Models (SCM) or Structural Equation Models [145, 144, 148, 13, 111].

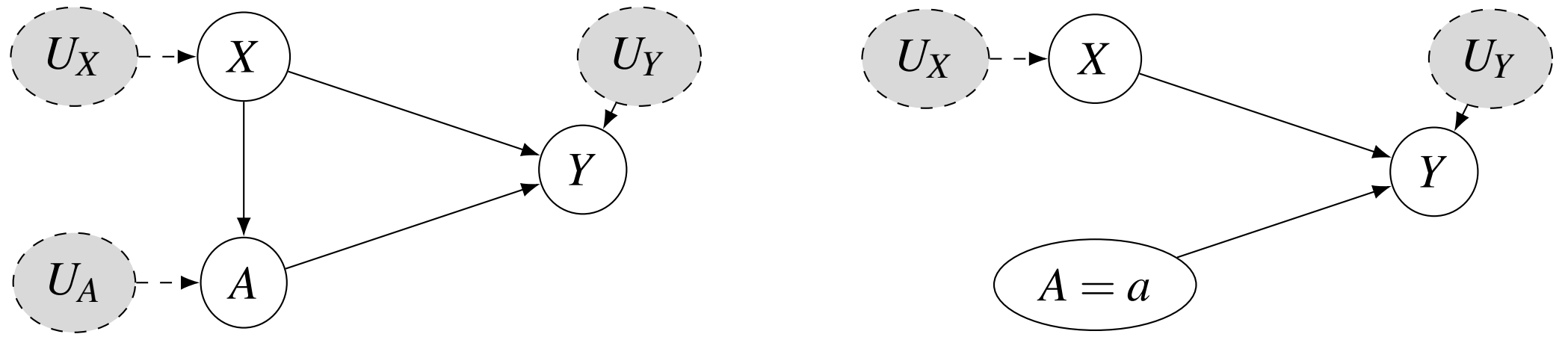

Figure 4.6 Example of causal graph with 3 endogenous variables (left). Intervention on $A$ is expressed via a different graph (right) where all incoming edges in $A$ are removed and the variable $A$ is set to the value $a$.

We model the underlying causal relationships among features by means of a Directed Acyclic Graph (DAG) $G=(V,E)$, with $V$ set of vertices (or nodes) and $E$ set of directed edges (or links) — see Figure 4.6. Nodes of the graph $G$ represent the variables $\widetilde{X}$ used as predictors in the model. Moreover, we denote with $U$ exogenous or *latent* variables, representing factors not accounted for by the features $\widetilde{X}$. In causal graph theory, edges in the graph represent not only conditional dependence relations, but are interpreted as the causal impact that the source variable has on the target variable. To work at the level of *interventions* this is all that is needed. Modeling causal knowledge is complex and challenging since it requires an actual understanding of the relations, beyond statistical evidence. Different causal discovery algorithms have been proposed to identify causal relationships from observational data through automatic methods [76].

To work at the level of *counterfactuals* one needs much more than the sole graph structure: a Structural Causal Model is a a triplet $(\widetilde{X},U,F)$ where $F$ are a set of assignments of the form

$$\widetilde{X}_i = f_i(\mathbf{pa}(\widetilde{X}_i),U_i), \quad i=1,\dots,d; \tag{4.19}$$

where the functions $f_i$ represent the precise way in which the parents of each node variable $\widetilde{X}_i$ ($\mathbf{pa}(\widetilde{X}_i)$) — namely the variables having a direct causal impact over $X_i$ in the assumed (or inferred) graph $G$ — together with the latent variable $U_i$, influence the value of $X_i$. These relations are called Structural Equations (SE) and, besides describing which variables causally impact which (that is already encoded in the graph $G$), they also determine *how* these relations work. One typical simplifying assumption on SCM is to work with Additive Noise Models [148]:

$$\widetilde{X}_i = g_i(\mathbf{pa}(\widetilde{X}_i)) + U_i, \quad i=1,\dots d.$$

In general, assumptions like this are needed in order to be able to perform computations using SCM starting from observational data. A detailed presentation of the assumptions underlying the use of graphical models and SCMs (e.g. modularity, markovianity, causal sufficiency, etc…) is beyond the scope of this thesis [148, 80, 56, 144].

As mentioned above, one of the limits of counterfactual models is that, since they make "predictions" regarding something that has not happened, the choice of a specific counterfactual model is in general not falsifiable with observable data [148, 55].

### 4.5.2 Fairness Metrics in Causality-Based Settings

Fairness at the level of interventions can be formally expressed as follows [107]:

$$P(\hat{Y} = 1 \mid do(A = a), X = x) = P(\hat{Y} = 1 \mid do(A = b), X = x), \quad \forall a, b \in \mathscr{A}, x \in \mathscr{X}. \quad (4.20)$$

In words, if you take a random individual, force it to be, e.g., female ($do(A = a)$) and she happens to have $X = x$, you want to give him the same chance of acceptance as for a random individual forced to be male ($do(A = b)$) that also happens to have $X = x$. We refer to it as *Intervention Fairness*. Incidentally, notice that it may be unfeasible to compute quantities in (4.20) on the basis of observational data and the graph, without assuming an SCM (or without physically performing the experiment); indeed it may be that the events whose probability needs to be computed in order to get the post-intervention distribution have no observations in the dataset at hand [144, 143].

The same requirement can be set at the counterfactual level and is known as CounterFactual Fairness (CFF) [111]:

$$P(\hat{Y}_{A \leftarrow a} = 1 \mid A = a, X = x) = P(\hat{Y}_{A \leftarrow b} = 1 \mid A = a, X = x), \quad \forall a, b \in \mathscr{A}, x \in \mathscr{X}, \quad (4.21)$$

which, in words, reads: if you take a random individual with $A = a$ and $X = x$ and the same individual if she had $A = b$, you want to give them the same chance of being accepted.

The difference between the two levels is subtle but important: roughly speaking, when talking about interventions one is considering the average value over exogenous factors $U$ that, after the intervention, are compatible with the conditioning, while counterfactuals consider only the values of $U$ that are compatible with the factual observation (namely, the distribution $\mathrm{P}(U \mid A = a, X = x)$) to begin with, and then perform the intervention and consider the consequences. In other words, counterfactuals consider only events that take into account actual observed value of $A$ (and $X$ as well). As a further clarification [111], suppose the following structural equation: $X = A + U$. Then equation (4.20) compares two individuals with $U = x - a$ and $U' = x - b$, i.e. two different individuals that, with the interventions $do(A = a)$ and $do(A = b)$ both happen to have the observed value $X = x$. Equation (4.21) instead take the individual with $U = x - a$ and then perform the interventions $do(A = a)$ (which has no consequences) and $do(A = b)$ which implies on $X = b + (x - a) \neq x$.

Equations (4.20), (4.21) make use of Pearl's *do*-calculus [143–145], where $do(A = a)$ — the *intervention* — consists in modifying the causal structure of the problem by exogenously setting $A = a$, thus removing any causal paths impacting on $A$ – the theoretical equivalent of randomized experiments (see Figure 4.6 (right)). Effectively, post-intervention variables have a different distribution $P_{do(A=a)}(X,A,Y)$, and samples (or observations) from this distribution can be drawn only if the intervention $do(A = a)$ can be physically carried out (e.g. in randomized experiments). In many real life situations this is not feasible, and *do*-calculus provide a mathematical machinery that allows, in some specific cases, to compute quantities related to the (non observable) post-intervention distribution $P_{do(A=a)}(X,A,Y)$ employing only quantities related to the (observable) distribution $P(X,A,Y)$, given the graph — a concept called *identifiability* [143, 145, 144, 148, 80].

The notation $\hat{Y}_{A\leftarrow b}$, on the other hand, stands for the three steps of counterfactual calculus, *abduction, action, prediction* [144, 145]:

- *abduction* is where you account for observed values and compute the distribution of $U \mid \{A = a, X = x\}$;
- *action* corresponds to implementing the intervention $do(A = b)$;
- *prediction* consists in using the new causal structure and the exogenous conditional distribution $\mathrm{P}(U \mid A = a, X = x)$ to compute the posterior of $\hat{Y}$.

As in the observational setting, causality-based criteria have a group and an individual version: equations (4.20), (4.21) correspond to the individual form, but nothing prevents from taking the version unconditional on $X$, i.e. holding on average on all the individuals, or even to condition on only some subset of $X$, as in CDP.

### 4.5.3 Group vs. Individual Fairness in Causality-Based Setting

We call *Expectation Intervention Fairness* the condition:

$$P(\hat{Y} = 1 \mid do(A = a)) = P(\hat{Y} = 1 \mid do(A = b)) \quad \forall a, b \in \mathscr{A}, \tag{4.22}$$

i.e. requiring that, on average, an individual taken at random from the whole population and forced to be woman ($do(A = a)$) should have the same chance of being accepted as a random individual forced to be man. Analogously, we can define *Expectation Counterfactual*

*Fairness (ECFF)* as:

$$P(\hat{Y}_{A\leftarrow a} = 1 \mid A = a) = P(\hat{Y}_{A\leftarrow b} = 1 \mid A = a) \quad \forall a, b \in \mathscr{A}, \tag{4.23}$$

which states that, on average, the acceptance rate for a random woman ($A = a$) should be the same as the one given to a random woman forced to be a man. Similar definitions can be given conditioning on partial information $R$.

Summarizing, we can visualize CFF as the following process: the conditioning on some values $\{A = a, R = r\}$ represents the group of people that we take into account (a single individual for $R = X$), and any consideration we do on them must hold on average over that group; given that group, we force a flipping of the sensitive attribute from $A = a$ to $A = b$ (and here we are going in a new, non-observable, distribution), and this will trigger a cascade of causal consequences on the other features $X$ (namely, on the descendants of $A$ in the causal graph). Then, we compare the model outcomes averaged on the observed group and on the counterfactual group.

Intervention Fairness is slightly different: you take again all the individuals compatible with the conditioning – which this time is $R = r$ without fixing the sensitive $A$ – then force them first to have $A = a$, and then to have $A = b$ (with all the causal consequences of these interventions), and require them to have, on average, the same acceptance rate.

This may resonate with the notion of FTU: when flipping $A$ for an observation you want the outcome of the model not to change. Indeed *Fairness Through Unawareness is a causality-based notion of fairness*, at least in its formulation of not explicitly using $A$. The very concept of flipping $A$ is nothing but an intervention. Notice, incidentally, that the flipping of $A$ *without any impact on other variables* corresponds to assuming a causal graph where $A$ has no descendant, or, better, corresponds to assuming that all the changes caused by the flipping of $A$ on other variables are legitimate, i.e. considered fair. Moreover, the fact that FTU is difficult to be measured without having access to the model is due to its nature of being a non-observational notion, but it requires fictitious data to be assessed – a dataset with $A$ flipped, i.e. a *dataset not sampled from the real distribution of* $(A, X, \hat{Y})$. This is similar to the way in which CFF can be assessed: compare the predictions over a dataset with the predictions on the same dataset but with $A$ flipped *and* with all the changes caused by this flipping (knowledge of the underlying SCM is thus needed). This, in turn, reveals the subtle difference among CDP in the form $\hat{Y} \perp\!\!\!\perp A \mid X$ and FTU: CDP is an observational notion, i.e. can be measured, in principle, by having access to (realizations from) the distribution of $(\hat{Y}, A, X)$, while FTU does not. We say "in principle" since it's very likely that in real-world datasets there will be very few observations corresponding to $X = x$ (typically one), thus resulting in a poor estimation of the distribution of $\hat{Y} \mid X$.

Causality-based notions are richer than the observational ones, and permit a further possibility: to select which causal paths from $A$ to $\hat{Y}$ are considered legitimate and which, instead, we want to forbid. In the job listing example, we may consider that the type of degree of the applicants is fundamental for the job position, and thus crucial to making decisions. But it could well be that the type of degree is correlated, and even "caused" by the gender of the applicants: women and men may have different attitudes towards choosing the preferred degree. In this situation, if we require CFF, we would compare a man with some degree to his "parallel self" in the counterfactual world where he is a woman. In that world, however, it's very likely that her degree is going to be different. Thus, requiring CFF means somehow prevents the decision-maker to employ the degree type for the assessment. Taking into account situations like this one is not very difficult. Supplementary Material of [111] and [44] introduce the following definition, known as path-specific CounterFactual Fairness (pCFF):

$$\begin{aligned} P(\hat{Y}_{A \leftarrow a, X_F \leftarrow x_1} = 1 \mid A = a, X_F = x_1, X_F^c = x_2) = \\ P(\hat{Y}_{A \leftarrow b, X_F \leftarrow x_1} = 1 \mid A = a, X_F = x_1, X_F^c = x_2), \quad \forall a, b \in \mathscr{A}, \forall x_1, x_2. \end{aligned} \tag{4.24}$$

where $X_F$ correspond to the variables descendants of $A$ that we consider fair mediators to make decisions (e.g. the degree type), and $X_F^c$ is its complement with respect to $X$, namely $X = (X_F, X_F^c)$. In words, take a man ($A = a$) with a specific degree $x_1$ and other features $x_2$, and force him to be a woman, letting all the causal consequences of this flipping to happen *with the exception of the degree, kept fixed at* $x_1$, then compare their outcomes.

Incidentally, notice that if we allow $X_F$ to contain all the descendants of $A$, then we end up with a notion that we may call *direct Counterfactual Fairness (dCFF)*, namely the only path that we are concerned of is the direct causal path from gender $A$ to the decision $\hat{Y}$. In this case, we don't let any causal consequences of the gender flipping to happen when assessing fairness. This is, again, strictly connected to FTU.

Figure 4.7 is a schematic and qualitative visualization of many of the points discussed above: the dimension of group vs. individual is "controlled" by how much information we condition on, while the dimension of the causal impact of $A$ on the decision is controlled by the fraction of paths causally connecting $A$ to $\hat{Y}$ that are considered fair. Figure 4.7 is meant to be the causal analogue of Figure 4.4. Of course there are many possible cases that, for simplicity, we have omitted from Figure 4.7, namely all the Intervention notions, and the path-specific notions valid on average in broader groups. For instance, one could think of a measure taking into account only the $A$ flip, without causal consequences, and valid on average over $X$ as well (i.e. a group notion), namely $P(\hat{Y}_{A \leftarrow a, X \leftarrow X} = 1 \mid A = a) = P(\hat{Y}_{A \leftarrow b, X \leftarrow X} = 1 \mid A = a), \quad \forall a, b \in \mathscr{A}$.

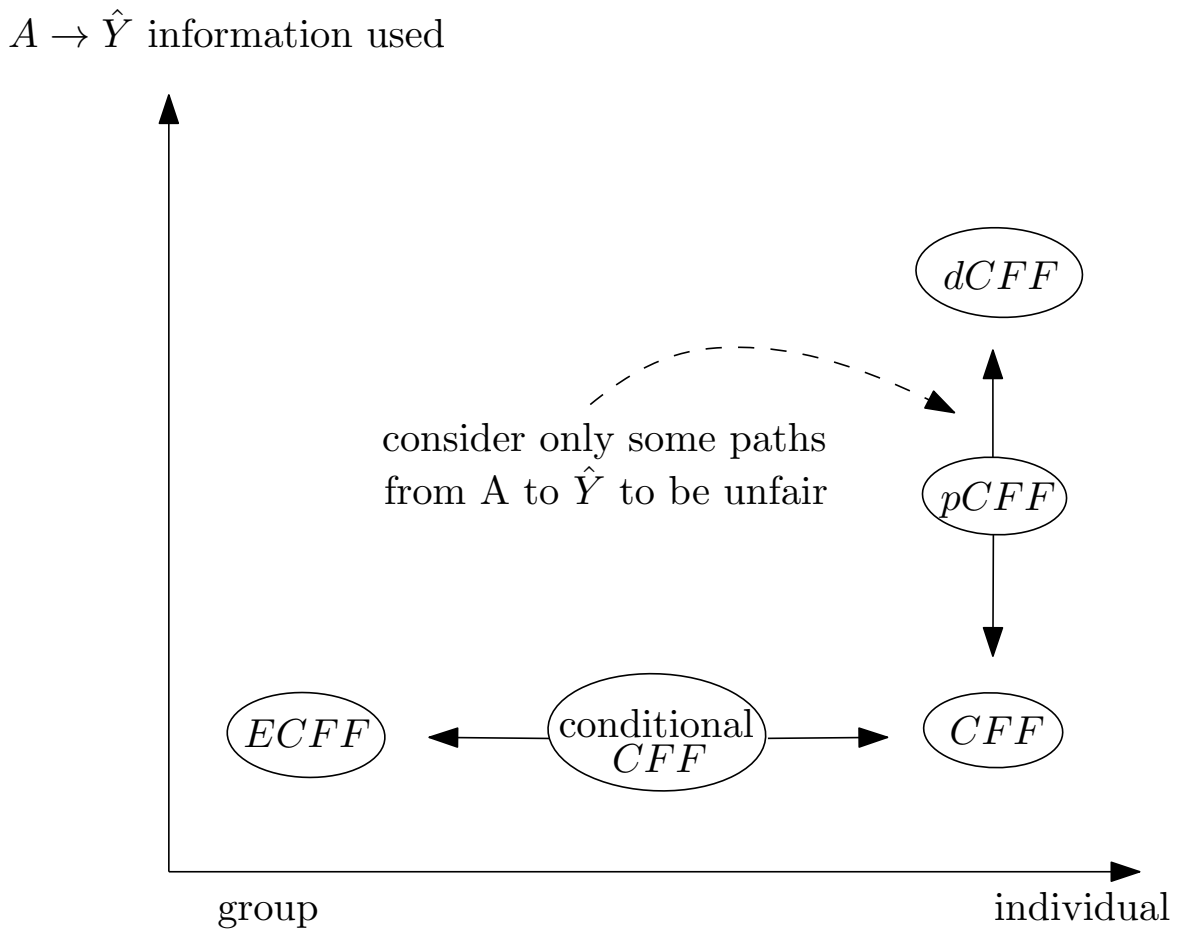

Figure 4.7 Landscape of causality-based fairness criteria.

If we focus on CFF, it can be reached by training a model on the space $(\Sigma, U_A, Y)$ where $\Sigma$ are the *non-descendants* of $A$, i.e. variables not caused by $A$, *directly or indirectly*; $U_A$ denotes the information inside descendants of $A$ that is not attributable to $A$. A decision system $\hat{Y} = f(\Sigma, U)$ is counterfactually fair by design (4.21). This closely reflects what we discussed in Section 4.2 about fair representations: train a model on a new space "cleaned" from all $A$ information. Indeed, in that case, we searched for statistical independence of $\hat{Y}$ on $A$ while here we look for "causal independence" of $\hat{Y}$ on $A$. In this specific sense, counterfactual fairness can be reached via a form of *preprocessing* of the dataset, and is very similar, in spirit, to the concept of fair representation.

We may then wonder why the fair representation approach, reaching DP, is a notion of group rather than individual fairness, while CFF is considered an individual notion: the point is simply that *we employed two different concepts of individual fairness in the observational and in the counterfactual setting*. Namely, in the observational setting we expressed disparate treatment as the event in which two individuals with same $X$ but different $A$ are given different outcomes. In general, fair representation is not guaranteed to prevent such scenario. In the counterfactual context, instead, we consider disparate treatment when an individual and her "counterfactual self" with $A$ flipped are given different outcomes. CFF is designed precisely to prevent this. Notice that this concept, translated in the observational setting, is analogous to requiring a similar treatment for two individuals similar in $\mathscr{Z}$, not in $\mathscr{X}$. Indeed, if you take an individual with some $\{A = a, X = x\}$ and flip her gender *while taking into account the interdependence of* $X$ *and* $A$, she will be transformed in the same point $Z$.

This is an example of the fact that the concept of individual fairness is strongly dependent on the concept of similarity that one decides to consider.

Summarising, CFF and fair representation are very similar in the way in which they deal with disparities, namely they both try to remove all (causal) information of $A$ from the feature space, and then use the "cleaned" space to make decisions. In this respect, CFF can be seen as both a "causal analogue" of DP-compliant methods, in the sense that all the ways in which $A$ may impact the decision are forbidden, and an individually fair notion in the sense that it imposes a condition on an individual basis.

## 4.6 Discussion on the Fairness Metrics Landscape in Machine Learning

The notion of fairness in decision-making has many nuances, that have been reflected in the high number of proposed mathematical and statistical definitions. Notice, moreover, that this aspect is not limited to ML or artificial intelligence: the problem of how to define and assess bias discrimination in decision-making processes is present largely independently of "who" is making the decision. The growing attention to this issue in the domain of automated data-driven decisions can be imputed to the fact that these processes can potentially amplify biases *at scale*, and could possibly do that without *human oversight*. It is well known that AI systems usually compute their outcomes leveraging very complicated relationships among input features, making not only the users but often also the developer blind with respect to the reasons underlying those outcomes. This issue, which has triggered in recent years the flourishing stream of research of Explainable AI [79, 32], is intertwined to that of bias detection and fairness in ML, since the problem of discrimination becomes even subtler when carried out by a "black-box" algorithm whose rationale is obscure or ambiguous.

Even if a lot of work has been done in this respect, still there is confusion on the interplay among different fairness notions and metrics to assess them. In this chapter, we tried to highlight some important aspects of the relationships between fairness metrics, in particular with respect to the clash "individual vs. group" and "observational vs. causality-based" and we have argued that we are facing rather a landscape of intertwined possibilities than clear-cut distinctions.

The next part of this thesis will be dedicated to exploring the process of mitigation once the specific fairness metric has been determined for implementation.

As we argue in this work, the various notions of fairness and the corresponding metrics are intertwined, with individual and group fairness being actually on the same "dimension", depending on the the amount of sensitive information contained in non-sensitive features we are willing to accept in specific cases. Moreover, we have highlighted the connections between observational and causality-based criteria, which also have group and individual aspects.

# Part II

# Mitigating Bias

# 5

# Fairness Mitigation

Fairness mitigation refers to the process of addressing specific aspects of the ML pipeline in order to remove the effect of unfair bias [198]. Given a specific fairness metric, different strategies can be implemented in order to fit a model by pursuing both algorithmic performance and an optimal value of the chosen metric.

Numerous efforts have been made to achieve fairness and mitigate bias in artificial intelligence across various domains [128]. Typically, methods aiming to address algorithmic biases can be categorized into three main approachess [128, 142]: *pre-processing*, *in-processing*, and *post-processing*.This distinction arises from the point in the development pipeline of an AI-based decision system where bias mitigation is implemented (refer to Figure 5.1).

## 5.1 Pre-Processing Mitigation

Pre-processing techniques focus on transforming the data to eliminate underlying discrimination [53]. By modifying the training data, these approaches aim to address issues related to biased, discriminatory, or imbalanced distributions of sensitive or protected variables. The goal is to alter sample distributions of protected variables or apply specific data transformations to remove discrimination from the training data [102]. Pre-processing is valued for its flexibility in the data science pipeline, as it allows modeling techniques to be chosen independently.

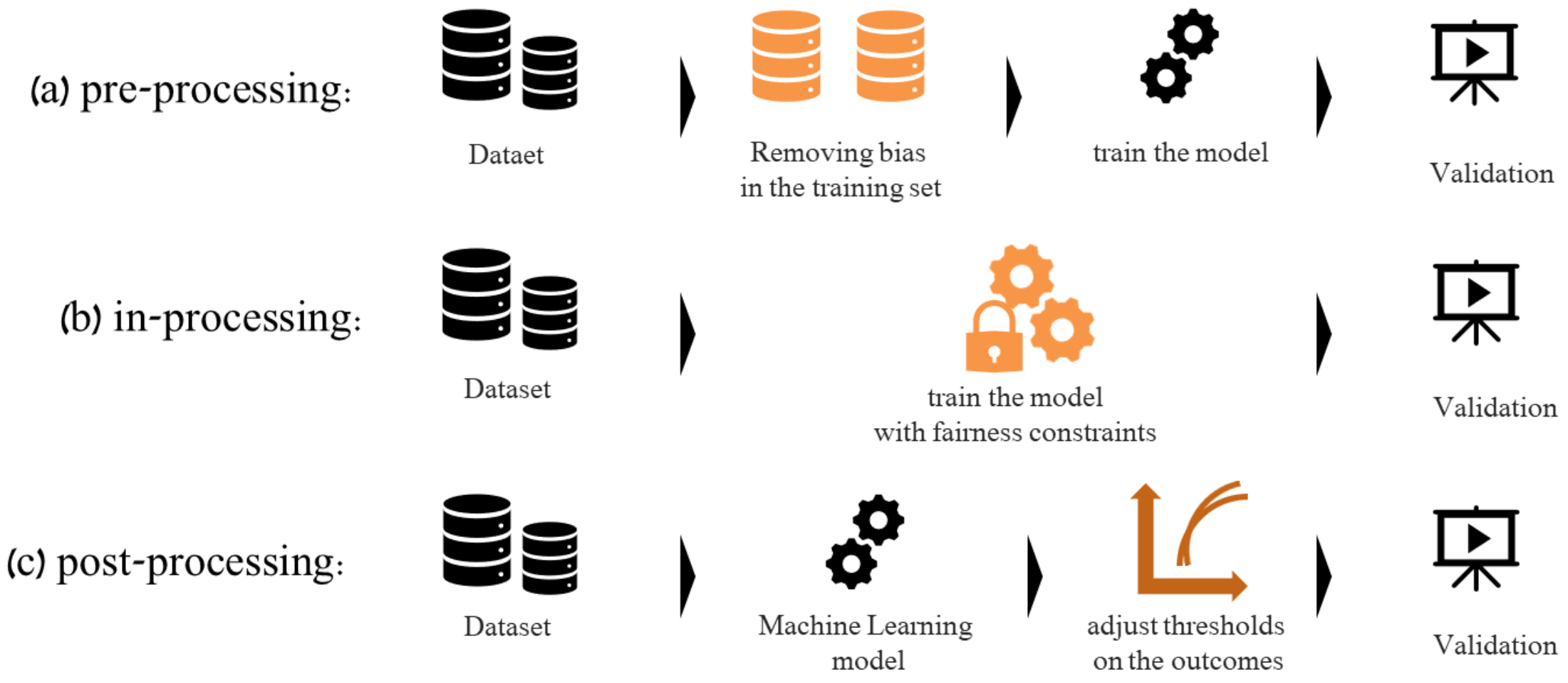

Figure 5.1 Schema illustrating the three distinct fairness mitigation approaches.

For instance, Kamiran and Calders [101] propose three strategies centered around removing discrimination from the training dataset. In these methods, a classifier is trained on the cleaned dataset. The rationale for these approaches is that, since the classifier is trained on discrimination-free data, it is likely that its predictions will be (more) discrimination-free as well. The first method, known as *massaging the data*, involves altering the target variable labels to eliminate discrimination from the training data. The second method is less invasive as it avoids modifying the class labels; instead, it assigns weights to the data instances to create a discrimination-free dataset. This technique is referred to as *reweighing*. As reweighing necessitates the learner's ability to handle weighted data instances, an alternative solution is proposed that eliminates this requirement. In this approach, called *sampling*, the dataset is resampled to eliminate discrimination. These techniques will be employed on real-world credit lending data in Section 5.6.2.

## 5.2 In-Processing Mitigation

In-processing techniques involve modifying state-of-the-art learning algorithms during the model training process to eliminate specific bias [50]. This method is highly tailored to the specific underlying model and is, therefore, difficult to generalise. By incorporating changes into the objective function or applying constraints, in-processing adjusts the learning procedure for machine learning models [185]. The main challenge tackled by in-processing approaches is the indirect discrimination introduced by dominant features, distributional effects, or the trade-off between accuracy and fairness. These techniques often integrate fair-ness metrics into the model optimization functions to achieve a balance between performance and fairness.

A representative work [189] has investigated this idea and presented a general formulation of using adversarial learning to mitigate unwanted bias. This article shows that the framework can be used to achieve various fairness conditions by adjusting the inputs of the adversarial network based on the fairness definitions. In Section 5.6.3, we will present an example of the implementation of this technique along with additional detailed information. For a comprehensive survey of in-processing modeling techniques, we recommend referring to [177]. The paper categorizes these approaches into *explicit* and *implicit* methods based on where fairness is integrated into the model. Explicit fairness mitigation can be achieved by explicitly adding fairness regularizers or constraints to the objective functions of machine learning models, while implicit methods focus on refining latent representation learning. The authors also provide an overview of how each category of methods can be applied to mitigate group-level bias, individual-level bias, and other mixed scenarios.

## 5.3 Post-Processing Mitigation

Post-processing occurs after model training, where a holdout set is accessed that was not used during the training process [53]. When the algorithm can only treat the learned model as a black box without the ability to modify training data or learning algorithms, post-processing is the applicable approach [27, 81]. Post-processing recognizes that the output of an ML model may be unfair to certain protected variables or subgroups within them. As a result, post-processing techniques apply transformations to the model's output to improve prediction fairness. This approach is highly flexible since it requires only access to predictions and sensitive attribute information, not the actual algorithms or ML models. Thus, it is well-suited for black-box scenarios where the entire ML pipeline is not exposed.

The algorithmic fairness community has provided formal proofs and implementations for optimal *post-processing* solutions satisfying existing notions of group fairness, as described in the last column of Table 5.2 [81, 49, 16]. These include finding a so-called decision rule, which transforms the ML prediction into a final decision. Hardt et al. [81] and Corbett-Davies et al. [49] showed that among rules satisfying *DP*, *CSP*, *TPR parity*, or *FPR parity*, the optimum always takes the form of group-specific thresholds.[1] Furthermore, Baumann et al.

[1] For the fairness criterion *CSP*, the group-specific thresholds additionally depend on the "legitimate" attributes $L$.

[16] showed that among rules satisfying *PPV parity* or *FOR parity*, the optimum always takes the form of group-specific upper- or lower-bound thresholds. This means that, in certain situations, it can be optimal to assign a positive decision to the 'worst' individuals of one group (i.e. those with the lowest predicted scores) and omit the most promising ones. This can happen if, for a utility-maximising decision-maker, it is overall better to 'sacrifice' the 'best' individuals of the smaller group in favour of 'keeping' the 'best' individuals from the larger group. For the fairness notions that combine two parity constraints (i.e. *separation* and *sufficiency*), some randomisation is needed to satisfy both constraints at the same time. Among rules satisfying *separation* or *sufficiency*, the optimal decision rules always take the form of randomised group-specific upper- or lower-bound thresholds (see Hardt et al. [81] for *separation* and Baumann et al. [16] for *sufficiency*).

We will apply post-processing mitigation techniques in both sections 5.6.4 and 5.7.

## 5.4 Causal Methods

Causal methods-based approaches acknowledge that the data used for training machine learning models frequently mirror some underlying discriminatory patterns [41]. Their primary goal is to unveil causal relationships within the data and reveal dependencies between sensitive and non-sensitive variables [107, 124]. Therefore, causal methods are particularly adept at identifying proxies of sensitive variables.

The majority of the existing work on these solutions has been focused on individual fairness and can be loosely considered as pre-processing methods since they consist in training a ML model on a transformed dataset. For example, [111] proposes a method to produce counterfactually fair outcomes, where all protected information causally impacting the decision is removed from the dataset (i.e. an individual is given the same decision that the decision-maker would have been given in the counterfactual world where sensitive features are different). [44] extends this counterfactual fairness, accounting for the fact that not all the causal impact of the sensitive information on the decision is in general unfair, thus mitigating only with respect to variables that are part of unfair causal paths. Despite the substantial implementation of causal methods in the pre-processing stage, it is inappropriate to categorize them solely within this category. The literature indeed recognizes their utilization for causal-based mitigation during both the in-processing (see e.g. [78] and post-processing (see e.g. [134]) stages. In Section 5.6.5, we will showcase a real application of implementing causal counterfactual fairness in the context of credit lending.

## 5.5 Discussion on Fairness Mitigation Approaches

Pre- and post-processing techniques offer implementation advantages as they do not directly alter the underlying ML method [41]. This allows leveraging (open source) ML libraries for model training without significant alterations. However, these approaches lack direct control over the optimization function of the ML model itself. This limitation could lead to legal implications due to modifications in the data and/or model output [14]. Additionally, models may become less interpretable [120], which can be in conflict with current data protection legislation's requirements for explainability [170, 114] and the forthcoming transparency requirements outlined in the AI Act [169] (see Section 7.1.1 for more comprehensive information regarding the regulations influencing the utilization of AI.).

On the other hand, in-processing approaches enable the optimization of fairness notions during model training. This involves making changes to the optimization function, allowing for a more targeted fairness optimization. However, it is essential that the optimization function is accessible, replaceable, and/or modifiable, which might not always be feasible or straightforward in some cases. In certain situations, the complexity of the model or the constraints of the ML framework could hinder the implementation of in-processing approaches.

Each approach comes with its unique trade-offs and considerations, and the choice of which approach to adopt should be carefully evaluated based on the specific requirements and constraints of the AI system and the ethical considerations surrounding fairness and transparency.

Various companies and public institutions have made an effort to encompass fairness metrics and mitigation techniques through specific software tools, and toolkits, such as IBM AI Fairness 360 [18], Google What-If Tool [180], and Aequitas [163]. Despite the progress made, these solutions are usually context-agnostic. However, each industry and process has its particularities and it is needed to research and formalize ad-hoc solutions from institutions. The next sections describe the results of our implementation of the most common mitigation techniques in the banking sector, in particular for a credit-lending use case.

## 5.6 Experiment on Fairness Mitigation using Real-World Data on Credit Lending using `BeFair`

As pointed out in Chapter 4, there is no such thing as the fairness metric and thus there is neither a single nor definitive way to mitigate bias in all situations.

To assist data scientists at Intesa Sanpaolo in their efforts to achieve fairness mitigation, we have developed a comprehensive toolkit called `BeFair`. This toolkit offers a range of functionalities designed to implement fairness in real-world use cases. Its capabilities encompass bias detection within input datasets and model outcomes, bias mitigation utilizing various strategies from existing literature, performance evaluation using pertinent metrics, performance/fairness trade-off comparison across different strategies, and interpretation of feature relationships through causal graphs.

In the subsequent part of this section, we will focus on the credit lending use case. Following that, we will provide a brief overview of the mitigation strategies integrated within the `BeFair` toolkit, as well as an insight into the outcomes summarised in Table 5.1 obtained from applying these strategies.

### 5.6.1 Data Assessment on Credit Lending

Decisions concerning credit lending are indeed highly susceptible to unfairness. To evaluate each application, financial institutions often request a certain amount of personal information whose examination might potentially lead to, even unintentionally, discrimination. The dataset we used for these experiments comes from an anonymised portion of past loan granting applications, whose actual final outcome had no dependence in any way on an ML model.

The real dataset consists of 250,000 entries, collected in 2018, representing the granting of personal loans. The dataset contains over 50 financial attributes, such as "income" and "bank rating", and two binary sensitive attributes: *gender* and *citizenship*. The decision to grant a loan, which is used as a target variable, is made by human loan experts, supported by a careful process of controls. Throughout the following sections, we perform the assessment and apply the mitigation strategies over the feature *citizenship* as a mere example, to be able to point out that without intention a standard model can inject discrimination in the predictions even if no significant bias exists in input data.

### 5.6.2 Pre-Processing Implementation

We implement three different pre-processing techniques: suppression, massaging and sampling. In *Suppression* [101] the transformed dataset is derived by removing both the sensitive variable and the features with highest correlation with it. We removed all variables with correlation higher than 15%. *Massaging* [100, 101] consists in relabelling the target label of some observations in order to reach Demographic Parity for the "massaged" target. To choose which observations must be relabelled, an auxiliary classifier is trained (the literature proposes to use a Bayesian classifier but we have obtained better results using a Random Forest (Random Forest (RF)). Finally, *Sampling* [101] simply consists in over - or under - sampling observations in order to reach Demographic Parity.

Once the dataset has been transformed, we apply a standard Random Forest to get the mitigated outcomes.

### 5.6.3 In-Processing Implementation

We implement two different algorithms: Adversarial Debiasing and Reductions.

*Adversarial Debiasing* [189] is based on the simultaneous training of two competing classifiers (corresponding to a Generative Adversarial Network).

In the first one, the predictor $P$ tries to accomplish the task of predicting the target variable $Y$ given the input variables $X$ by modifying its weights $W$ to minimise some loss function $L_P(\hat{Y},Y)$. The second one, the adversary $A$, tries to accomplish the task of predicting the sensitive variable, given $\hat{Y}$ by modifying its weights $U$ to minimise some loss function $L_A(\hat{Z},Z)$ and consequently backpropagates the error through the predictor $P$.

If the Adversary model is trying to estimate the sensitive attribute given only $\hat{Y}$ the result will satisfy Demographic Parity. Rather, if it is trying to estimate the sensitive attribute given $\hat{Y}$ and the true label $Y$, the result will satisfy Equality of Odds.

We rely on the Python module `AIF360` [19] to apply Adversarial Debiasing in order to either impose Demographic Parity (Adversarial Debiasing imposing DP (AdvDP)), Equality of Odds (Adversarial Debiasing imposing EOdd (AdvEOdd)) or Conditional Demographic Parity (Adversarial Debiasing imposing CDP (AdvCDP)). The latest is achieved by training a different Adversarial model for each subgroup of the variable we condition on. In our use case, we condition on three different levels of credit risk: high/medium/low.

*Reductions* approach [2] is based on the idea of finding a classifier that minimises the classification error subject to a specific fairness constraint, effectively reducing the problem to a sequence of cost-sensitive classification problems. More specifically, it takes an arbitrary ML model and trains it multiple times updating, at each iteration, the weights to be assigned to each observation in order to fulfill the chosen fairness constraint. It is actually a hybrid between an in-processing method, since it works by imposing fairness during training, and a post-processing method, since it can be applied to any ML models, treating them as black-boxes.

We rely on `Fairlearn` Python module [27] for the actual implementation of the Reductions approach.

We train different models (starting from a logistic regression) using the demographic parity metric. In particular, first, we train 50 models using the GridSearch algorithm (ReductionsGS) and select the best one. Second, we train 2 models using the ExponentiatedGradient (ReductionsEG) with 0.001 and 0.01 values for the constraints [2].

### 5.6.4 Post-Processing Implementation

As mentioned in the previous section, these are a set of techniques that basically consists in computing the mitigated decisions as a function $\hat{Y} = f(R,A)$, where $R$ is the outcome of any given (in general biased) classifier (in our case a Random Forest) and $A$ the sensitive attribute, such that it satisfies a desired fairness metric.

We implement a simple algorithm to impose Demographic Parity (ThreshDP) and one to impose Equality of Opportunity (ThreshEopp) and we rely on `Fairlearn` Python module [27] to compute post-processing mitigation enforcing Equality of Odds (ThreshEO)[2]. Conditional Demographic Parity (ThreshCDP) is enforced by choosing a threshold not only group-dependent but also depending on the level of the variable we are conditioning on, which is chosen to be a 3-level credit risk, as for the adversarial debiasing case.

### 5.6.5 Counterfactual Fairness Implementation Through Causality

`BeFair` includes the implementation and particularization of counterfactual fairness [111] based on the domain knowledge of our specific use case.

[2]Demographic Parity and Equality of Opportunity can be easily enforced by computing group-wise thresholds; while, in general, Equality of Odds requires some form of randomization, since it may not be possible for each group to select a single threshold classifier reaching the same true positive rate and false positive rate. See [81] and `Fairlean` documentation [27] for more details.

Counterfactual fairness requires establishing a causal graph that represents how the variables influence each other (including input features and outcome). This is usually represented by a Directed Acyclic Graph (DAG) in which each variable is represented by a node and arrows represent causal relationships.

We have defined our specific causal graph using multiple causal discovery algorithms. In particular, we employed the Python `Causal Discovery ToolBox` [99], including different graph modeling algorithms on observational data (i.e. SAM, PC) and the NOTEARS algorithm [195] included in the Python library `CausalNex`. Once the results were obtained and integrated, a manual revision has been performed to verify the validity of each relation detected (represented by arrows). A group of domain experts participated in the validation process to determine the final causal graph that is included in `BeFair` (Figure 5.2).

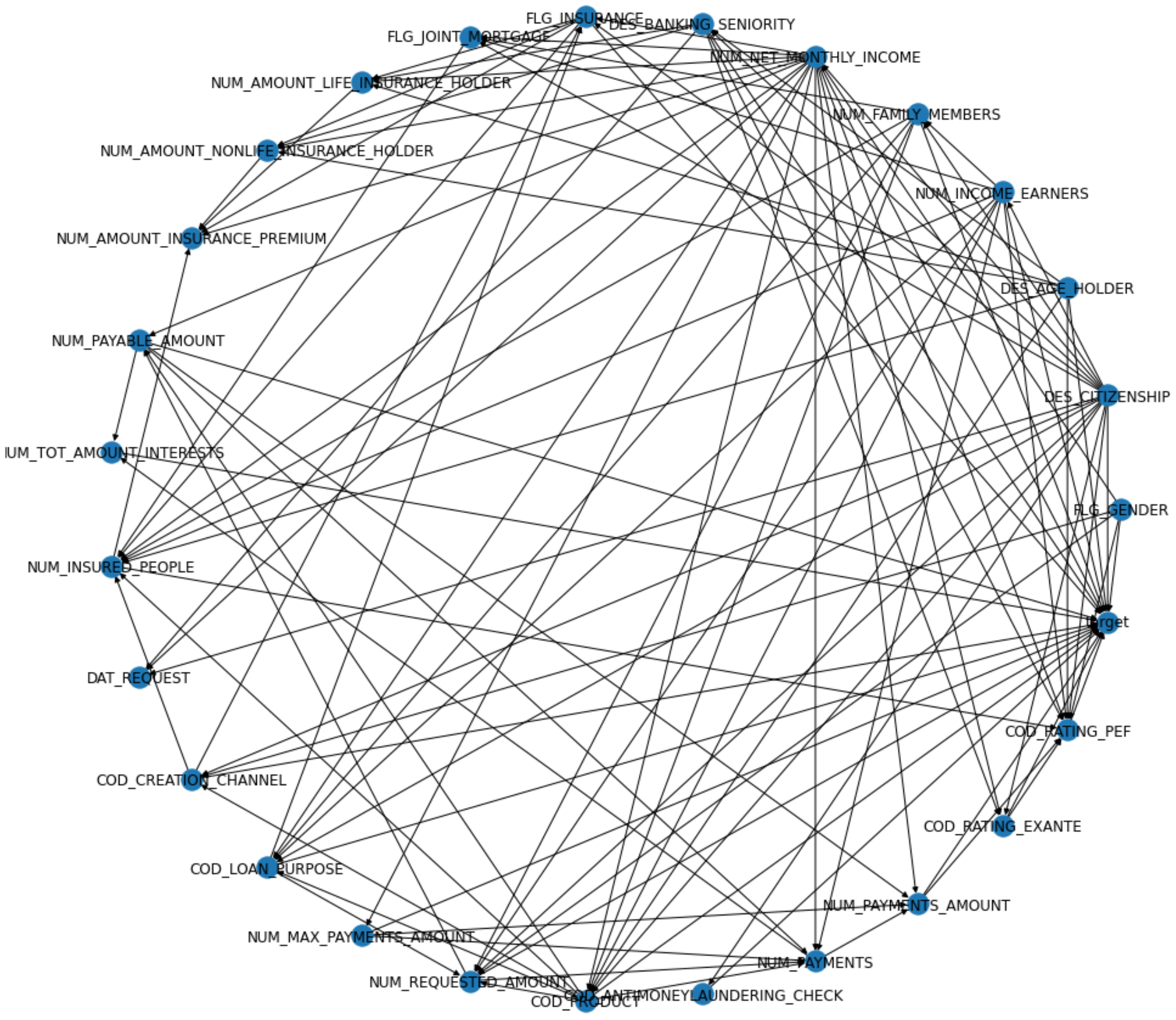

Figure 5.2 Causal graph for the credit lending use case.

Table 5.1 Fairness and performance assessment of mitigation strategies implemented in BeFair, as discussed in Section 5.6 with respect to different metrics. Each panel is devoted to a particular family of mitigation techniques. Fairness metrics are expressed in terms of difference over sensitive groups, thus the lower in absolute value the better. Bold (underline) highlights the best (worst) value per column. Results are discussed in Section 5.6.7.

| | | **fairness** | | | | **performance** | | |
|---|---|---|---|---|---|---|---|---|
| **family** | **type** | **DP** | **EO** | **EOpp** | **PP** | **AUROC** | **Accuracy** | **F1** |
| no mitigation | Logistic | 0.324 | 0.272 | 0.272 | **0.032** | 0.817 | 0.761 | 0.823 |
| | Random forest | 0.221 | 0.202 | -0.104 | 0.068 | **0.838** | 0.804 | 0.875 |
| | Neural network | 0.219 | 0.198 | 0.104 | 0.072 | 0.830 | 0.811 | **0.876** |
| pre-process | FTU | 0.164 | 0.124 | 0.058 | 0.095 | **0.838** | 0.812 | **0.876** |
| | Suppression | 0.099 | -0.053 | 0.065 | 0.152 | 0.753 | 0.748 | 0.840 |
| | Massaging | -0.004 | 0.062 | 0.062 | 0.163 | 0.818 | **0.868** | 0.803 |
| | Sampling | 0.080 | 0.012 | 0.012 | 0.115 | 0.835 | 0.791 | 0.851 |
| | CFF | 0.218 | 0.192 | 0.104 | 0.070 | 0.832 | 0.810 | 0.874 |
| in-process | AdvDP | -0.034 | 0.073 | 0.063 | 0.176 | 0.823 | 0.802 | 0.869 |
| | AdvEO | 0.102 | 0.029 | -0.010 | 0.148 | 0.819 | 0.805 | 0.871 |
| | AdvCDP | 0.147 | 0.101 | -0.050 | 0.112 | 0.830 | 0.807 | 0.872 |
| | ReductionsGS | 0.012 | 0.077 | 0.049 | 0.159 | 0.812 | 0.794 | 0.864 |
| | ReductionsEG | 0.007 | 0.084 | 0.051 | 0.161 | – | 0.794 | 0.864 |
| post-process | ThreshDP | **0.003** | 0.099 | 0.056 | 0.164 | – | 0.805 | 0.872 |
| | ThreshEO | 0.082 | **0.006** | 0.006 | 0.138 | – | 0.812 | 0.873 |
| | ThreshEOpp | 0.100 | 0.048 | **0.005** | 0.119 | – | 0.809 | 0.874 |
| | ThreshCDP | 0.186 | 0.159 | 0.072 | 0.083 | – | 0.810 | 0.875 |

Given the defined causal graph, we have developed a CFF model based on level 3 described in [111]: the main idea is to model the data using an additive error model with deterministic residuals (error terms of the model) and then fit the CFF model on non-descendants of the sensitive attribute(s) and the residuals. The process to obtain counterfactual outcomes entails changing the value of the sensitive attribute and propagating the effect of the change throughout the causal graph to obtain the new values of the new input features. The model trained on the non-descendants of the sensitive feature(s) and the residuals are then counterfactually fair by design, meaning that an individual and its counterfactual version are given the same outcome.

### 5.6.6 Model Comparison in BeFair

Sections 5.6.2, 5.6.3, and 5.6.4 provide a comprehensive overview of various mitigation techniques and metrics, demonstrating the extensive range of possibilities. In light of this, we propose two approaches that aim to synthesize the trade-offs among these techniques and metrics. These approaches are designed to assist in identifying the most suitable model based on the specific domain's requirements and constraints.

- *Trade-off fairness-performance.* This indicator is inspired by $F_\beta$-score used in ML, whose value for $\beta = 1$ results in the harmonic mean of precision and recall:

$$(1+\beta^2)\frac{(1-|\phi|)\times\pi}{\beta^2\times(1-|\phi|)+\pi}; \tag{5.1}$$

  where $\pi$ and $\phi$ are the preferred performance and fairness metrics, respectively and $\beta$ is the weight associated with the performance metric.
- *Constrained performance.* Once chosen an upper bound $\Phi$ for a desired fairness metric $\phi$, this indicator corresponds to the highest possible performance given that fairness constraint

$$\max_{\phi\leq\Phi}\pi. \tag{5.2}$$

The optimal choice is then given by the model maximizing the selected indicator.

In our `BeFair` implementation we have considered the following parameters to compare models, although other metrics could be included easily. For the performance metric $\pi$ we have considered accuracy, precision, recall, and F1. For the fairness metric $\phi$ we have included Demographic Parity, Equal Opportunity, Predictive Parity, and Equality of Odds.

`BeFair` allows users to configure the model comparison. Figure 5.3 depicts an example of a graph generated by `BeFair` showing the trade-off among fairness ($x$-axis) and performance ($y$-axis). In particular, it shows the comparison using DP and F1 metrics for all the mitigation strategies implemented (blue dots in the graph) and models without mitigation (orange dots). The best strategy (*y_prepro_massaging*) is identified, in this example, using the constrained performance approach with $\Phi = 0.05$, reaching an F1 = 0.875.

### 5.6.7 Discussion on the Result

Table 5.1 summarises the results of the mitigation strategies introduced in Section 5.6 relative to the sensitive attribute of *citizenship* for the credit lending use case.

Fairness quantification is done via the most used statistical fairness dimensions, namely $Y$-independent (DP), recall based (EO, EOpp) and precision based (PP); while performance is monitored via usual metrics such as Area Under the ROC (AUROC), accuracy and the F1 score (i.e. harmonic mean of precision and recall).

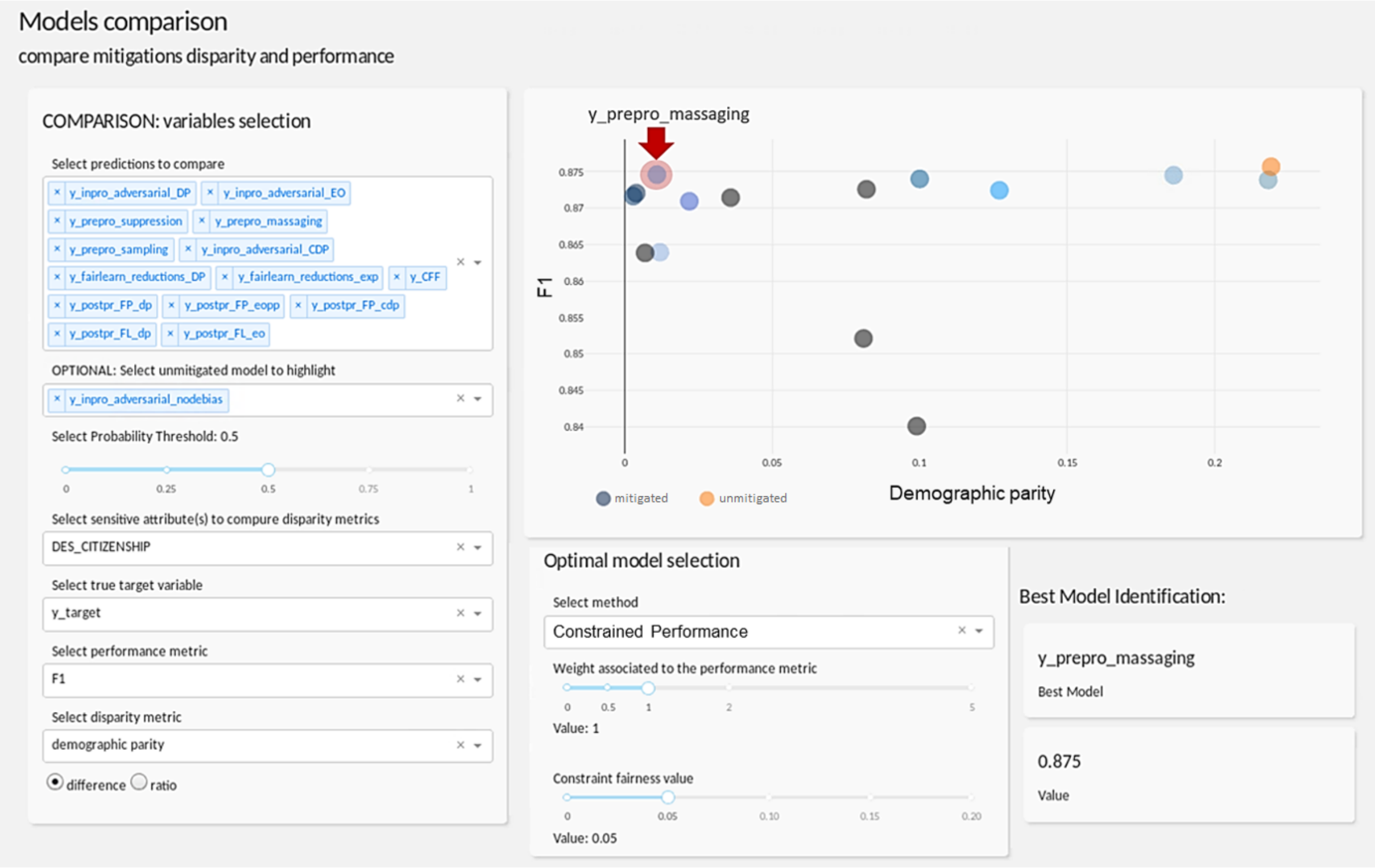

Figure 5.3 Models comparison: snapshot of `BeFair`. Several mitigation strategies can be selected, together with the desired protected attribute and the chosen performance and fairness metrics to be used (left panel). The right panel is devoted to the performance-fairness plane (top) and to the selection of the optimal model (bottom) (see 5.6.6).

First of all, it is clear from the top panel that naively applying a common ML model to the entire dataset results in the amplification of bias with respect to almost all the possible fairness dimensions. This simple fact is *per se* a sufficient reason to promote and foster the need for attention on fairness issues in ML models, in particular in its banking sector applications.

The intuitive and simplest pre-processing methodology of suppression seems not to be able to reach the same level of mitigation as the other strategies, moreover paying a higher price in terms of predictive performance. Thus, bias mitigation while preserving performance is not as simple as removing a bunch of variables from the training dataset. Better results come from massaging the dataset.

Fairness Through Unawareness [3], as expected, is not able to mitigate bias with respect to group metrics, since it does not take into account the sensitive information still present in the dataset via correlations to other variables.

[3]The underlying classifier is a plain Random Forest.

The in-processing techniques are able to reduce DP and EO maintaining the same performance as the unmitigated models. The downside of using these methods is that they consist of training models specifically designed to meet some fairness constraint and they cannot be generalized to arbitrary models.

Post-processing techniques, selecting group-wise thresholds over a trained model, have binary 0/1 outcomes, thus for them, it is not possible to compute the Area under the ROC. However, they seem to be overall pretty good in all fairness dimensions without loosing much in performance. On the other hand, they can be criticised for the fact that they, almost by design, consist in treating protected groups differently.

Further clarification must be made for techniques aimed at imposing Conditional Demographic Parity (AdvCDP, ThreshCDP). These techniques result in rather poor performance when assessed with respect to the fairness dimensions present in Table 5.1. This is expected: imposing CDP means going in the direction of a more individual notion of fairness, where equality is requested only for people having some common conditions (in this example the same level of credit risk). This, of course, does not mean that these strategies produce less fair outcomes, it only means that they are fair with respect to a different notion of fairness.

Remains fundamental to carefully consider the notion of fairness appropriate for a specific domain and task. It is highly desirable that people with different expertise work together along the model development life-cycle, especially in its first steps and their iterations, where crucial decisions are taken about the notion of fairness and the information that can be used safely. We will explore this crucial step in Section 7, while Section 8 will introduce a tool designed to assist in making this delicate decision.

A similar argument holds for the counterfactual model, which indeed performs poorly in terms of group fairness dimensions, being it a method trying to enforce an individual notion of fairness. Moreover, it must be taken into account that counterfactual models are unfalsifiable by design, namely, there are many alternative counterfactual realities compatible with the same causal graph and implying, in general, a different definition of what is fair, and no observation can be used to choose among them.

However, the process of building and validating a causal graph for a specific use case marks an important step in the comprehension of the network of interdependence among the variables involved. The more these connections and dependencies are known, the more it is possible to understand what it means, in that specific situation, to be unfairly discriminated against. Thus, we believe it represents a valuable tool for the achievement of fairness.

Notice that the fairness dimension related to equal precision (PP) seems to be worsened by most mitigation techniques: this is coherent with the fact that most techniques aim at reducing DP or EO, thus "losing ground" in a measure like PP which is related to precision. Once again, it is worth mentioning that it is not possible to be fair with respect to every possible dimension.

Even if the performance and fairness requirements for the specific case are well-defined, still it is needed to choose the model that best meets them simultaneously. For this purpose, we have proposed two different approaches: (i) use the *constrained performance* method when the fairness constraint is clearly defined as in the case of the EEOC instances, otherwise (ii) use the *trade-off fairness-performance* method.

## 5.7 Experiment on Fairness Mitigation using Synthetic Data with Specific Bias Generated by `Bias On Demand`

As stated in chapter 3.2, the main goal of `Bias On Demand` is to provide a simple generative model able to reproduce datasets with (combinations of) fundamental types of bias. Such datasets can be useful to illustrate how various biases may occur in data and to investigate them. In particular, we shall make reference to the two specific examples of biased features in *college admissions* and biased labels in *financial lending* to showcase the effect of measurement bias and historical bias on fairness mitigation results.[4] Similar to Binns [26] we use these two specific examples to show that considerations w.r.t. fairness primarily depends on the ethical and social assumptions about the underlying phenomenon. In line with his consideration, we emphasise that the assumptions about worldviews also determine the understanding of the type of bias present in the data, with significant implications on the performance and fairness of ML-based decisions and on the effectiveness of different bias mitigation strategies.

We first generate datasets simulating toy scenarios with different assumptions about the presence and magnitude of relevant biases.Then, each generated dataset is used to train and test a supervised ML classifier[5] that aims to maximise performance by utilising all available variables. Alongside the unmitigated ML model, we train the same model *blinding* the sensitive variable $A$ (i.e. implementing the *FTU* approach), and we use *post-processing* mitigation strategies to enforce *DP* and *TPR parity*.We evaluate the outcomes both in terms of predictive performance (we shall use the overall *accuracy*) and of fairness (through the group differences with respect to the group fairness metrics introduced in chapter 4.3.

[4] In the full paper [15], we provide additional results for experiments that showcase other types of biases.

[5] Specifically, for the experiments presented here we use a *Random Forest* [23], but any other supervised ML classifier could be used as well. Also, notice that all bias examples presented here result in a disadvantage for individuals of the group $A = 1$ if left unmitigated.

Table 5.2 Group fairness criteria

| **Conditioning on $Y, D$** | **Group fairness criterion** | **Mathematical representation** | **Post-processing bias mitigation** |
|---|---|---|---|
| Unconditional | DP | $P(D=1 \mid A=0) = P(D=1 \mid A=1)$ | Corbett-Davies et al. [49] |
| | CDP | $P(D=1 \mid L=l, A=0) = P(D=1 \mid L=l, A=1)$ | |
| Conditioned on $Y$ | Separation | $P(D=1 \mid Y=i, A=0) = P(D=1 \mid Y=i, A=1), i \in \{0,1\}$ | Hardt et al. [81] |
| | TPR parity | $P(D=1 \mid Y=1, A=0) = P(D=1 \mid Y=1, A=1)$ | Corbett-Davies et al. [49], |
| | FPR parity | $P(D=1 \mid Y=0, A=0) = P(D=1 \mid Y=0, A=1)$ | Hardt et al. [81] |
| Conditioned on $D$ | Sufficiency | $P(Y=1 \mid D=j, A=0) = P(Y=1 \mid D=j, A=1), j \in \{0,1\}$ | |
| | PPV parity | $P(Y=1 \mid D=1, A=0) = P(Y=1 \mid D=1, A=1)$ | Baumann et al. [16] |
| | FOR parity | $P(Y=1 \mid D=0, A=0) = P(Y=1 \mid D=0, A=1)$ | |

In these experiments, our central objective is to evaluate the effect of different biases on fairness mitigation strategies. As a result, we exclusively focus on post-processing techniques, as our aim is not to optimize outcomes but rather to understand the interplay between bias and mitigation. Table 5.2 provides a summary of the group fairness metrics discussed in Chapter 4.3 and the corresponding post-processing techniques applied in this section.

### 5.7.1 Example 1: Biased Features in College Admissions

For the first example, we focus on the decision-making context of college admissions, where the task is to determine which candidates are more suitable for a degree program. Let us assume that the committee responsible for the admission decision heavily relies on SAT scores of applicants and that these are not independent of individual sensitive characteristics. We shall consider two alternative assumptions: *a) SAT scores are a faithful representation of applicants' skills and competencies*, vs. *b) SAT scores do not faithfully represent applicants' skills and competencies*. A general underlying assumption is that, absent any bias, skills, and competencies should be uniform across sensitive groups. Case *a)* is an example of the assumption CS $\approx$ OS. Thus, as argued in Section 2.1, SAT score disparities should be a consequence of some form of historical bias impacting the actual skills and competencies of applicants. On the other hand, case *b)* is the result of a measurement bias, where SAT scores are not the proper way to assess skills and competencies, and this creates undesired disparities. Referring to our notation in Section 3.2, case *a)* represents a form of *historical bias on R*, and case *b)* represents a form of *measurement bias on R*.

**Case a): historical bias on $R$** The generative model in this scenario is $Y = f(R,Q) + \varepsilon$, $R = R(A)$, $Q \perp\!\!\!\perp A$, with $Y$ depending on $A$ through $R$. Figure 5.4a shows the effect of different magnitudes of historical bias on $R$ (denoted by the parameter $\beta_h^R$). In the unconstrained case, all group fairness criteria are violated, and the group disparities are proportional to the size of $\beta_h^R$. Interestingly, blinding the model w.r.t. the sensitive attribute (i.e. *FTU*) has no effect since the dependence on $A$ is embedded in $R$. However, all other bias mitigation techniques manage to ensure the associated group fairness criteria. Post-processing the ML model to achieve *DP* (requiring equal acceptance rates across groups) is the only mitigation technique that is unconditional on $Y$. As a result, group-level differences in SAT scores are not reflected in the admission decisions, reducing the accuracy with increasing historical bias. However, as can be seen in Figure 5.4a, the between-group differences of other group fairness metrics (*TPR*, *FOR* and *PPV differences*) increase. Other bias mitigation techniques do not reduce the overall accuracy but also come at the cost of other fairness criteria, empirically confirming their theoretical incompatibility [108, 45]: for example, enforcing *TPR parity* increases *PPV* and *FOR differences* (even though it also brings the groups' acceptance rates closer together and thus has a positive effect on *DP* and *FPR differences*).

**Case b): measurement bias on $R$** In contrast to case *a)*, $Y$ and $R$ do not depend on $A$, with the only dependence on $A$ being in the proxy of $R$ ($P_R$). Figure 5.4b (and Figure 8b in Appendix B.2, which contains the full results) shows that, in general, the models can cope with the measurement bias on $R$ by leveraging the sensitive attribute $A$ (with a slight accuracy reduction), without any increase in unfairness. This is not the case when the model is blinded w.r.t. $A$ (*FTU*), i.e. the ML model can only cope with measurement bias on SAT scores as long as it is aware of the group memberships $A$. As can be seen in Figure 5.4b, *FTU* further reduces the accuracy and generates unfairness w.r.t. to all of the considered fairness metrics. This shows that *FTU* is not a suitable technique to deal with measurement bias on the features.

### 5.7.2 Example 2: Biased Labels in Financial Lending

For a second example, we focus on the scenario in which a bank uses an ML model to determine whether loan applications should be approved or denied. Let us assume that the bank notices that the labels are biased, i.e. the rate of repayment is not uniform with respect to gender. As in the first example, we are again considering two distinct scenarios, corresponding to the following two alternative assumptions: *a) historical bias on Y, i.e. the repayment rate disparity reflects a real mismatch in creditworthiness between men and* *women*; and *b) measurement bias on Y, i.e. the observed repayments are a skewed measure of real creditworthiness.*

**Case a): historical bias on $Y$.** Analogously to case *a)* in the first example (Section 5.7.1), the observed disparity represents a historical bias on $Y$ and is a consequence of structural discrimination, e.g. via factors like income disparities. As can be seen in Figure 5.5a, the resulting effects on the fairness and accuracy of the outcomes are very similar to the ones with historical bias on the labels $R$ (shown in Figure 5.4a). Only the bias mitigation technique *FTU* produces very different results: in contrast to historically biased features, the *FTU* approach is able to reach equality of acceptance rates in the case of historically biased features. Indeed, the generative equation reads $Y = f(R,Q,A) + \varepsilon,\ R,Q \perp\!\!\!\perp A$, which is why blinding $A$ is enough to achieve *DP*.

**Case b): measurement bias on $Y$.** For this scenario, the observed proxy $P_Y$ of the true outcomes $Y$ is increasingly biased with larger values of $\beta_m^Y$.The ML model is trained on the biased label $P_Y$, and also all bias mitigation techniques are conducted on $P_Y$. However, the final results are measured w.r.t. the real $Y$, which is unobservable in reality. Hence, Figure 5.5b shows that with increasing measurement bias on the labels, the accuracy continuously decreases, as the trained model is unaware of the bias in observed proxy $P_Y$ for the true label $Y$ (see Figure 9b in the appendix for the full results).

In this case, we assume that instead of measuring the actual creditworthiness of applicants, the repayment rate results are skewed in favour of men, for whom conditions are easier.[6]

In this case, the underlying phenomenon reads $Y = f(R,Q) + \varepsilon,\ R,Q \perp\!\!\!\perp A$, and the observed dependence on the sensitive attribute comes entirely from the proxy $P_Y$. The classifier is trained on this proxy, which is why it is calibrated against $P_Y$ but increasingly miscalibrated against the real $Y$ for group $A = 1$ the larger the measurement bias $\beta_m^Y$.Consequently, the produced outcomes are unfair w.r.t. the true $Y$, as visualised in Figure 5.5b. More precisely, measurement bias on the labels shifts the calibration curve of the ML model against $Y$ upwards, i.e. predicted scores underestimate the ratio of positives.

[6] Binns [26] mentions that such discrimination against women can be the result of bank clerks being systematically more lenient with loan repayment deadlines for men. This means that men ($m$) are more likely to end up repaying their loan despite not being more creditworthy compared to women ($w$), i.e. $\mathbb{E}(P_Y \mid A = m) > \mathbb{E}(P_Y \mid A = w)$ but $\mathbb{E}(Y \mid A = m) = \mathbb{E}(Y \mid A = w)$.

Most *post-processing* bias mitigation techniques fail to achieve any group fairness criteria. However, similarly to the decreasing accuracy, this is due to the fact that, in Figure 5.5b, accuracy and fairness are measured w.r.t. $Y$, and the "distance" between $Y$ and $P_Y$ grows with $\beta_m^Y$. Only *FTU* and enforcing *DP* through *post-processing* are effective in mitigating measurement bias on the labels. Both methods behave very similarly since they do not depend on the observed outcome $Y$ (and because the feature $R$ is free of any bias, i.e. it does not depend on $A$).Notice that those two techniques manage to fully mitigate any measurement bias on $Y$. For *FTU*, this effect occurs since using the unbiased feature $R$ without being aware of the group membership $A$ makes it impossible for the ML model to capture the group-level disparities in $P_Y$. In contrast, for *DP*, this is due to the linear implementations of the measurement bias and of the effect of $R$ on $Y$ (through $S$) using the parameters $\beta_m^Y$ and $\alpha_Y$, respectively (see Equations (3.6) and (3.5)). Namely, this shifts the distribution of $P_S$ (and, thus, its mean $\overline{P_S}$), which is equivalent to flipping the label from $Y = 1$ to $P_Y = 0$ for those individuals of group $A = 1$ that have $\overline{S} > P_S > \overline{P_S}$. Notice that in the lending scenario, we are considering, $S$ could represent an individual's true probability of repaying on due time. Thus, the linear shift of $S$ makes sense to replicate the lower leniency of bank clerks towards women (denoted by $A = 1$) when it comes to the repayment deadline. H

### 5.7.3 Discussion on the Result

#### Connecting worldviews and bias mitigation techniques

As outlined in both examples of Section 5.7, the type of bias present in a given dataset may depend crucially on assumptions about moral worldviews and, ultimately, about the data generation mechanism. This is even more important in light of the fact that mitigation strategies behave differently when facing different types of bias: measurement and historical bias have very similar patterns on data observable statistics but quite different consequences on the choice of the most appropriate bias mitigation strategy, as exemplified in the difference between Figures 5.4a and 5.4b, and between Figures 5.5a and 5.5b.

In Chapter 8, we will further explore the intricate relationship between worldviews and bias mitigation techniques.

**The biased label problem.** Our findings show that the solutions proposed by Corbett-Davies et al. [49], Hardt et al. [81], Baumann et al. [16] to post-process predicted scores effectively manage to mitigate biases as long as there is no measurement bias on the label.

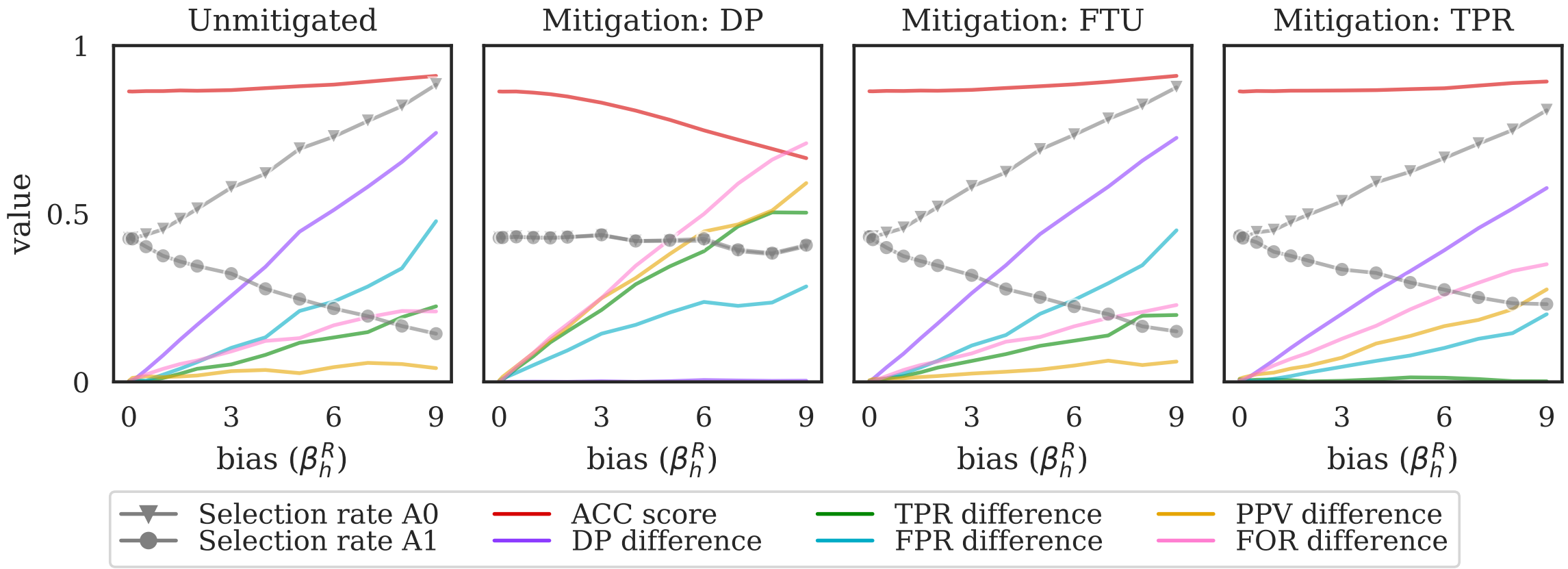

(a) Historical bias on $R$

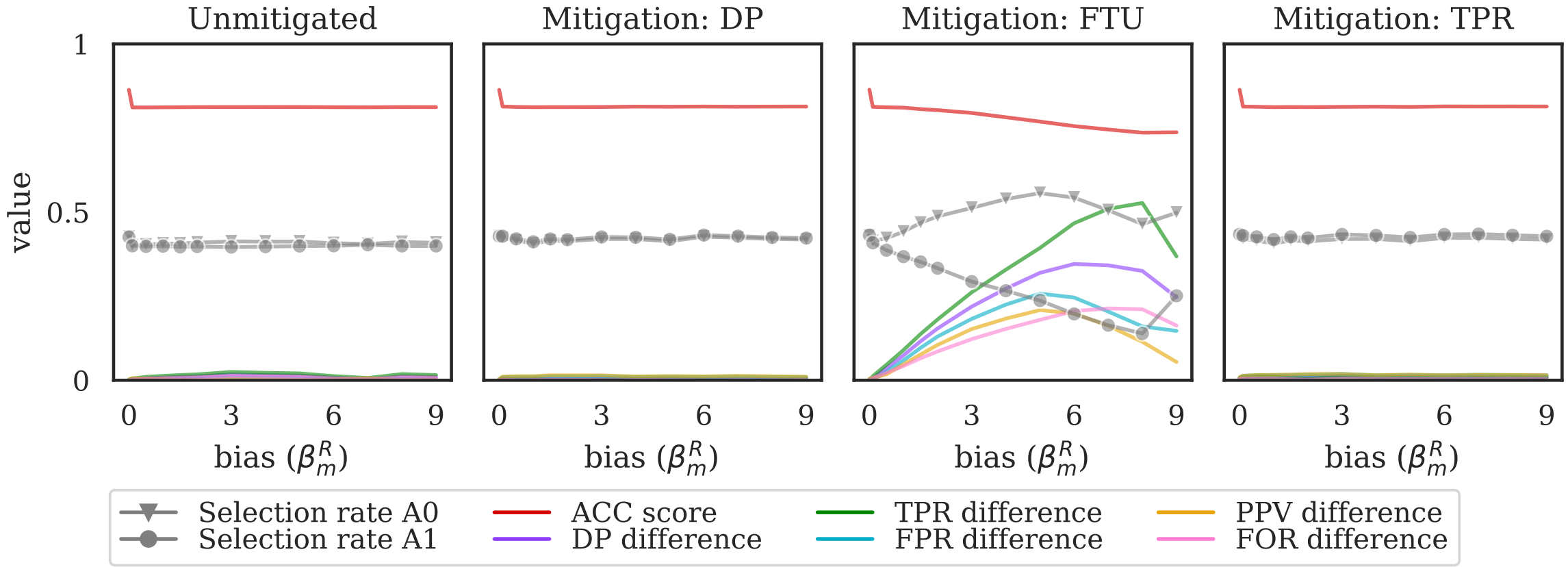

(b) Measurement bias on $R$

Figure 5.4 Accuracy and fairness metrics for biased features $R$ in the college admission example.

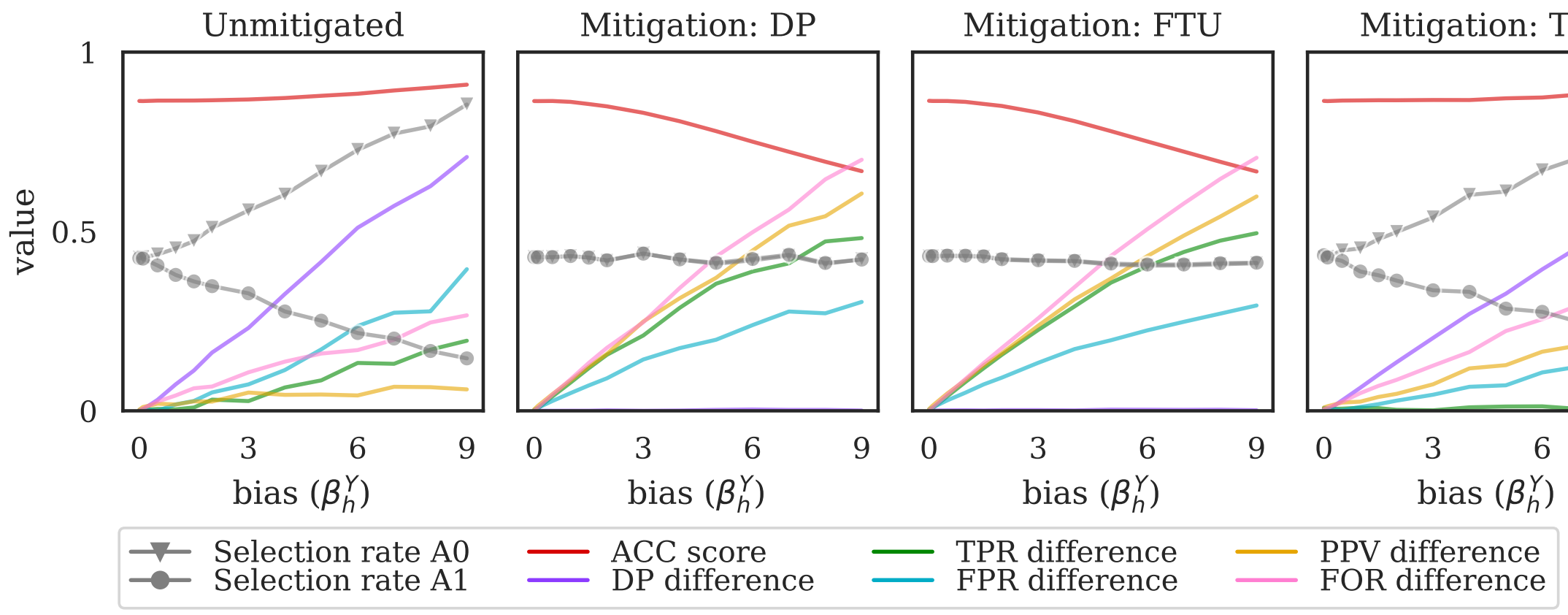

(a) Historical bias on $Y$

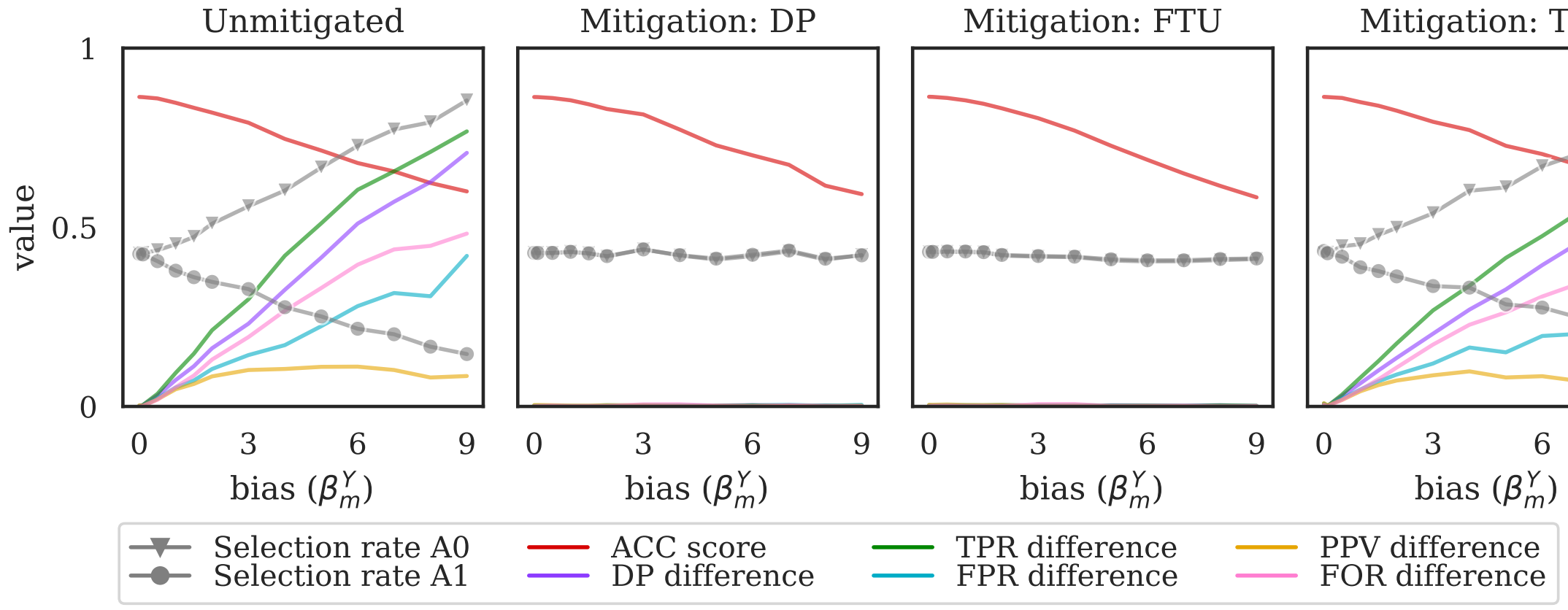

(b) Measurement bias on $Y$

Figure 5.5 Accuracy and fairness metrics for biased labels $Y$ in the financial lending example. Notice that all metrics in (b) are computed with respect to the "true" target $Y$.

As Figure 5.5b shows, the case of measurement bias on labels is particularly subtle: having access only to the (biased) proxy $P_Y$, it is only possible to control the bias when imposing fairness criteria that do not make use of target variable, namely *DP* and *FTU* in our experiments. For all other criteria (*TPR*, *FPR*, *PPV*, and *FOR parity* or combinations of these), one would mitigate the group differences of errors with respect to $P_Y$ (and not to $Y$), and thus would be erroneously induced to judge the model as fair and accurate when it is not.

**On the trade-offs of ML-based decision-making systems.** Our findings empirically confirm the existence of different trade-offs emerging when enforcing fairness in ML-based decision-making systems [72]. There is a trade-off between fairness and performance (e.g., measured by accuracy), as well as between different notions of fairness.

Figures 5.4-5.5 show that in certain cases, a biased dataset results in lower overall performance compared to an unbiased dataset. Furthermore, the application of bias mitigation techniques generally comes at an increasing cost, in terms of performance, as the bias increases. However, this is not necessarily the case in all situations, meaning that the performance-fairness trade-off may sometimes be negligible, which is in line with the findings of Rodolfa et al. [157]. As our experiments show, depending on the bias present in the dataset, certain bias mitigation techniques are 'cheaper' in terms of accuracy. At the same time, not all post-processing bias mitigation techniques result in equal outcomes for the affected individuals. Assessing whether the system is fair for the affected individuals requires normative choices w.r.t. what constitutes a just outcome. Several works emphasise how the moral appropriateness of certain notions of fairness heavily depends on the context, such as Baumann and Loi [17], who provide an ethical argument in favour of the *sufficiency* criterion in the context of personalised insurance premiums or Binns [25], who argues that in other applications (such as the selection of candidates from a pool of job applicants) *equality of opportunity* might be more appropriate.

Our findings empirically confirm that there is a trade-off between different notions of fairness: enforcing some fairness criteria may come at the cost of others. This is a logical consequence of the mathematical incompatibility between certain fairness criteria [108, 45]. As can be seen in Figures 5.4-5.5, group-level differences in *PPV*s are usually relatively small compared to acceptance rate differences, and enforcing *TPR* or *FPR parity* oftentimes brings those acceptance rates closer together. Furthermore, our experiments show that *DP* mitigates every group fairness criteria in the case of measurement bias on the label $Y$ or on the feature $R$. However, as explained in Section 5.7.2, this is due to the linear implementations for those biases. Alternatively, if there is historical bias, clear trade-offs between the different bias mitigation techniques and w.r.t. accuracy emerge.

**On the effect of blinding.** Despite its simplicity, *FTU* should be applied with particular care, as other works have pointed out [116, 63, 101, 43, 48]. Even though the reason for applying *FTU* is not primarily to achieve group fairness, we believe that it is still relevant to take its effects on different group fairness criteria into account. In fact, in some very particular cases, *FTU* can be effective even w.r.t. group fairness metrics, e.g. when the observed proxy for the target variable $P_Y$ is wrongly assumed to be free of bias (see Figure 5.5b). However, in general, it is not an effective bias mitigation technique w.r.t. the fairness of the produced outcome for the affected individuals, as shown for the example of historically biased features $R$, where *FTU* has no effect whatsoever since the information on group membership is redundantly encoded in $R$ (see Figure 5.4a). And, in fact, there are even cases in which the application of *FTU* leads to biased results and performance deterioration even when the unconstrained model does not (see Figure 5.4b).

# 6

# `FFTree`: A Flexible Tree to Mitigate Multiple Fairness Criteria

As detailed in [185], the mitigation strategies proposed in prior studies, as well as those introduced in the preceding chapter, typically lack flexibility with respect to one or more of the following aspects: (i) they are specifically designed for a specific fairness criterion and, as a consequence, cannot accommodate more than one simultaneously (e.g., disparate treatment and disparate impact); (ii) they cannot ensure fairness with respect to multiple sensitive features simultaneously (e.g., gender and race); (iii) They are only limited to a narrow range of classification models. Moreover, the proper mitigation model depends on the chosen definition of fairness and, finally, they are typically designed as black boxes, i.e., they are not directly interpretable.

To overcome these limitations we introduced `FFTree` [34], a transparent, flexible, and fairness-aware decision tree. As a novelty, `FFTree` enhances the classical approach introduced in [29] with a method to find "fair" splits designed to work with different fairness criteria. The method presented in this work aims to support the user in the creation of fair decision systems through the following peculiarity:

- *flexibility*. Depending on the application domain, the user can select the proper fairness criteria choosing between a metric in the family of disparate impact (DI) or disparate mistreatment (DM). Moreover, `FFTree` permits to implement multiple fairness defini-tions (MD) or mitigate on multiple sensitive (protected) groups (MS) simultaneously. The user can also set the required level of fairness as an input parameter to meet different business needs (BN) or regulatory requirements.

- *transparency*. Since it is based on a decision tree, `FFTree` is transparent by design. Furthermore, the constraints introduced in the system are directly derived from the chosen fairness criteria and explicitly added to the model.

We evaluate `FFTree` on both: (i) a benchmark dataset [62] well-known in the literature, and (ii) a real-world private dataset owned by Intesa Sanpaolo Bank containing about 250,000 records related to the granting of personal loans.

We show that `FFTree` is competitive with state-of-the-art fair trees in the common features, and thanks to its novel functionality, it can support users in very different application domains.

## 6.1 `FFtree` Implemented Fairness Criteria

In this section, we describe the criteria of fairness implemented in `FFTree` to set the stage for the discussions and experiments presented in later sections.

To provide clarity and facilitate understanding for the reader, we will briefly repeat the presentation of the criteria and their corresponding abbreviations. Despite their prior detailed introduction in Chapter 4, this repetition is useful considering their frequent use throughout this chapter. We use $Y$ to indicate the *"target label"*, (i.e., the information that we want to predict), and $\hat{Y}$ is the prediction, which may be the output of a model. $S$ indicates the *"sensitive label"* (i.e., the class based on which we explore the concept of fairness). Finally, $X$ represents all the other inputs that are useful in obtaining the prediction.

### 6.1.1 Disparate Impact (DI), Disparate Mistreatment (DM), and Disparate Treatment (DT)

Disparate impact is a "group" fairness notion closely related to the concept of *Independence* [13]. Independence means that the prediction $\hat{Y}$ does not depend on a sensitive attribute S:

$$\hat{Y} \perp\!\!\!\perp S. \tag{6.1}$$

This can also be expressed as follows:

$$P(\hat{Y}=1 \mid S=a) = P(\hat{Y}=1 \mid S=b), \quad \forall a,b \in S. \tag{6.2}$$

The most popular metrics used to measure independence are *disparate impact* and *demographic parity*. These are expressed as the ratio and the difference respectively between the percentage of positive outcomes between the different classes of the sensitive attribute.

Disparate mistreatment was introduced by Zafar et al. [184] to refer to all group fairness notions relying on disparity errors (i.e. all group metrics dealing with comparisons among model predictions $\hat{Y}$ and true outcomes $Y$).

The concept of disparate mistreatment can be subdivided into *separation* and *sufficiency* [13].

Separation means that given the true class of the target label, the model prediction is independent of the sensitive class. Separation is satisfied if:

$$\hat{Y} \perp\!\!\!\perp S \mid Y. \tag{6.3}$$

In other terms:

$$P(\hat{Y}=\hat{y} \mid S=a, Y=y) = P(\hat{Y}=1 \mid S=b, Y=y), \forall a,b \in S,\ y \in \{0,1\}. \tag{6.4}$$

There are two relaxed versions of this criterion:

- *Predictive equality (PE)*: This refers to equality of the false positive rate across groups:

$$P(\hat{Y}=1 \mid S=a, Y=0) = P(\hat{Y}=1 \mid S=b, Y=0), \quad \forall a,b \in S. \tag{6.5}$$

- *Equality of opportunity (EO)*: This represents the equality of false negative rates across groups:

$$P(\hat{Y}=0 \mid S=a, Y=1) = P(\hat{Y}=0 \mid S=b, Y=1), \quad \forall a,b \in S. \tag{6.6}$$

Sufficiency means that a prediction from the model, the true class of the target label is independent of the sensitive class. Sufficiency is satisfied if:

$$Y \perp\!\!\!\perp A \mid \hat{Y}. \tag{6.7}$$

While separation deals with error rates in terms of the ratio of errors to the ground truth (for example the number of individuals whose loan request is denied compared to those who would have repaid), sufficiency takes into account the number of individuals who will not repay of those who are given a loan.

A fairness criterion that focuses on this type of error rate is called *predictive parity (PP)* [45]:

$$P(Y = 1 \mid S = a, \hat{Y} = 1) = P(Y = 1 \mid S = b, \hat{Y} = 1), \forall a,\ b \in S, \tag{6.8}$$

i.e. the model should have the same precision across sensitive groups.

All the measures of this family of fairness metrics are *conservative* [153]: they are always respected in the case of a perfect predictor. This property is important because they do not require any changes to obtain fairness, i.e. they conserve the *status quo* [131].

Disparate treatment, also known as *individual fairness*, is based on the following principle: *similar individuals should be given similar decisions*. This principle deals with a comparison between single individuals rather than focusing on groups of people sharing certain characteristics. The first attempt to deal with a form of individual fairness was presented in [63], where this concept is introduced as a Lipschitz condition on the map $f$ from the feature space to the model space:

$$dist_Y(\hat{y}_i, \hat{y}_j) < L \times dist_{\widetilde{X}}(\widetilde{x}_i, \widetilde{x}_j), \tag{6.9}$$

where $dist_Y$ and $dist_{\widetilde{X}}$ denote suitable distances in the target space and feature space, respectively, and $L$ is a constant.

Another way to represent disparate treatment is to define similar individuals as couples belonging to different groups with respect to sensitive features but with the same values for all the other features. In this approach, we require that the outcome should be unchanged if we take an observation and change only its protected attribute $S$. This concept is usually referred to as *Fairness Through Unawareness (FTU)* or *blindness* [176], and is expressed as the *requirement to avoid explicitly employing protected attributes when making decisions*.

### 6.1.2 Business Needs (BN)

In many real domains, the minimum level of fairness that should be reached is not always clear [39]. A well-known statement from the U.S. Equal Employment Opportunity Commission (EEOC) [66] adopted the so-called 80% rule, stating that the disparate impact should be no less than 80:100. At the time of writing there is no evidence of other fairness regulations in which the metric and the maximum level that should be adopted are specified.

The permitted level of discrimination is an important parameter for companies since under certain conditions the level of fairness is negatively correlated with performance [22] (e.g. more fairness means less performance). In general, the methods used to reduce discrimination do not allow the user to set the required level of fairness as an input.

### 6.1.3 Multiple Fairness Definitions (MD)

The previous fairness notions account for either direct (or intentional) or indirect (or unintentional) unfairness [7]. More in detail, *disparate treatment* takes into consideration direct unfairness, which occurs when a decision-making process directly leverages the sensitive feature information to put a group of people sharing a value of a sensitive feature on a relative disadvantage. Contrarily, *disparate impact* and *disparate mistreatment* account for indirect unfairness, i.e., when a decision-making process unintentionally or indirectly exploits the correlation between sensitive features and the target variables to penalise the sensitive population subgroups.

Furthermore, whereas DI and DM both account for indirect unfairness, their typical application domains drastically differ. In an application where there is no ground truth information for judgments and the past decisions utilised during training are unreliable and hence cannot be trusted, the disparate impact criteria are particularly appropriate [14]. On the other hand, domains where the user can completely trust the target variable, such as fraud detection or image recognition, require fairness criteria that conserve the *status quo*, where is possible to define an *optimal classifier* as *perfectly fair* [166]. Since, in a real complex use case is unlikely to develop perfect classification systems, the user aims to not introduce disproportions of errors between different population subgroups, by using DM criteria.

Despite those guidelines, it is not always easy to determine the most appropriate fairness criterion for the domain under consideration. Attempting to simultaneously obtain different fairness criteria is a reasonable and unarguable strategy. The literature has not frequently taken this path because applying multiple fairness criteria can result in significant performance losses [45] or even in the absence of algebraic solutions [13] (see Section 6.3.2 for further details). However, investigating the best results that mitigation on multiple fairness criteria can produce in practice, even if weak, may still be interesting [185].

### 6.1.4 Multiple Sensitive Features (MS)

Although different fairness criteria are taken into account using various formalisations, as previously discussed in chapter 4.3.6, the majority of them do not take the mitigation on multiple sensitive features [31] into consideration. However, this approach is of limited practical usefulness, because real-world scenarios usually present multiple sensitive attributes. For instance, it is easy to expect that the loan will be granted based on fairness criteria on gender, age, and citizenship. When taking into account several sensitive features, a typical approach is to construct all possible combinations of the sensitive feature values (for example, man-citizen and woman-not-citizen) and add restrictions for each combination independently to prevent "fairness-gerrymandering" [105].

## 6.2 Related Works on Interpretability and Discrimination-Aware Decision Tree

Our work is closely related to the *discrimination-aware decision tree*, which is an intersection between *fairness* and *interpretability* in supervised machine learning. Regarding the fairness mitigation methods extensively outlined in Chapter 5, this system is classified as an in-processing mitigation approach. This categorization stems from the direct imposition of mitigation constraints within the algorithm's learning function (see section 6.3 for further details on the implementation).

### 6.2.1 Interpretability in Machine Learning

Lipton [103] offers a comprehensive classification of the desired attributes and techniques for achieving interpretable AI. We adopt the interpretation of interpretability as proposed by Biran and Cotton [9], denoting it as the extent to which an observer can comprehend the rationale behind a decision. Explanation stands as one avenue through which an observer can attain such comprehension, but it's important to note that alternative avenues also exist, such as formulating inherently more understandable decisions or employing introspection. As in [132], we consider 'interpretability' and 'explainability' to be synonymous.

More in deep, an explanation is essentially an "interface" between humans and an automatic decision system that is at the same time both an accurate proxy of the decision model and comprehensible to humans [79]. At a very high level, there are two approaches in the literature for producing explanations: (i) "black box" explanations and (ii) "transparent box" design [79]. The former, also known as *post hoc* interpretability, involves reconstructing an explanation for the decisions generated by a ubiquitous opaque decision system, whereas the latter consists of developing decision systems that are interpretable by design. Explanation technologies are an immense help to companies in terms of producing AI systems that are trustworthy[1].

In the field of machine learning, interpretability is defined as the ability to explain or to provide meaning in terms that are understandable by humans [61, 79]. An interpretable model should provide both *comprehensibility* and *accuracy* [8, 61, 69, 97]. Comprehensibility is related to the complexity of the model: an interpretable model must have the lowest complexity as possible [69]. An accepted proxy for measuring the complexity of a predictive model is based on the model size. In contrast, accuracy is related to the performance of the model. The development of an interpretable model that can ensure the expected levels of performance is the most common aim among the works in this field [33]. In the literature a small set of existing interpretable models is recognised: *decision trees*, *rules* and *linear models* [69, 91, 156]. These models are considered to be easily understandable and interpretable by humans. An interpretable model may be developed to provide both predictions and explanations (transparent box design), or simply to provide explanations of a non-interpretable but well-performing model (black box explanation) [79]. In the second case, another important driver for evaluating the ability of an interpretable model to mimic the behaviour of a black-box is the *fidelity* [175].

The adoption of AI depends mainly on the trust that users have in it. Interpretability offers a great opportunity to promote trust in AI [86]; when AI is used to support decision-makers such as judges or doctors, these users need to understand the reasons behind a suggestion to trust the AI system and decide to adopt it [109].

`FFTree` is based around the use of a decision tree engine to provide transparent rules to the users of the model, and can be used as both a transparent decision model and a surrogate for a black box.

### 6.2.2 Discrimination Aware Decision Tree

A decision tree classifier [29] is one of the most popular algorithms in machine learning [159] given its intelligibility. It takes the form of a tree-like structure consisting of nodes, branches, and leaves. At each internal node of the tree, an evaluation is performed, and each branch represents the result of this evaluation. Each leaf collects all points that gave the same answers to all the evaluations on its path. The paths from the root to the leaves represent the classification rules, and a decision tree can be represented as a set of decision rules in *if-then* format [68, 150].

[1] As detailed in [86], dimensions of trustworthy AI include: security, safety, privacy, non-discrimination, fairness, accountability (re-traceability, replicability), auditability, environmental, well-being, robustness, and explainability.

In a classical decision tree, each evaluation is based on local optimisation of the overall accuracy introduced by the new decision rules, and this is known as *information gain* (Information Gain (IG)). In a discrimination-aware decision tree, fairness criteria are taken into consideration in terms of the IG in the evaluation at each internal node.

The first *"fair tree"* was introduced in [102], while the authors of [191] and [6] have proposed important developments.

Kamiran et al. introduced the following two methods for incorporating discrimination awareness into the decision tree-building process:

- *Dependency-Aware Tree Construction.* When evaluating the splitting criterion for a tree node, not only its contribution to the accuracy but also the level of discrimination - defined as DP - caused by this split is evaluated.
- *Leaf Relabeling.* After the completion of the construction of the tree, the label of selected leaves is changed in such a way that discrimination is lowered with a minimal loss in accuracy.

In their work, the best results were achieved by combining the accuracy gain with the discrimination gain with an addiction function in conjunction with additional leaf relabeling operations, thus combining in-processing and post-processing fairness interventions.

Zhang and Ntoutsi tackled the problem of building a *fair tree* within real-world applications where data is generated in streaming and its characteristics may change over time [40]. In order to simultaneously address these challenges, they introduced a fairness-aware decision tree for data stream with the following main contributions:

- *Reformulate the fair information gain in the splitting criterion.* They combined fairness gain and information gain to a joint objective function that evaluates the suitability of a possible splitting attribute in terms of accuracy and fairness. As in [102] the discrimination is defined using DP. The distinction is that the fairness gain is directly defined in terms of the reduction in discrimination after the split than mediating between the entropy w.r.t sensitive attribute.
- *Adaptive fairness-aware learner for online stream classification.* Their *fair tree* builds upon the Hoeffding tree classifier [60] that is able to adapt to the changes in the underlying data distribution making confident attribute splitting decisions in an online fashion.

In their experiments, [191] showed in comparison with [102], a slight reduction in performance, but a better fairness-performance trade-off, while guaranteeing data-stream adaption.

Finally, Aghaei et al. introduced a novel *fair tree* with the following contributions:

- *More mitigation choices*. They formalised two types of discrimination: disparate impact - as in the previous *fair trees* - and disparate treatment, mathematically for both classification and regression tasks. Moreover, the trade-off between accuracy and fairness is conveniently tuned by a single, user selected parameter.
- *Computational optimization*. They proposed a unifying mixed-integer programming framework for designing *optimal* and *fair* decision trees for classification and regression.

They conducted extensive computational studies claiming that their framework improves the state-of-the-art to yield non-discriminating decisions at a lower cost to overall accuracy.

As a limitation, all of the works described here search for a gain of a function that combines a criterion of fairness and accuracy, at each node of the tree. However, this method does not ensure the reduction of the fairness below an *explicitly* settable level - large gains in accuracy with slight reductions in fairness might be preferred by the combined function as well as a large reduction in fairness (*over the requirements*) with not optimal performance (*under fairness constraint*). Moreover, these methods are specialised for specific fairness criteria for the mitigation of a single sensitive attribute - the extension to other fairness criteria or the mitigation of multiple sensitive attributes is complex and requires a *different* new formalisation of the fair information gain.

In this work, we tackle the fair-gain problem by searching for the best IG subject to a fairness constraint. Our approach to this problem offers several benefits: (i) there is no need to compare classification performance with fairness in the evaluation at each internal node, thus allowing the formalisation of several fairness criteria without having to intervene in the fair information gain function; (ii) each rule is compliant with the fairness constraint, meaning that we can prune the decision tree and continue to be fair; (iii) we can add constraints to easily handle more than one fairness definition or more than one sensitive attribute simultaneously; and (iv) is possible to set the required level of fairness as an input parameter to meet different types of business or regulation requirements and optimise performance under it.

Table 6.1 Capabilities of different discrimination-aware decision trees in terms of mitigating meeting different fairness criteria: disparate treatment (DT), disparate impact (DI), disparate mistreatment (DM), multiple fairness definitions (MD), multiple sensitive features (MS) and business needs (BN).

| *State-of-the-art Fair Tree* | DI | DT | DM | MD | MS | BN |
|---|---|---|---|---|---|---|
| Kamiran et al. | ✓ | ✗ | ✗ | ✗ | ✗ | ✗ |
| Zhang and Ntoutsi | ✓ | ✗ | ✗ | ✗ | ✗ | ✗ |
| Aghaei et al. | ✓ | ✓ | ✗ | ✗ | ✗ | ✓ |
| `FFTree` (our method) | ✓ | ✓ | ✓ | ✓ | ✓ | ✓ |

Table 6.1 compares the capabilities of different discrimination-aware decision trees in terms of meeting different fairness criteria through a scheme inspired by Zafar et al. [185]. Since `FFTree` covers all the fairness criteria could be adopted in a larger variety of application domains.

## 6.3 How does `FFTree` Work

In a classic decision tree, given the domain $dom(X_i)$ of the input features $X = X_i,\ i = 1,..,n$ and the target label $Y$, the goal of a splitting criterion is to find a rule to separate a node $D$ into *"purer"* branches $D_j$. The concept of purity is related to the target label $Y$, whereas the splitting criterion is determined by iteratively observing the variables $X$ and their domains.

In formal terms, an increase in purity is called IG, and can be expressed as the differences in the entropy $H$ of $Y$ between the node and the average for the branches created by the splitting criterion $a$:

$$IG(a) = H_Y(D) - \sum_{j=1}^{m} \frac{|D_j|a|}{|D|} H_Y(D_i|a) \tag{6.10}$$

Among all the splitting criteria under consideration, the one that locally (at the node) optimises the IG is chosen (see Figure 6.1 (a)). In our approach, the best splitting criterion is the one that locally optimises the IG, from among the options that satisfy the imposed fairness constraints $FC$.

$$\max IG(a) \quad \text{s.t.} \quad FC(a|S) \leq \delta$$

Where $FC$ is a fairness constraint, $S$ is a binary sensitive attribute, and $\delta$ is the maximum level of discrimination locally permitted (BN in Tab.6.1).

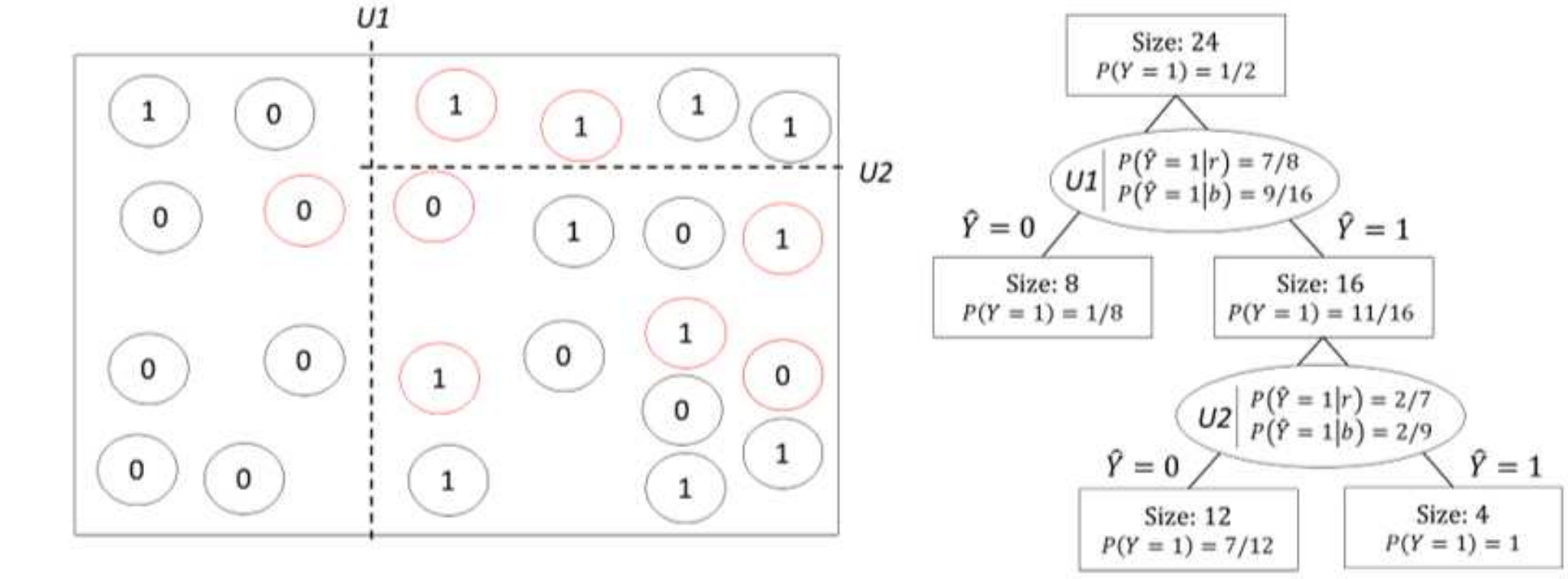

(a) Example of a decision tree without fairness constraints

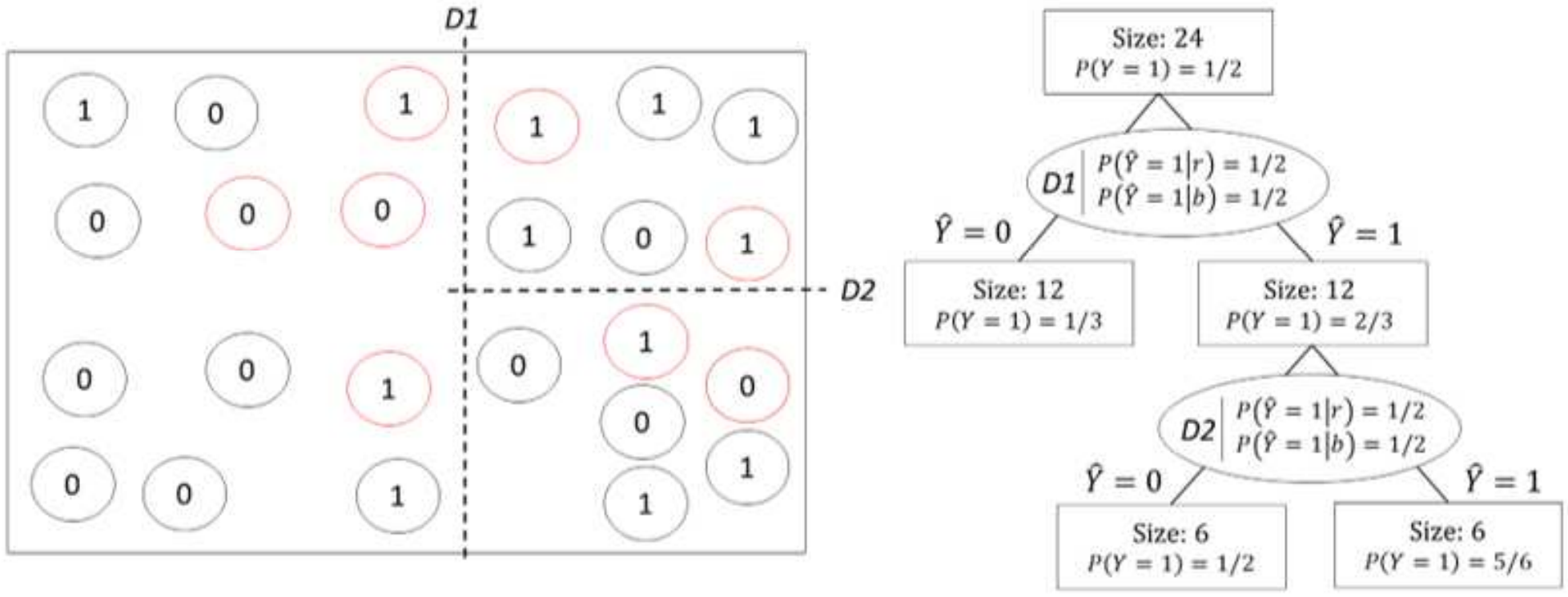

(b) Example of a decision tree in which demographic parity is imposed

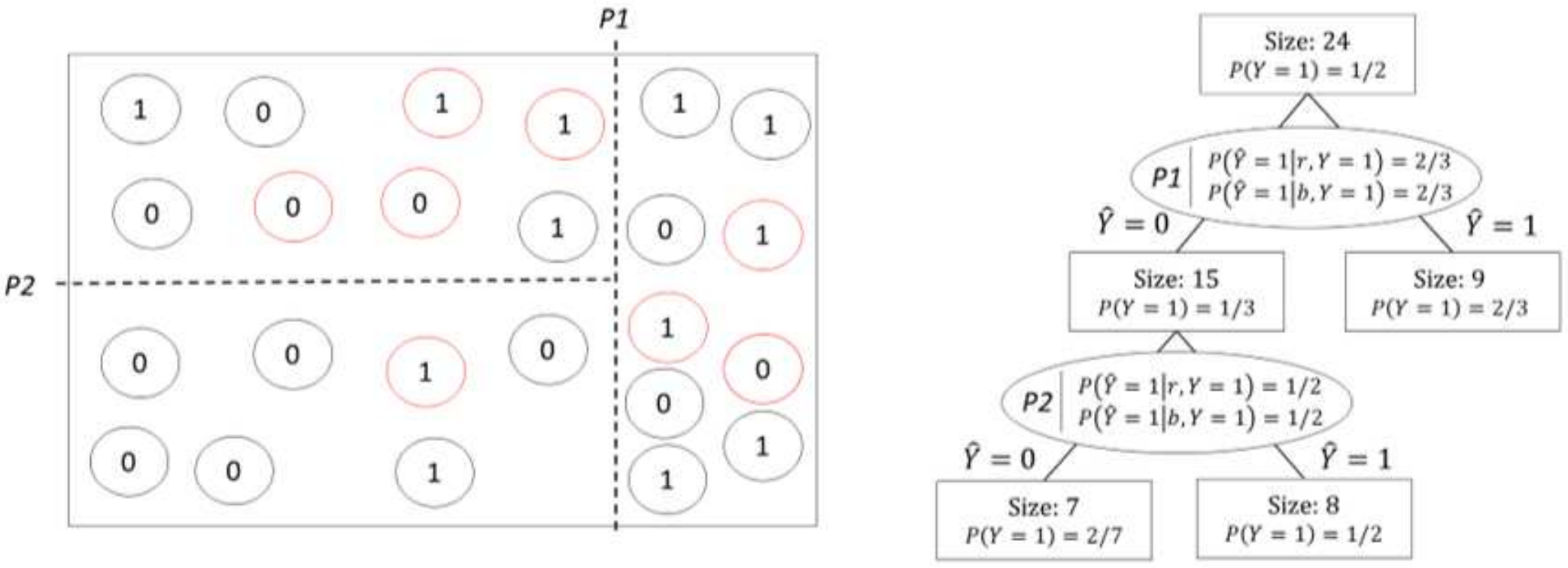

(c) Example of a decision tree in which predictive parity is imposed

Figure 6.1 Three examples of decision trees with different splitting criteria. The number in each circle represents the target variable $Y$, while the colour of the border (red ($r$) or black ($b$)) represent the sensitive attribute $S$.

At the time of writing, `FFTree` allows us to choose $FC$ among the observational[2] fairness equations defined in Section 6.1 (Demographic Parity (DI), Predictive Equality (DM), Equal Opportunity (DM), and Predictive Parity (DM)), expressed as the absolute value of the difference between the two probability terms (e.g. $PE = |P(\hat{Y} = 1 \mid S = a, Y = 0) - P(\hat{Y} = 1 \mid S = b, Y = 0)|$. `FFTree` permits to choose more than one fairness metric and more than one sensitive attribute $S$ (MD and MS as in Tab.6.1). In this case, all the fairness criteria must be satisfied by the best split; if no solutions are found the node becomes a leaf. The "sensitive" features $S$ are not used to find the best IG, therefore the *"fairness through blindness"* criterion is always provided (DT in Tab.6.1).

We set the number $m$ of partitions generated by the splitting criterion to two, as in a classic decision tree. To calculate the fairness metrics we compute the local prediction $\hat{Y}$ of each split assigning a value of one to the branch with the higher percentage of $Y$ and zero to the other branch. Calculating the metrics of *separation* and *sufficiency* requires $S$, $\hat{Y}$ and $Y$ (see Figure 6.1 (c)), while *independence* requires only $S$ and $\hat{Y}$ (see Figure 6.1 (b)).

Algorithm 1 shows the pseudo-code related to the local research for the best question, subject to multiple fairness criteria.

The output of `FFTree` is a rank where the score of each instance is expressed as the probability to get a positive outcome $P(Y = 1)$ according with the leaf calculate using the training-set. The classification could be obtained by comparing the score with user-configurable thresholds $\tau$.

### 6.3.1 Properties

When `FFTree` is trained to provide perfect independence in each local branch ($\delta = 0$), the resulting global tree is *fairness-threshold-invariant*, since by changing the classification threshold $\tau$, the output continues to provide perfect independence. In formal terms:

$$\delta = 0 \Rightarrow \hat{Y} \perp\!\!\!\perp S \quad \forall \tau, \quad where \quad \hat{Y}_i = \begin{cases} 1 & if P(Y_i = 1) > \tau \\ 0 & if P(Y_i = 1) \leq \tau \end{cases} \tag{6.11}$$

This fairness criterion was introduced by Jiang et al. in [95] and is known as *strong demographic parity*. Moreover, in this configuration, `FFTree` is also *fairness-pruning-invariant* - thanks to the implementation of the constraints at each local level, the user can prune the

[2] A fairness criterion is called "observational" if by knowing the joint distribution of the random variables $(\hat{Y}, S, Y)$ is possible determine its value without ambiguity.

**Algorithm 1** Find Best Question

```
Require: X, S, FC, δ as defined in section 6.3;
for all non sensitive features X_i do
    for all dom(X_i)_j do
        if X_i is numeric then
            Split the rows in two branches through the question X_i > dom(X_i)_j
        else
            Split the rows in two branches through the question X_i = dom(X_i)_j
        end if
        calculate IG_i j through eq. 1
        for all chosen methods FC_k do
            for all chosen sensitive attribute S_w do
                if FC_k|S_w ≥ δ_k w then
                    IG_i j = 0
                end if
            end for
        end for
    end for
end for
return Xi, dom(X_i)_j with the max(IG_i j)
```

resulting tree without losing fairness. In the Figure 6.1 (b) a practical example of these properties is shown. Also when $\delta > 0$, the optimisation at local level implies that all the rules shown by `FFTree` are consistent with the chosen fairness constraint, while other fair trees, based on the *fairness gain*, could be composed of a series of rules which, if taken individually, could be unfair. We suggest to tune the $\delta$ hyperparameter to identify at the global level the best performance-fairness trade-off for the application domain (see Section 6.4.3).

Finally, the main feature that distinguishes `FFTree` is *flexibility*. In this work, we presented the method encapsulating the most adopted fairness criteria, introduced in Section 6.1. However, the adopted approach easily allows developers to implement new criteria to meet other possible users need.

### 6.3.2 Limitations

As described in chapter 4.3.5, meeting different fairness criteria at the same time is not always possible. These criteria constrain the joint distribution in non-trivial ways. We should therefore suspect that imposing any two of them simultaneously will over-constrain the space to the point where only degenerate solutions remain. Tab.6.2 resumes the assumptions

Table 6.2 incompatibility relationships between fairness criterion [13].

| Assumption | Mutually exclusive criteria |
|---|---|
| S and Y are not independent. | Independence and sufficiency |
| Y is binary, S and Y are not independent, $\hat{Y}$ and Y are not independent. | Independence and separation |
| All events in the joint distribution of (S,$\hat{Y}$,Y) have positive probability, S and Y are not independent. | Separation and sufficiency |
| S and Y are not independent, $F(Y)$ is a binary classifier with a nonzero false positive rate. | Separation and sufficiency |

under which different fairness criteria become mutually exclusive (see [13] for proof of these propositions).

FFTree searches for the best local existing solution under the required constraints. As long as a solution within the application data exists, the tree grows. Therefore, FFTree gives the possibility to implement more fairness criteria at the same time, but does not guarantee the presence of a solution; it is up to the user's sensitivity to understand how many and which constraints to configure.

FFTree's favourable and novel properties in providing the user flexibility and greedy compliance to fairness constraints, can cost a loss in classification performance higher than that introduced by the other fair trees. The works of Zhang and Ntoutsi and Aghaei et al. remain the *state-of-the-art* for online data stream classification and regression tasks, respectively.

## 6.4 Experiments

**Experiments Overview.** We conduct experiments to verify the flexibility of FFTree; we want to validate the hypotheses expressed in Tab.6.1 where it is stated that FFTree can be adopted to satisfy DI, DM, BN, MD, and MS. Through the experiments, we want to show that the performances of our approach are comparable to that of the other *fair trees* in the *state-of-the-art* on common features (DI as in Tab. 6.1) and to demonstrate further situations in which the new features introduced by FFTree can be useful in real applications (DM, BN, MD and MS as in Tab. 6.1). In particular, in Section 6.4.1 we compare FFTree with the other fair trees in literature in satisfying demographic parity. In Section 6.4.2 we show the results of FFTree when the goal is to achieve a fairness metric in the DM family. Section 6.4.3 aims to demonstrate different fairness-performance trade-offs changing the input parameter $\delta$. Finally, Section 6.4.4 shows an experiment where multiple constraints are implemented to provide MD and MS.

The fairness metrics calculated in the experiments are explicitly formalised in Section 6.1.

**Experiments Settings.** The experiments are performed using two datasets: (i) *Adult* [62], a benchmark dataset that is available online, and on which the other *fair trees* were evaluated, and (ii) a real dataset owned by Intesa Sanpaolo containing data on the granting of personal loans.

The *Adult* dataset consists of over 30,000 records, each of them representing a person in terms of 14 employment and demographic attributes. The sensitive features are the binary variable “gender” and the multi-class variable "race". We encoded the sensitive variable "race" to create the binary variable "Caucasian". The learning task involved predicting whether the annual income of a person would exceed 50K dollars.

The real dataset used is the same as the one described in section 5.6.1, which concerns credit lending.

For both datasets, we encode all the variables as *dummies* to increase the interpretability of the rules provided by the `FFTree` . First, all the *numeric* variables are encode in range representing the quartiles. Then all the variables are encode through the so called *One Hot Encoding*. To validate the results, we adopt the *k*-fold cross validation using 5 as *k*. In the results, we show the median outcome in terms of accuracy and the related fairness.

### 6.4.1 Evaluating `FFTree` with state-of-the-art fair tree (DI)

As shown in Table 6.1, every *fair tree* in the literature is design to satisfy disparate impact (DI). In all these works an evaluation is performed with the *Adult* dataset.

With this experiment we intend to demonstrate that when $FC = DP$ the final results in fairness and performance are comparable. We selected two distinct configurations of the $\delta$ parameter (0.2 and 0.05) because the outcomes from the state-of-the-art evidenced different trade-off between fairness and performance. In Table 6.3 we compare our outcomes with the results obtained in [191] and [6][3]. When `FFTree` is train using $\delta = 0.2$ is competitive with Kamiran et al. in terms of performance, and provide a lower discrimination. Moreover, when `FFTree` is train using $\delta = 0.05$, the best result in terms of discrimination is achieved together with the lowest result in terms of performance. These results are consistent with the mitigation logic of the algorithms under examination; indeed, the fair trees in literature are aim to reduce discrimination by optimising performance, while FFTree optimise performance complying with the fairness constraint.

[3]Aghaei et al.’s results are approximate as a graph is shown in their paper without indicating exact values. As for `FFTree` we show the median result in terms of accuracy.

Table 6.3 Accuracy and demographic parity of different state-of-the-art fair tree on the *Adult* dataset. FFTree is trained with $FC$ = DP.

| Methods | DP | Accuracy |
|---|---|---|
| Kamiran et al. | .226 | **.839** |
| Zhang and Ntoutsi | .163 | .818 |
| Aghaei et al. | .060 | .815 |
| FFTree with $\delta = 0.2$ | .186 | .828 |
| FFTree with $\delta = 0.05$ | **.049** | .801 |

The following experiments show applications where FFTree is the only *fair tree* that can be employed thanks to its flexible novel functionality.

### 6.4.2 Disparate Mistreatment Mitigation (DM)

We conducted this experiments to verify FFTree's ability to mitigate disparate mistreatment by selecting EO, PP, and PE as fairness constraints, through the parameter $FC$. To validate the results we expect to obtain levels of accuracy in line with those obtained when trying to satisfy demographic parity while maintaining low levels of the respective fairness metric.

Tab.6.4 shows the results obtained with these mitigation criteria using the *adult* dataset and the real dataset on *loan granting*. In the latter dataset, the starting level of DP for the sensitive attribute of *citizenship* was 16.3%. A classic Random Forest classifier is able to replicate the loan officer's decision with an accuracy of 82%, while a standard tree classifier shown an on average reduction in performance of 7%. This result clearly indicates the cost in terms of performance of obtaining transparent rules. The performance losses increase further when fairness constraints are imposed.

For both datasets the greatest reduction in performance occurs when trying to achieve independence, while this is not necessarily true for separation and sufficiency. These results are in line with the *conservative* nature of these criteria.

Table 6.4 Accuracy and fairness of FFTree on the Adult and Loan Granting datasets (*FC* represents the selected fairness criterion).

| | Adult | | Loan Granting | |
|---|---|---|---|---|
| **Fairness Criteria *FC*** | ***FC* Value** | **Accuracy** | ***FC* Value** | **Accuracy** |
| Demographic Parity | 0.074 | 0.813 | 0.094 | 0.692 |
| Predictive Parity | 0.004 | 0.844 | 0.086 | 0.732 |
| Equal Opportunity | 0.035 | 0.829 | 0.065 | 0.721 |
| Predictive Equality | 0.013 | 0.821 | 0.047 | 0.711 |

Table 6.5 Accuracy and demographic parity of FFTree on the Adult and Loan Granting datasets for fixed $FC = DP$ and varying $\delta$.

| | Adult | | Loan Granting | |
|---|---|---|---|---|
| $\delta$ | **DP** | **Accuracy** | **DP** | **Accuracy** |
| 0.05 | 0.049 | 0.801 | 0.042 | 0.647 |
| 0.10 | 0.074 | 0.813 | 0.094 | 0.692 |
| 0.15 | 0.107 | 0.824 | 0.081 | 0.711 |

### 6.4.3 Business Needs (BN)

An important feature of FFTree is the possibility of defining the maximum level of discrimination allowed in each split in order to meet different business requirements or regulation constraints.

With this experiments we want to prove that changing the $\delta$ parameter is possible to obtain different (valid) trade-off between fairness and performance on both the benchmark and the real dataset.

Table 6.5 presents some interesting results: as expected, an increase in the allowed level of discrimination $\delta$ means an increase in performance, but it is not always true that an increase in performance means a decrease in fairness. In fact, for the real dataset, we observe a decrease in discrimination when $\delta$ is increased from 0.1 to 0.15. This phenomenon occurs because FFTree searches for the best solutions locally, and it is possible to find better global solutions, in terms of performance and fairness by relaxing the fairness constraint. These results suggest that it is appropriate to try different $\delta$ configurations to decide on the best trade-off for the application domain.

### 6.4.4 Multiple Fairness Criteria (MD, MS)

Finally, we carried out a set of experiments to investigate the ability of our model to implement multiple fairness constraints at the same time. Point D in the left scatter-plot of Figure 6.2 shows a configuration of FFTree in which the Demographic Parity of gender and race needed to be satisfied simultaneously (MS as in Tab.6.1). Both metrics were met at a marginal level with a small loss in performance compared to other models with only one constraint.

Similar results were obtained in the point G of Figure 6.2, where the aim was to satisfy *independence* and *sufficiency* criteria at the same time (MD as in Tab.6.1). As discussed in Section 6.3.2, the results arise from the dependencies between the data. The same experiments conducted on the real loan granting dataset, indicated a greater reduction in performance.

In subjective contexts, where companies try to support human decisions with automatic decision systems, such as the granting of credit, the loss in accuracy is not always so important. In fact, the previous human decisions, used to train the model, may not be consistent with each other when taken by different decision makers. Furthermore, these decisions may contain historical and selection biases [128]. Regardless of performance, fair and transparent rules, such as those provided by FFTree , can be successfully adopted if they are considered trustworthy by the business acumen of a human domain expert.

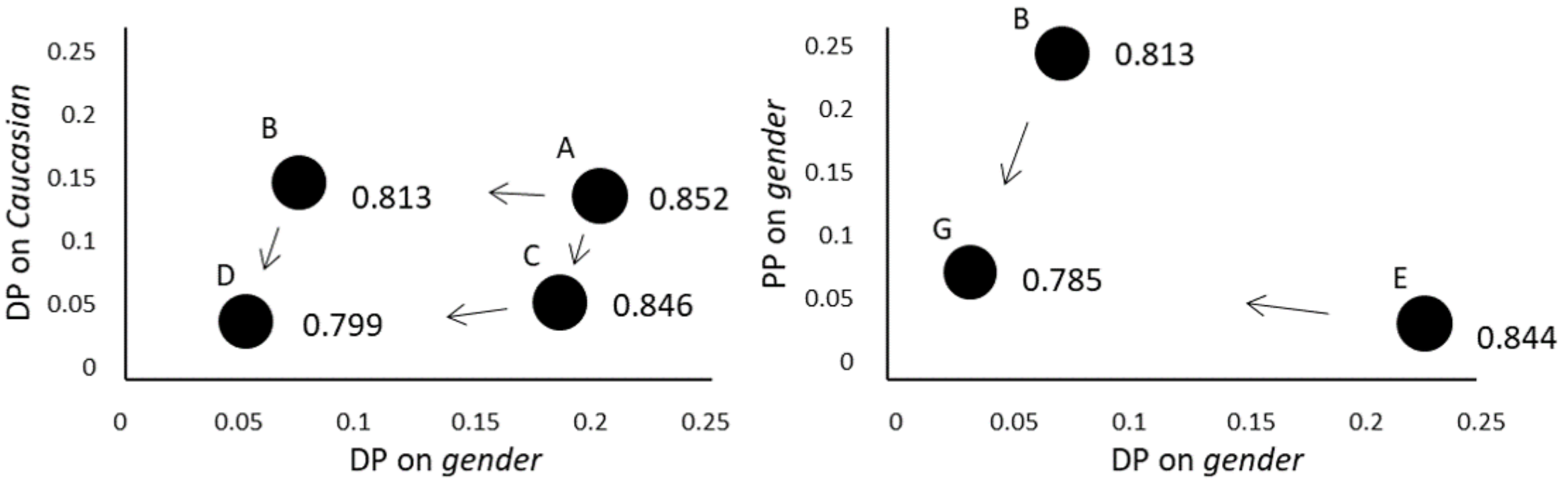

Figure 6.2 The following configurations of FFTree are represented: (A) unmitigated FFTree; (B) FFTree with FC = DP and S = gender; (C) FFTree with FC = DP and S = Caucasian; (D) FFTree with FC = DP and S = gender and Caucasian; (E) FFTree with FC = PP and S = gender; (G) FFTree with FC = DP and PP and S = gender. The number next to the bubble indicates the accuracy. The models were trained on the *Adult* dataset.

# Part III

# Accounting for Bias

# 7

# Addressing Fairness in the Banking Sector

Currently, policymakers, companies, and academic institutions are making efforts towards establishing guidelines and recommendations for Ethics in AI, which includes fairness in ML. One such initiative is AI4People[1], a multi-stakeholder forum in Europe with global activities around the promotion of a "good AI society". Within this context, as active members of AI4People, Intesa Sanpaolo and Fujitsu Laboratories of Europe are collaboratively engaged in an open innovation project. This endeavor demonstrates a proactive commitment to participating in initiatives aimed at addressing bias within industrial environments and advocating for the responsible development of AI-based systems. Specifically, in the subsequent sections of this chapter, we present a generic roadmap [35] for fostering fairness in machine learning, a result of this collaborative effort, which can be effectively applied across diverse banking use cases.

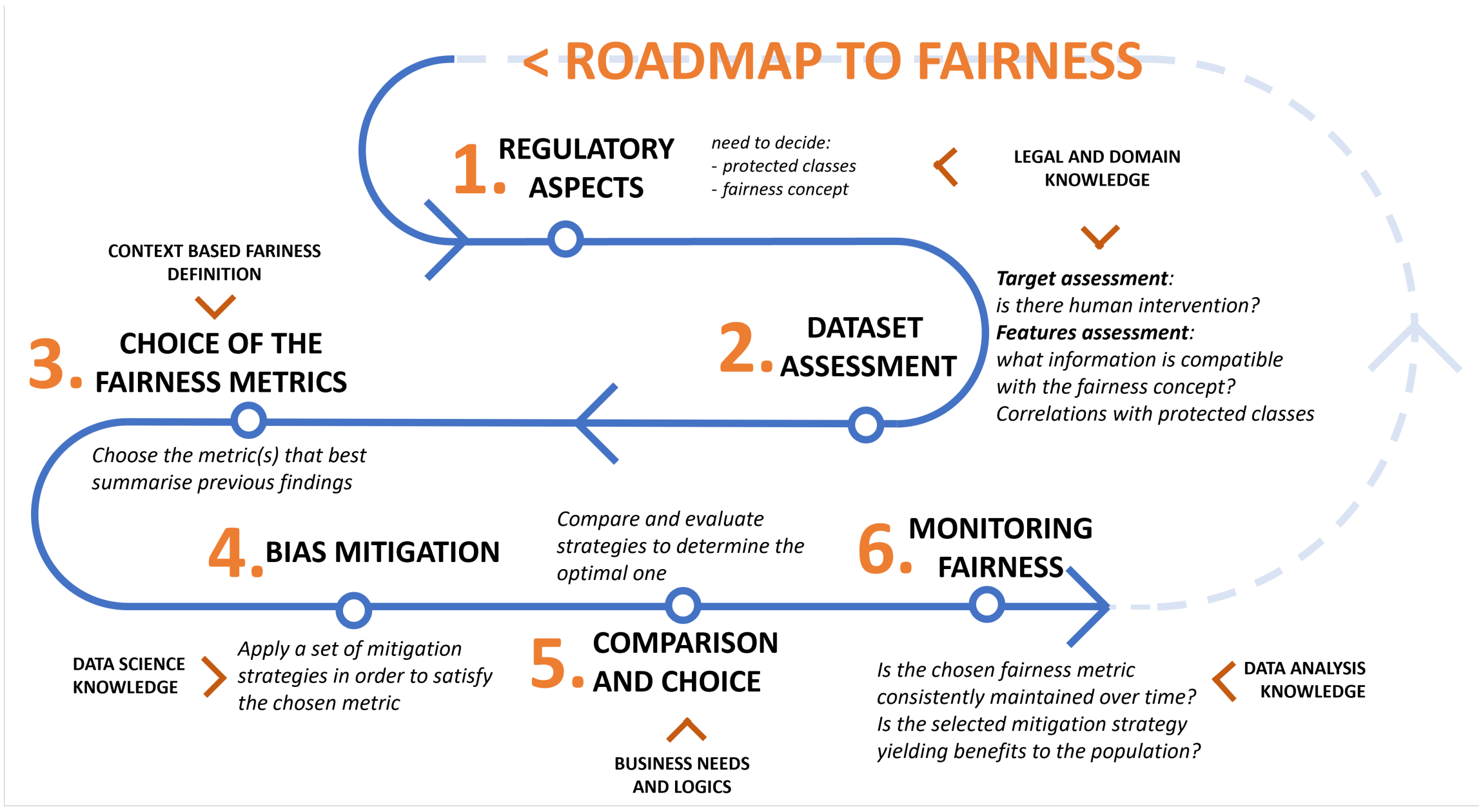

Figure 7.1 Schematic visualization of the proposed roadmap representing the process of pursuing fairness in ML projects.

## 7.1 Roadmap to Fairness

In this section, we present our proposal for a generic roadmap to enable fairness in ML. It encompasses six states: regulatory aspects, dataset assessment, choice of fairness metrics, bias mitigation, comparison/evaluations, and monitoring.

It is notable that the roadmap requires specific inputs from various expertise (such as legal and domain knowledge, expert knowledge, etc.). Indeed, pursuing fairness is a process far too intertwined with several ethical and social aspects to be treated as a purely technical issue.

As a final remark, this roadmap must be thought of as a flexible guideline, with steps back and iterations over specific points, in order to converge to the best possible solution between regulatory aspects, mathematical formulation, algorithmic performance, and fairness optimization. The process is at least as important as metric optimization.

The roadmap is outlined in the following steps (see Figure 7.1):

[1]AI4People, https://www.eismd.eu/ai4people/

### 7.1.1 Regulatory Aspects

[Domain knowledge, legal expertise] As we have seen in Chapter 4 there is no single notion of fairness and its definition is highly dependent on specific aspects of the use case at hand and of its domain. In order to decide what is the potentially *sensitive information* and the *concept of fairness* to be pursued, it is necessary to take into account legal and regulatory aspects.

The importance of this direction is unequivocally demonstrated by the European Union's proactive efforts to regulate AI, aiming to create a more favorable environment for the development and deployment of this innovative technology. In April 2021, the European Commission introduced the First Proposal for a "Regulation laying down harmonised rules on artificial intelligence" (Regulation laying down harmonised rules on AI (AI Act)) [169]. The European Parliament passed this landmark legislation in June 2023, bringing the bloc one step closer to formally adopting the world's first major set of comprehensive rules regulating AI technology. The AI Act, firmly rooted in EU values and fundamental rights, seeks to foster trust in AI-driven solutions while promoting their responsible advancement by businesses. In the US, the White House Office of Science and Technology Policy has recently published a Blueprint for an AI Bill of Rights with five overarching guiding principles for AI systems design, development, and use, one of which is titled *"Algorithmic Discrimination Protection"* [181]. Alongside, binding legislation has been proposed, now under discussion [136] —the *Algorithmic Accountability Act*. The UK government published in July 2022 an AI Regulation Policy paper in which one of the pillars is *"Embed considerations of fairness into AI"* [58], while more advanced and binding legislative proposals are still to come.

However, the fact that, sooner or later, there will be regulations to clearly prescribe what is and what is not fair in each situation is unlikely, unreasonable, and in many ways undesirable. Thus, it is crucial that in each domain and even use case, people working on it, being them developers, scientists, or domain experts, consider carefully the consequences of including or not potentially sensitive attributes and the variables correlated to them and what could mean, from the point of view of the final user, to be unfairly discriminated, of course taking into account all the regulatory aspects relevant to the specific situation. In any case, the advent of regulations is itself a (slow) process, during which companies and service providers can play a role by actively doing research and building their own policies and best practices.

### 7.1.2 Dataset Assessment

[Domain knowledge, legal expertise]

In situations where a clear legal requirement regarding the specific fairness metric is absent, the identification of the most appropriate metric for the available data becomes a critical task. Once sensitive variables have been defined, it is essential to evaluate which information within the dataset can effectively contribute to generating fair decisions. The evaluation process encompasses two fundamental aspects: the target variable and the feature variable designated for the subsequent training of the machine learning system.

**Target Variable.** The nature of the ground truth variable, whether it results from human judgment or factual information, significantly influences its suitability for measuring fairness. Metrics such as Equality of Odds heavily rely on the ground truth as a target to quantify fairness and cannot be used if the target itself is prone to some form of bias. Notice that, in the credit lending example, even a variable that may seem objective, such as the actual repayment or not of a loan, is subject to a form of selection bias, since it is information available only for people who were granted a loan in the first place and that are customers of a single bank [183].

In Chapter 8, we will present `FairView`, a tool designed for evaluating the target variable and assisting practitioners in determining the appropriate fairness metric.

**Feature Variables.** Within the feature variables, the consideration is directed toward information that aligns with the objectives of the AI-based decision-making system. For instance, in the credit lending use case, the correlation between a sensitive attribute like gender or citizenship and income could be considered. Income is undeniably a pivotal factor for determining repayment probabilities; thus, even if it disproportionately affects decisions between genders, it might be deemed fair.

Conversely, one may consider that the differences in income (e.g. with respect to gender) are themselves due to historical bias and thus opt for the use of techniques to remove the dependence between income and the sensitive attributes, in order to mitigate this historical bias as well (see e.g. [183] and references therein for more insights). When opting to utilize a variable correlated with the sensitive attribute for algorithm training, it is crucial to provide an objective and reasonable justification. Objective justifications align with anti-discrimination laws and are an essential consideration in this context [160].

### 7.1.3 Choice of the Fairness Metric(s)

[Domain knowledge, data ethics expertise, data science expertise] Given the outcome of steps 1 and 2, it is possible to choose, among available fairness metrics, the one(s) that best embodies the chosen fairness concept, given the dataset assessment. In particular, some choices to be made are:

- target dependent / target independent;
- group / individual;
- observational / causal.

Depending on the use case, one may decide to monitor more than one metric, e.g. both a group and an individual notion of fairness.

In many respects individual and group fairness can be thought of as two extremes of a continuum of possible metrics, roughly depending on what kind of variables one is willing to accept on the basis of the chosen concept of fairness. Namely, one could condition over all the non-sensitive variables, thus enforcing a form of Conditional Demographic Parity which is equivalent to simply removing the sensitive feature only; or one could not condition at all, thus enforcing the group notion of Demographic Parity, e.g. by removing all the information of the sensitive feature present in the dataset (i.e. the variable itself and all its correlations with other variables). Intermediate forms of Conditional Demographic Parity lie between these two extremes.

When the alignment between available fairness metrics and the specific context remains elusive after Steps 1 and 2, ethical considerations become the compass steering the course. Within this intricate landscape, the role of data ethics assumes a central role, taking on the responsibility of providing a structured framework to justify decisions.

At the heart of the third stage of this roadmap lies a crucial objective. It goes beyond the mere selection of a fairness metric for the subsequent phase. Instead, it encompasses a robust justification grounded in regulatory compliance, concrete data evidence, and ethical consideration – forming the foundation for the company's accountability. This comprehensive justification not only guides decision-making but also provides a credible and defensible basis. These three stages ensure that the chosen path isn't merely a metric selection but a well-informed, responsible decision backed by a strong framework.

### 7.1.4 Bias Mitigation

[Data science expertise] Different strategies can be implemented in order to fit a model by pursuing both algorithmic performance and an optimal value of the chosen metric. As discussed in Chapter 5, these strategies are usually classified into pre-processing, in-processing, and post-processing techniques, depending on the specific point of the algorithmic pipeline in which fairness optimization is implemented.

The selection of appropriate models for implementation is heavily contingent upon the constraints established in prior stages. These constraints encompass the necessity for respecting the chosen fairness metric, the identification of sensitive variables to be considered for mitigation, and the permissible variables for training the Machine Learning model. Moreover, a multitude of additional domain-specific factors necessitates consideration, including the nature of the problem (e.g., regression, binary classification), the type of data (e.g., textual, image, tabular), and the desired level of interpretability demanded by the end-user of the model.

### 7.1.5 Comparison of the Strategies and Evaluation

[Domain knowledge, data science expertise] Upon the development of a set of mitigation strategies, the subsequent step entails a rigorous and methodical comparison and evaluation process. This comprehensive assessment considers both algorithmic performance and alignment with chosen fairness metrics. The aim is to identify the most effective strategy that balances efficiency and fairness.

Evaluating algorithmic performance involves analyzing how each strategy impacts accuracy, efficiency, and robustness across various data scenarios. Simultaneously, the chosen fairness metrics are rigorously evaluated to gauge each strategy's bias mitigation effectiveness.

In this stage is important to foster collaboration between data scientists, domain experts, and stakeholders. The holistic approach ensures an optimal strategy selection that enhances predictive capabilities and builds trust in AI-driven decision-making.

### 7.1.6 Monitoring Fairness

[Domain knowledge, data analyst expertise, governance, legal expertise]

The journey towards fairness doesn't conclude with the implementation of an optimal mitigation strategy —- it's an ongoing commitment. Establishing a system of checks and balances is imperative to prevent biases from taking root once again.

A fundamental aspect of this continuous commitment involves establishing a robust system for monitoring and governing fairness. These mechanisms act as safeguards, preventing biases from reemerging and promptly alerting when fairness boundaries are breached. Central to this approach is the establishment of predefined thresholds, carefully calibrated to encompass acceptable deviations from established fairness standards.

Anticipating and planning for corrective interventions is a crucial element of this framework. By considering potential corrective measures in advance, organizations can respond effectively to unforeseen deviations. This governance structure should encompass a range of calibrated actions, each tailored to the severity of the situation. From temporarily suspending AI applications in sensitive contexts to re-train algorithms, this spectrum of responses ensures a resilient approach to upholding fairness. In Chapter 9 we will explore a real application involving the monitoring of fairness over time.

In addition to monitoring fairness, ensuring continuous compliance holds significant importance. Companies need to stay informed about evolving AI regulations and ensure the alignment of all AI-based solutions with these regulations. This highlights the critical governance role that companies play in maintaining responsible AI practices. They must be adept at recognizing deployed AI decision-making systems that fall short of new regulatory constraints and plan for their evolutions by revisiting the initial point of this roadmap.

## 7.2 Discussion on the Roadmap

Fairness has many dimensions. We have seen that several techniques are available both to assess and to mitigate them. However, it is still not clear which dimension should be pursued in each specific situation.

Our roadmap is an attempt to provide a general guideline to address fairness in ML projects and focuses on the fact that different expertise should work together along the process in order to properly take into account the context and the social impact of the technological service/product being developed.

Our `BeFair` toolkit, presented in section 5.6 allows data scientists and developers to embed several bias mitigation techniques and assessment metrics within their ML projects and to compare these with a rationale of fairness/performance trade-offs. These can be used to eventually make practical decisions with respect to the aforementioned roadmap.

We believe that more research is needed on the ethical and legal side to disentangle and explicitly elaborate on various categorisations of bias that concur to form the overall discrimination in specific domains. It will facilitate understanding the information that can be safely exploited in different situations.

# 8

# `FairView`: an Evaluative AI Support for Addressing Fairness

As emphasized throughout this thesis, and particularly in the preceding chapter, the greatest challenge for companies is deciding which definition is suitable for each context of AI systems application so that it can be held accountable. It is necessary to choose because, under certain conditions, the different conceptions of fairness are mutually incompatible [13]. Furthermore, it is important to note that various mitigation strategies yield distinct outcomes, influencing the system's population differently.

Deciding which notion of fairness is more appropriate and understanding whether an AI system behaves fairly in decision-making requires extensive contextual understanding and domain knowledge [197]. In addition to the roadmap presented in the previous chapter, the literature has introduced other checklists [123, 1, 178] and guidelines [162, 96, 161, 126] to assist companies in responding to this difficulties and to be able to identify the more suitable fairness framework for each application domain.

An important and common question to answer when deciding on the proper concept of fairness is the reliability of the target variable that the model has to replicate (as in the *dataset assessment* step of our roadmap). In particular, building on the moral framework discussed in Chapter 2, it is fundamental to understand if that target variable quantified in the OS is objective [149], i.e. faithfully represents the real phenomenon to be predicted (contained in CS), or may already be distorted by some form of bias (see, e.g. [126]). For example, in a hypothetical situation where it is known that there is direct discrimination in the target variable, the WYSIWYG moral framework is hardly acceptable and, consequently, the implementation of target-based mitigation strategies. On the other hand, assuming WAE in such a context would permit not relying blindly on the data contained in OS and implementing mitigation measures so that the decision-making process is more congruent with the real phenomenon. Unfortunately, a concrete and sure understanding of whether the target variable is reliable or contains some form of bias remains a major issue, especially if it results from a human decision.

In this chapter, we present `FairView` [38], our approach that aims to support the choice of which worldview to assume when addressing fairness in developing an AI system. This approach is based on evaluative AI, a new paradigm in explainable AI proposed by Miller [133], which generates evidence that supports human judgments by explaining trade-offs between any set of possibilities instead of making recommendations. In our situation, the options available are the different moral frameworks on which grounding the fairness metric to adopt in a specific decision-making process, and the hypotheses that will serve as the foundation for the evidence come from potential distortions between the various metric spaces, such as the presence of direct discrimination in the observed target variable. To show these potential distortions, we propose to leverage global contrastive explanations. By highlighting the presence - or the absence - of potential discrimination in the observed data, we provide actionable insights to developers and decision-makers, allowing them to properly decide which worldview - WAE, WYSIWYG, and nuances in between - may be preferable to assume when an AI-system operates in a specific context. This work represents one of the first attempts to link XAI with moral frameworks with which the ethical-philosophical literature formalises fairness. In addition, a lack of context-specific frameworks that support companies in making high-stakes decisions can be recognised, and this work aims to offer a tool for further research in this area. The same [133] argues that evaluative AI may be specifically helpful in cases of high-stakes decision-making processes.

## 8.1 Related Works on Contrastive Explanation and Evaluative AI

`FairView` investigates fairness by exploiting contrastive explanations [115], following the evaluative AI paradigm [133]. The motivation behind contrastive explanations is that people usually understand the reason for an event in relation – or contrast – to some other event that did not occur rather than explaining the causes for that event [132]. This means a human explanation is more likely to answer a question like "Why P rather than Q?" instead of "Why P?" even though Q is often implicit in the context. This is called contrastive explanation. However, people usually expect to see a specific event happen while another occurs, making the observed event the fact and the expected event the foil [174]. In particular, ContrXT [127] inspires our approach, which follows this principle. ContrXT (**Contr**astive e**X**plainer for **T**ext/**T**abular classifier) is an approach that utilizes Binary Decision Diagrams (Binary Decision Diagrams (BDD)) to trace the decision criteria of a black box text or tabular classifier. It achieves this by training surrogate decision tree models at different temporal stages, which serve as global post-hoc explanations of the logic behind the black box model. ContrXT provides global, model-agnostic, and time-contrastive explanations by encoding the decision logic changes through BDDs. These explanations show why the model has modified its behaviour over time, allowing for a deeper understanding of the new factors influencing its decisions. The decision trees employed in ContrXT play a crucial role as they capture the decision-making process of the black box model and provide comprehensive explanations of its logic. These global post-hoc explanations enable users to understand how the model arrives at its decisions, enhancing transparency and interpretability.

In our approach, we shift from comparing the behaviour of machine learning classifiers trained at different time intervals to compare the behaviour between different sensitive groups using tabular data before training a machine learning model. By examining the behaviour across different sensitive groups, we aim to gain valuable insights that can support selecting the appropriate moral framework to address fairness concerns based on the observed data.

However, the contrastive explanation typically gives a recommendation and justifies it [133]. This persuasive approach does not align with the human cognitive process [85], as it would be desirable when human decision-makers are ultimately accountable. A more effective approach for high-stakes decisions can be found in the support provided by evaluative AI [133]. This is a machine-in-the-loop paradigm in which decision support tools provide evidence for and against decisions made by people, rather than provide recommendations to accept or reject. Recent works, such as [37], demonstrate that Evaluative AI is compatible with contrastive explanations. FairView is a decision support tool that assumes this innovative paradigm in explainable AI: it does not provide recommendations. Instead, it allows the decision-makers to gain evidence for the trade-offs between moral worldviews to determine which is more suitable for addressing fairness given a specific observed space.

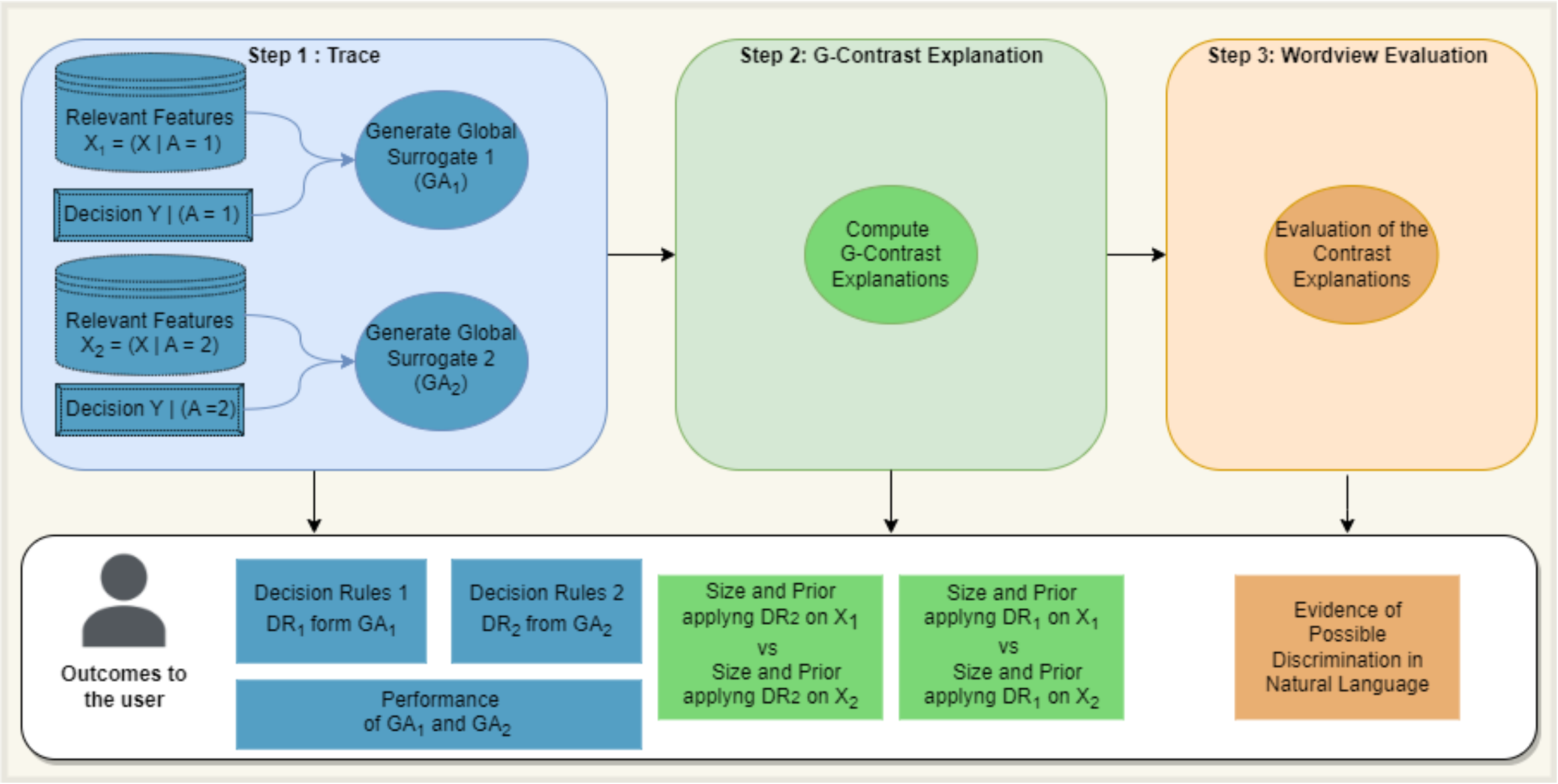

Figure 8.1 Graphical overview of how FairView works with a binary sensitive feature $A$.

## 8.2 FairView Formalisation

FairView is designed to support users in the pre-development phase of AI-based decision-making systems. In this context, a key challenge arises in determining the most suitable fairness mitigation approach, which carries significant accountability implications. To tackle this challenge, FairView permits to evaluate the consistency of the hypotheses aligned with the WAE and WYSIWYG principles in the data. In particular, by leveraging group contrastive explanation techniques, FairView offers evidence to uncover potential instances of direct or indirect discrimination within the observed data.

Direct discrimination refers to treating somebody unfavourably because of their membership to a particular group, characterized by sensitive attributes, such as gender or race [77]. Formally, when a decision $Y$ depends on relevant attributes for the problem $X$ and sensitive attributes $A$, i.e. $Y = f(X,A)$. In this case, two individuals $a$ and $b$, with the same relevant attributes ($X_a = X_b$), but belonging to different sensitive groups ($A_a \neq A_b$), could obtain different decisions ($Y_a \neq Y_b$). Indirect discrimination arises when the decision $Y$ depends only on the relevant attributes $X$, but for some reason, $X$ is not independent of $A$, i.e. $Y = f(X)$ *with* $X \not\perp\!\!\!\perp A$. In this case, $a$ and $b$ in the previous example would have the same decision ($Y_a = Y_b$). Still, since the relevant attributes $X$ used by the decision-making process are not distributed in the same way between sensitive groups (e.g. males have, on average, higher salaries than females), the percentage of positive decisions varies among sensitive groups ($P(Y = 1|A = 1) \neq P(Y = 1|A \neq 1)$). Although both situations result in a non-independence between the decision $Y$ and sensitive attributes $A$, indirect discrimination under certain conditions could be tolerable, i.e. assuming a WYSIWYG worldview, and compliant with the principle of objective justifications [160].

Understanding whether the decision $Y$ results from direct or indirect discrimination is crucial when it serves as the target variable for further training the machine learning model. Moreover, training a machine learning model on $Y$ without using $A$ is insufficient for removing the effects of direct discrimination because of the latent effect of $A$ on $X$ [77]. However, in cases where this target variable exhibits a correlation with sensitive attributes, it remains unclear whether this correlation arises from direct or indirect discrimination. FairView helps users understand the underlying factors that influence the observed decision outcomes by thoroughly examining the data and identifying relevant patterns. In particular, users may come across patterns that result in different proportions of favorable outcomes between population subgroups. This observation can indicate that the sensitive attribute was utilized in the decision-making process, implying the presence of direct discrimination. Alternatively, it may suggest that other relevant information not present in the available data was considered. In both cases, these patterns serve as evidence of distortion between the observed space and the construct space, supporting the foundational hypothesis of the WAE moral framework. On the other hand, when FairView identifies patterns that do not exhibit differences in treatment between population subgroups, it provides evidence that does not contradict the foundational hypothesis of the WYSIWYG moral framework.

For obtaining these evidences, FairView is designed with three fundamental steps: (i) trace, (ii) G-contrast explanation, and (iii) worldview evaluation (see Fig.8.1).

In the **trace** step, a decision tree is trained within each class of the sensitive feature $i \in A$, using the features relevant to the problem $(X|A = i)$ to learn how to reproduce the decision $(Y|A = i)$. The resulting decision trees serve as a surrogate transparent explanation of the logic used to classify the element of a given population group. At the end of this phase, a set $j$ of decision rules $DR_{ij}$ for each sensitive group is encoded via a global surrogate $GA_i$ that mimics their real decision-making process. A decision rule is a series of minimum requirements, such as $income > z$ & $guarantees > g$ that lead to a positive decision. The decision to train as many decision trees as there are sensitive classes in the relative subsamples, rather than a single one on the entire dataset, is made to ensure that the logic used to classify the minority classes is not ignored or overshadowed by the majority classes. By focusing on separate subsamples for each sensitive class, this step aims to provide a more nuanced understanding of the decision-making process and identify potential decision rules that might be specific to certain sensitive groups. In the **g-contrast explanation** step, FairView first applies the decision rules $DR_{ij}$ of each surrogate $GA_i$ on each set of relevant features $(X|A = i)$ to obtain their **size** (the number of instances that respect the decision rule from $GA_i$) and **prior** (the percentage of instance that respect the decision rule from $GA_i$ and have a positive outcome). For example, suppose that the sensitive attribute is composed of two classes: female ($A = 1$) and male ($A = 2$), and one of the decision rules, learned by training the decision tree on $X|A = 1$ to learn when $Y|A = 1$ is, as before, $income > z$ & $guarantees > g$. The g-contrast step applies this decision rule both on $X|A = 1$ and $X|A = 2$ to calculate how many females and how many males respect the rule (size) and the percentage of how many females and how many males respecting the rule have a positive target variable (prior). Each decision rule is contrasted among sensitive groups to obtain its size and prior. It is important to note that poor performance of the surrogates used as the basis for these explanations may indicate that the available data $X$ are insufficient to reproduce the logic used to make the decision $Y$ in question, and a user should consider whether it is appropriate to proceed with the development of more complicated models without first enriching them.

Finally, in the **worldview evaluation** step, FairView provides users with a natural language outcome based on the results of the g-contrast explanation, which supports the underlying worldviews. Specifically, decision rules with a low level of difference in prior are presented as evidence in favor of indirect discrimination toward the WYSIWYG framework, while decision rules with a high level of difference in prior are presented as evidence of direct discrimination toward the WAE framework. Users can customize the threshold at which they consider the difference in prior to be significant. The size associated with each decision rule indicates the magnitude of evidence supporting a particular worldview. By analyzing decision rules supporting WAE and WYSIWYG, decision-makers can be facilitated in deciding which type of fairness mitigation to implement. Additionally, they can repeat the analysis by assessing a new target variable and input data.

### 8.2.1 `FairView` as a Open Source Tool

FairView has been implemented through Python as a pip package, publicly available to the community on GitHub. The input consists of each sensitive group's training features $X$ and classification decision $Y$. It is possible to use either Scikit-learn [146] decision trees or the RuleFit algorithm [73] for generating surrogates. Users also have the option to specify the threshold that determines the significance of the difference in priors and the hyperparameters that constitute the complexity of the surrogate, such as the maximum number of decision rules (leaf nodes) in the case of a Decision Tree (DT). Our GitHub repository contains a

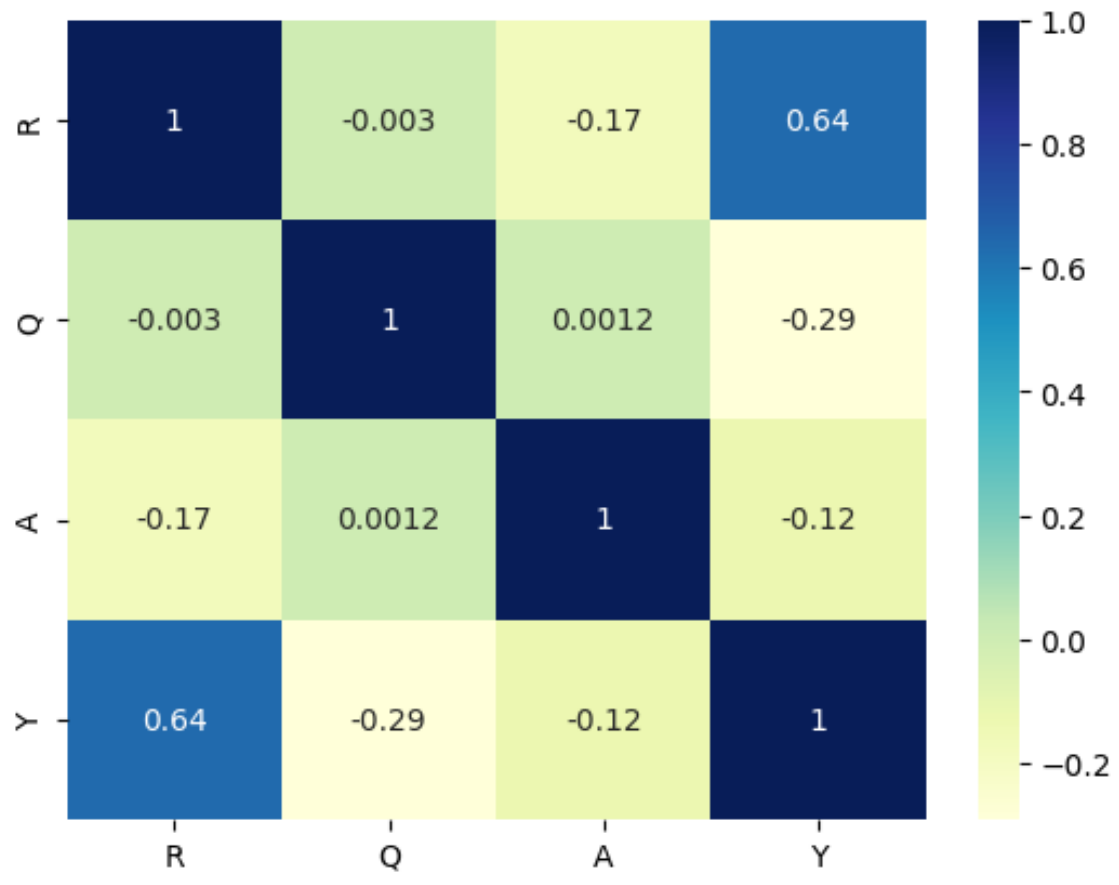

(a) Correlation Matrix on the Dataset with Historical Bias.

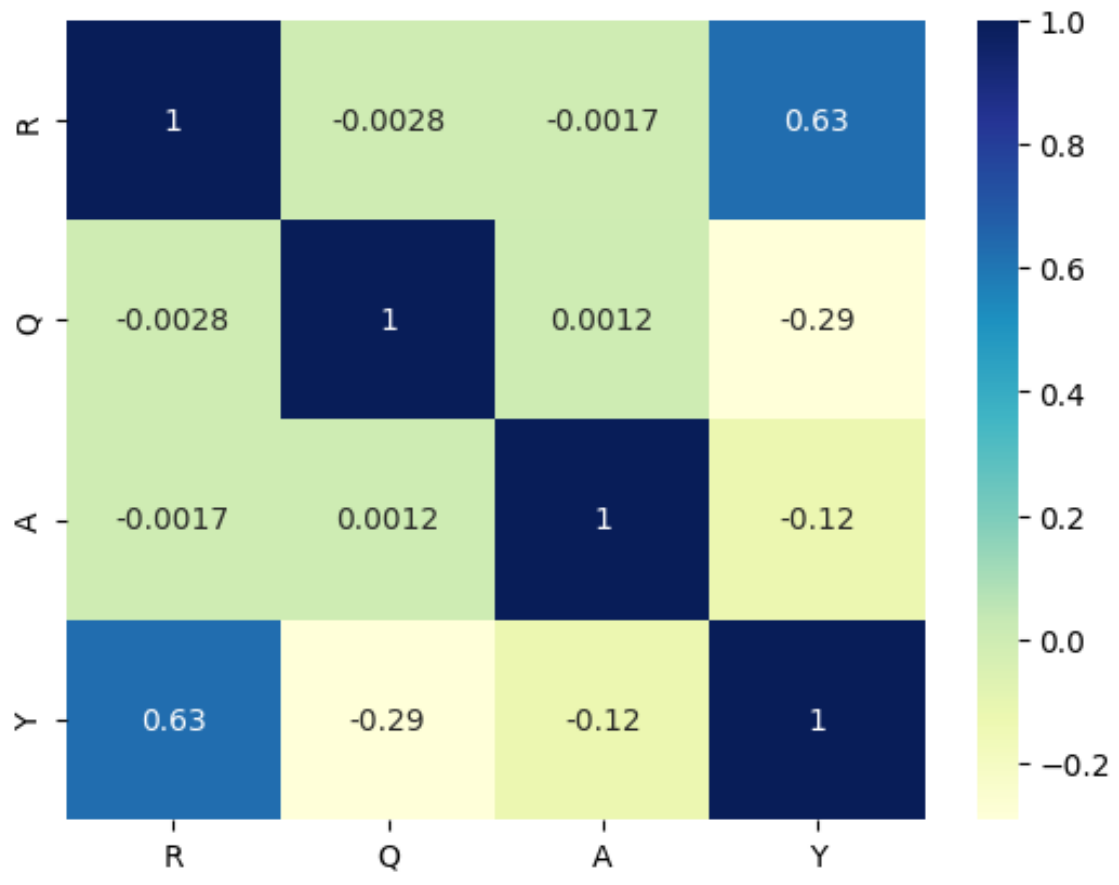

(b) Correlation Matrix on the Dataset with Measurement Bias.

comprehensive guide detailing the software usage and illustrative example notebooks to facilitate understanding and implementation.

The following section serves as an example to provide a clearer understanding of how FairView works.

## 8.3 `FairView` Showcase on Synthetic Dataset generated by `Bias On Demand`

We showcase a simple initial application of FairView to two synthetic cases generated using `Bias On Demand` [15], the model framework for generating synthetic data with bias presented in Chapter 2. In particular, we generate ($a$) a dataset with *historical bias* on a relevant input feature $R$ and; ($b$) a dataset with *measurement bias* on the target variable $Y$[1]. The first can be seen as a case of indirect discrimination, while the second is a case of direct discrimination. Both datasets correlate $A$ and $Y$, as shown in Fig.8.2a and 8.2b, making it difficult to distinguish between the two biases without prior knowledge of the process that generated them.

This application aims to demonstrate that with FairView, it is possible to distinguish the two cases and therefore decide the proper worldview to adopt in these two cases. We set the difference in prior being significant when greater than 5%.

[1]Historical bias is generated using $l_{hr} = 1.5$. Measurement bias is generated using $l_{my} = 1.5$

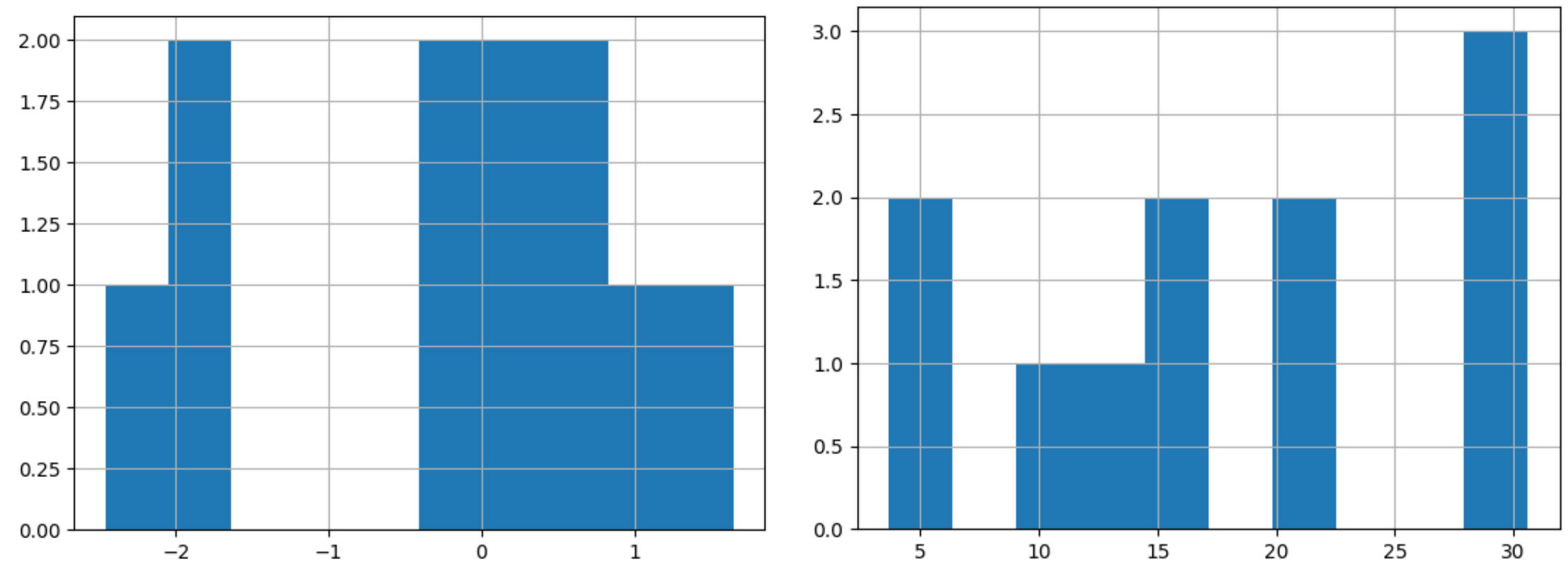

(a) Frequency of the difference in prior of each decision rule between groups of $A$ with Historical Bias.

(b) Frequency of the difference in prior of each decision rule between groups of $A$ with Measurement Bias.

*a) Results with historical bias on $Y$.* Since the sensitive feature $A$ is binary, two decision trees are trained in the Trace step. The first, trained when $A = 0$, has learned six decision rules with 85.2% *accuracy* and 85.6% *F1*, while the second, trained when $A = 1$, has learned five decision rules with 86.8% *accuracy* and 86.9% *F1*. In the G-contrast explanation step, the rules are contrasted between groups of $A$ by observing the difference in priors. Fig. 8.3a summarises these differences. Since the differences are all between -5 and 5, the *worldview evaluation* step suggests following only WYSIWYG.

*b) Results with measurement bias on $Y$.* The trace step is analogous to the case with historical bias, with a similar number of decision rules learned and performance achieved. What changes are the results in the G-contrast explanation step: Fig. 8.3b shows that all decision rules have differences in prior greater than 5%, and three decision rules have a deviation up to 30%. As a consequence, the *worldview evaluation* step suggests that it is preferable to enforce a WAE worldview.

The FairView evaluation in both cases results be consistent with the nature of the types of bias in the datasets. Indeed, as clarified by Friedler et al., measurement bias is a distortion between CS and OS that is recognised by a WAE worldview since it assumes equality among groups in CS and requires mitigating any differences in CS in the name of fairness between individuals in PS. On the other hand, historical biases impact CS, which is then fairly reported in OS. Therefore, if a WYSIWYG worldview is assumed, structural differences between individuals are legitimate sources of inequality.

## 8.4 Discussion on FairView

**On the G-contrast Step in the Landscape of Fairness Metrics.** The prior comparison in the G-contrast step regarding fairness metrics can be seen as a form of stronger conditional demographic parity. Building upon the discussion in Chapter 4.4 regarding the comparison between group vs individual fairness, CDP compared to DP is one step further towards a more individual notion of fairness. Our approach checks if the decisions are independent *given* a specific decision rule, taking another step towards individual fairness. Additionally, it addresses the challenge faced by CDP in automating the selection of the variables on which imposing conditionally equal treatment.

One might wonder why not directly employ individual fairness metrics. As discussed in chapter 4.2, the limitation of individual fairness lies in the intricate concept of "similar individuals". Designing an appropriate distance metric in the feature space to embody ethical-based similarity is nearly as complex as defining fairness itself. In comparison, our proposal provides a reasonable and easily understandable explanation of potential discrimination since it is computed following relevant patterns that synthesize the underlying decision-making process, offering a well-balanced trade-off between considerations of group and individual fairness.

**On the Worldview Evaluation step From a Philosophical Perspective.** Finding a pattern that results in different proportions of favourable outcomes between population subgroups, may indicate that the sensitive attribute was used to make the decision, i.e. that there is direct discrimination or that other relevant information not present in the available data was used to make the decision. Both conclusions represent evidence of distortion between OS and CS; consequently, fairness metrics that rely on the WYSIWYG worldview are hardly justifiable. For this reason, WYSIWYG is hardly preferable when there are differences in prior. WAE is the most appropriate moral framework when these differences cannot be objectively justified based on available data. Unfortunately, rarely in real cases are there situations as easily delineated as in the provided example. More frequently, a mix of patterns with varying degrees of different treatment. In these cases, observing the coverage of the population by a specific pattern can provide valuable insights for guiding the appropriate ethical direction forward.

In general, businesses often opt for WYSIWYG when there isn't substantial evidence undermining the reliability of the target [161]. However, there may be cases where WAE is followed because of strong moral beliefs that WAE in the potential space (see [84] for details) or for strategic reasons [90, 40], as we will discuss in the next chapter.

# 9

# Towards Fairness Through Time

Fairness definitions usually lack in considering the effect that the imposed constraints can cause on the disadvantaged groups [117]. Moreover, in many real-world scenarios, the attributes of the population can change over time, consequently, fairness cannot be treated as a static problem [191]. Some recent works that model fairness as a multi-stage game are worth mentioning [193, 137], since they consider the convergence between individual and group fairness as the *optimal situation*. In particular, [89] claims that group fairness helps to approach convergence with individual fairness in a long-term vision, while the opposite is not true, since *“group-level discrimination remains unchanged when subject to decisions that respect notions of individual fairness"*. Formalizing the dynamics of fairness is as important as monitoring the actually occurring effects in the population over time.

In this last Chapter of this Part on *accounting for* bias, we analyse a real-world scenario about credit lending where we assume to choose as a fairness metric the demographic parity with the goal of producing *long-term* benefits for the population.

## 9.1 Motivating Example

Imagine a forward-thinking bank that operates with a unique advantage – it has no limit to the amount of credit it can grant. The bank’s ultimate goals are to maximize its profits by strategically selecting clients who are likely to repay their loans (good clients) and attract as many good clients as possible. To achieve these objectives, the bank relies exclusively on financial status as the determining factor for loan approvals, without considering any demographic-sensitive information about the clients. This practice aligns with the principle of individual fairness, where clients are evaluated based solely on their financial qualifications, ensuring equal treatment for all applicants.

However, the bank starts noticing an intriguing trend. Clients from disadvantaged minority groups, who have historically faced challenges accessing affordable credit, often appear as bad clients based on their financial status alone. This is due to a life bias that has persisted over time, leading to lower financial statuses among these minority clients.

Upon closer examination, the bank realizes that this bias has created a cycle of disadvantage for these minority clients. Denied access to credit at reasonable rates, they have struggled to start businesses or invest in opportunities that could potentially improve their financial status. As a result, they consistently appear as bad clients based on financial metrics.

Recognizing the potential for positive change, the bank decides to employ the principle of demographic parity. By actively addressing the historical bias and granting loans to minority clients who may not meet the standard financial criteria but have the potential to thrive, the bank aims to break the cycle of disadvantage. This decision aligns with the bank's long-term vision of not only maximizing profits but also contributing to the financial empowerment of all individuals, regardless of their background.

Over time, the bank's approach proves to be successful. The loans granted to minority clients lead to the initiation of businesses and investments that stimulate economic growth within their communities. As these businesses flourish, the financial statuses of minority clients and their communities improve significantly. Consequently, what initially appeared as bad clients based solely on financial metrics now demonstrate the capacity to be good payers, thanks to the elimination of historical bias and the opportunities provided by the bank's fair lending practices.

In this context, employing demographic parity in credit lending not only contributes to individual financial success but also leads to long-term benefits for the bank and the broader society. By recognizing the potential of clients who have been historically disadvantaged, the bank not only maximizes its profits but also becomes a catalyst for positive social change and economic growth.

This simplified example aligns with the WAEPS worldview[84] introduced in Chapter 2 and supports the decision to mitigate using demographic parity as a strategic approach to simultaneously address group and individual fairness alignment.

## 9.2 Problem Setting and Relative Challenge Questions

Let us assume that a bank decides to develop a ML model to perform loan granting, to replace the previous human decision-maker. To build this model, the data scientists can access a dataset $D_{t_1}$ at time $t_1$ composed by a set of financial features $X_{t_1}$ and the citizenship $C_{t_1}$. During the designing phase emerges the impossibility of providing simultaneously at $t_1$ individual and group fairness due to the correlation between the citizenship attributes and some important financial variables such as *salary* or *rating*. In particular, data scientists can deploy a model $\psi$ that satisfies DP, or a model $\phi$ that satisfies individual fairness. Following the motivating example, let us assume that under this condition the bank decides to pursue a notion of group fairness to favor convergence over time towards the optimal situation.We refer to *optimal situation* as when it is possible to achieve simultaneously Demographic Parity and individual fairness.

Considering fairness as a dynamic game might enable the following challenging questions.

**CQ1:** Will the outputs of $\psi_1$ continue to ensure Demographic Parity over time?

**CQ2:** How can the model be retrained over time if it has completely replaced the human decision-maker?

**CQ3:** How can XAI techniques be used to verify that the chosen fairness policy (ensuring DP) is helping to reduce individual discrimination over time?

The aim of this work is to investigate novel ways to answer the previous *challenging* questions by conducting empirical experiments on real data owned by Intesa Sanpaolo accounting for 800,000 personal loan granting documents dated between 2016 and 2019. This dataset is an extension in terms of records collected compared to the one presented in chapter 5.6.1

## 9.3 Related Works on Fair Stream Learning and Fairness for Sequential Decision Making

Applied fairness works commonly consider static objectives defined on a snapshot of the population at one instant in time since the most used open datasets in the fairness domain [62, 11, 112, 155] do not contain data over several years. Reliability over time remains an open issue.

### 9.3.1 Fair Stream Learning

In stream learning, the distribution of the data can change over time: the so-called concept drift [110, 4, 75]. Therefore, the decision model should be able to adapt to the drifts through incremental learning from new occurrences (see, e.g. [192]). In particular, the model handles the drifts through (i) *informed adaption* - adapting only if it detects changes - or through (ii) *blind adaption* - updating constantly with the new data. In [93, 191, 190] the fairness constraints are taken into account together with the drifts in the data. Recently, [] propose a flexible ensemble algorithm for fair decision-making in the more challenging context of evolving online settings. This algorithm, called FARF (Fair and Adaptive Random Forests), is based on using online component classifiers and updating them according to the current distribution, which also accounts for fairness and a single hyper-parameters that alters fairness-accuracy balance. Baccarelli et al. [12] presents an Asynchronous Fair Adaptive Federated learning framework (AFAFed) for stream-oriented IoT application environments, which are featured by time-varying operating conditions, heterogeneous resource-limited devices (i.e., coworkers), non-i.i.d. local training data and unreliable communication links.

### 9.3.2 Fairness for Sequential Decision Making

In recent years, fairness has started having a strategic connotation [194]. [42, 74, 90] have shown that exists complex relations between model decision and effect on the underlying population. Therefore, many works aim at studying the impacts of imposing fairness strategies on underlying feature distributions. [117, 104, 89] proposed multi-stage models. In particular, [89] studied a model for long-term improvements in the labor market and proved that imposing the Demographic Parity constraint can lead to an equitable long term equilibrium. The authors in [117] show that consequential decisions can reshape population over time. They demonstrate that some fairness definitions do not promote improvement and may cause harm. They also introduced a one-step feedback model to quantify the long-term impact of classification on different groups in the population, by classifying them as *long-term improvement*, *stagnation*, or *decline*. [139, 94] focused in providing fairness in reinforcement learning to detect the policy that maximizes the cumulative rewards subject to certain fairness strategies; [82] constructs a user participation dynamics model where individuals respond to perceived decisions. The aim is to understand the impact of various fairness interventions on group representation. [118, 137, 173, 90] studied the long-term impacts of decisions on the group's qualification states with different dynamics. The authors in [137] investigate the role of dynamics of non-discrimination in the population, concluding that *"imposing fairness considerations on decision-making systems requires understanding the influence that they will have on the population at hand"*. Consistently, [173] agrees that *"algorithmic decisions lead to changes in the underlying feature distribution, which then feed back into the decision-making process."*

The main difference between *Fair Stream Learning* and *Fairness for Sequential Decision Making* is that in the first, drifts in the data are taken into account to build a stable fair-classifier, whereas in the second fairness strategies are assumed to cause drifts in the data. In our work, we address simultaneously these two fields of fairness. Understanding and formalizing dynamics between decisions and fairness remain a major challenge [54] as the works proposed in this direction have introduced several assumptions and simplifications. Moreover, in the credit context, the effects caused by the fairness strategy in the population could be overshadowed by many other factors, such as macroeconomic policies or immigration phenomena that cause drifts in the data.

In this work, instead of formalizing dynamic cause-effect relationships, we are interested in providing Demographic Parity through *blind adaption* and investigating methods based on XAI to observe the effects that mitigation policies actually cause between sensitive population subgroups over time.

This condition is achievable only if $X$ is independent of $S$, i.e. when there are no financial gaps between different classes of the protected attributes. Note that under this condition, *Counterfactual Fairness* [111] is also achieved, as there are no causal relationships triggered by sensitive attributes.

Assuming that it is not possible to achieve the optimal situation in the present, we are motivated to achieve it over time. Keeping the same notation introduced in [117], we propose extensions of some XAI methods to observe qualitatively *long-term improvement*, *stagnation*, or *decline* after each model retraining window. Our experiments aim to ensure a stable Demographic Parity over time - as the chosen fairness constraint - and to observe through XAI whether the level of individual discrimination, introduced by the group mitigation model, is decreasing. If this evidence occurs, it means that the population is having a *long-term improvement* and is moving towards the optimal situation.

Table 9.1 Demographic Parity of the various models tested on different temporal samples or after different stress tests. DP is calculated using *citizenship* as sensitive feature. Bold highlights $|DP|$ under 0.01 and underline highlights $|DP|$ over 0.1

| | Demographic Parity Evaluation | | | | | |
|---|---|---|---|---|---|---|
| Model Trained in 2018 | Real Data 2018 | Real Data 2019 | Pos. Shock Overall | Neg. Shock Overall | Pos. Shock Conditioned | Neg. Shock Conditioned |
| Unmitigated RF | <u>-0.21</u> | <u>-0.26</u> | <u>-0.15</u> | <u>-0.19</u> | -0.09 | <u>-0.44</u> |
| Massaging + RF | **0.003** | <u>-0.12</u> | 0.057 | -0.071 | <u>0.14</u> | <u>-0.16</u> |
| AdversarialDP | **0.005** | <u>-0.14</u> | 0.081 | -0.093 | <u>0.17</u> | <u>-0.2</u> |
| RF + TreshDP | **0.004** | <u>-0.12</u> | 0.061 | -0.073 | <u>0.16</u> | <u>-0.18</u> |

## 9.4 Experiment and Empirical Evidence

We replicate some models implemented in `BeFair`, as presented in Section 5.6, in order to achieve DP through *pre-processing*, *in-processing* and *post-processing*. Specifically, the employed models are *Massaging*, *Aversarial Debiasing*, and *TreshDP*. As in the previous chapter, we train after (before) pre-processing (post-processing) mitigation a random forest model.

We apply mitigation strategies for the sensitive feature *citizenship* since it is correlated with the most important financial variables to determine the granting of the loan. We focus only on this sensitive variable given that the *gender* does not show any bias in the data.

### 9.4.1 CQ1: Evaluating Fairness Stability Through Time

According to the state-of-the-art, the evaluation of the quality of trained models is often performed on a set of validation data extracted from the same temporal sample used for training. This validation isn't reliable in the financial perimeter because the model has to run in a live environment, hence ensuring both performance and fairness over time. Our experiments revealed that current approaches fail in satisfying the DP only one year after training.

The second column of Table 9.1 shows a clear decay of all mitigation models of Demographic Parity when the model's forecasts are computed with the data of the following year compared to the year used to build the training set, confirming what is already stated.

**Stress Tests.** A set of stress tests are conducted to shed light on this phenomenon. The trained models are used to predict the granting of the loan after (i) injecting shocks (positive and negative) that cause drifts in the most important variables (*"overall shock"*) and after (ii) injecting shocks (positive and negative) that cause drifts in the most important variables conditioned to only one class of the protected attribute (*"conditioned shock"*).

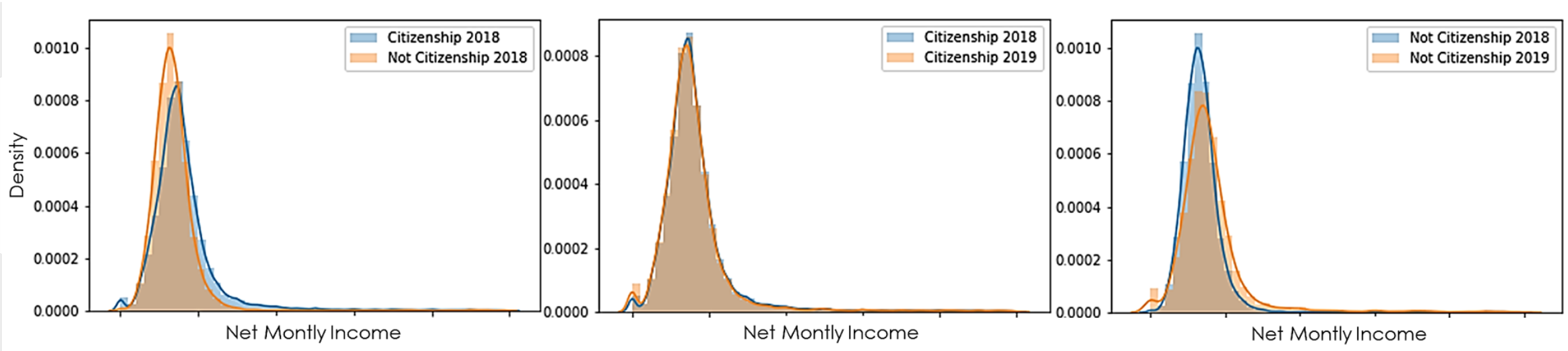

(a) Density plot: *Net Montly Income* of Citizenship vs non Citizenship in 2018

(b) Density plot: *Net Montly Income* of Citizenship in 2018 vs Citizenship in 2019

(c) Density plot: *Net Montly Income* of Not Citizenship in 2018 vs Not Citizenship in 2019

Figure 9.1 Density plot of the variable *net montly income* conditioned to vary combination of *citizenship* and *year*. Distribution values are blinded for data privacy.

Table 9.1 shows that the worsening of Demographic Parity occurs more evidently when *conditioned shocks* occur compared to drifts caused by *overall shocks*. This behaviour clarifies why the models failed to provide DP with real data as well. Indeed Figure 9.1 shows that in real data the *income* variable is affected by a drift for the *not citizen* (see Figure 9.1(c)) and not for the *citizen* (see Figure 9.1(b)). In particular, from Figure 9.1(c), it is possible to notice a drift to the right on the distribution, which is a good sign. Unfortunately, it also occurs an increase of the frequency of people with low income, which negatively affects the final group fairness results.

### 9.4.2 CQ2: Providing a Stable Fairness Through Time

As previously shown, the classical mitigation methods can fail in ensuring DP when deployed on a live environment. This usually happens as the marginal contributions assigned by the mitigation model to the protected attributes are no longer properly calibrated in relation to the conditional distribution drift of some important variables. Notice that, building a fair classifier that ensures *group* fairness - rather than *individual* - aims in improving the financial condition of the vulnerable class and it is consistent with the concerns about the trained *group-fair* classifier robustness over time.

A practical solution is to retrain $\psi_1$ using a dataset $D_{t_2}$ at time $t_2$ that contains the new financial conditions, resulting in a novel model $\psi_2$. This approach, which apparently may seem trivial, brings with it the complication of choosing correctly the target variable $Y_{t_2}$ to update the models. Indeed, in case the model is already deployed, the decision of the loan officer - needed to retrain the system - is no longer available. Furthermore, in the credit lending field, it is difficult to build a feedback system to reinforce the results of the model, as (i) it is impossible to observe whether the not granted loans would have been repaid (*false negative control*) and (ii) the defaults on the granted loans may not have occurred yet (*false positive control*).

Table 9.2 Fairness and Performance of the *Massaging* algorithm $\psi_2$ using different target label methods. Bold highlights $|DP|$ under 0.01 and area under the ROC Curve (AUC) over 0.8.

| Target Label Method | DP ($\psi_2$) | AUC ($\psi_2$) |
|---|---|---|
| Group Fair Model $\psi_1$ | **0.001** | 0.74 |
| Individual Fair Model $\phi_1$ | **0.003** | **0.83** |

We recall that our goal is to provide both an output (i) that satisfies DP and (ii) that leads to a convergence between individual and group fairness. To this end, our idea is to create an individual fair model, named $\phi_{t_1}$, using $D_{t_1}$ as input and the loan officer decision $Y_{t_1}$ as target. Since $\phi_{t_1}$ reproduces the behavior of the loan officer without introducing individual discrimination, its predictions on $D_{t_2}$ might be used to update the group mitigation model $\psi_2$.

To clarify the matter, Table 9.2 compares the results of the *Massaging* model $\psi_2$ retrained (i) using the prediction of $\phi_1$ on $D_{t_2}$ as the target (*individual fair model as target label method*) and (ii) using the prediction of $\psi_1$ on $D_{t_2}$ as the target (*group fair model as target label method*).

By observing the results, clearly, both models respect DP by design, but the model retrained with an individual target has superior performance. Performance is measured by knowing the actual decision of the loan officer.

### 9.4.3 CQ3: Monitoring Fairness Effects Through Time

Even though an AI algorithm might embed ethical principles by design to ensure social good, there is no guarantee that these will eventually occur during deployment over time. Moreover, it is important to monitor the consequences that mitigation strategies have on people subjected to model decisions. Unwanted distortions or complex effects could impact people in unexpected ways [167]. It is therefore essential to design monitoring systems to verify the effects of fairness strategies and eventually change them based on the acquired *know-how*.

The process described in the previous chapter makes monitoring fairness possible, explaining the changes in the model behaviour after each model retraining window [127]. Deploying a classifier that provides DP, when there are previous direct or indirect relationships between the sensitive attribute and the target variable, means assigning a positive marginal contribution to the disadvantaged class. The greater the marginal contribution, the greater the individual discrimination introduced by the model to ensure equality between groups. SHAP [121] allows to observe the marginal contributions assigned by the mitigation model to the classes of the sensitive attribute through the *Shapley values*. We propose to use an extension of

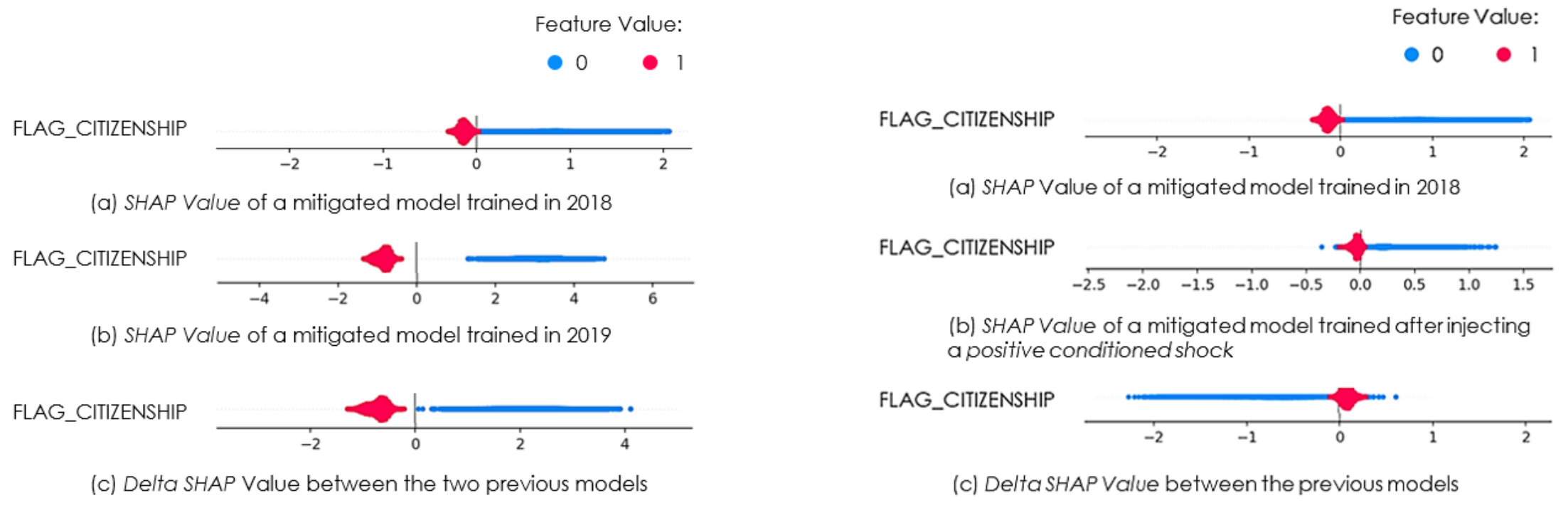

Figure 9.2 (a - left) Representation of the Shapley values of two mitigated models trained in different years and the relative differences. (b - right ) Representation of the Shapley values of two mitigated models trained in 2018, after injecting a positive conditioned shock and the relative differences.

SHAP – `FairX` – to observe the $\Delta$*Shapley values* calculated from the group mitigation models $\psi_1$ and $\psi_2$, trained in $D_{t_1}$ and $D_{t_2}$ respectively. If the marginal contributions assigned to the sensitive attributes decrease after retraining, it represents a *long-term improvement* because the updated model is now able to ensure *group* fairness by decreasing *individual* discrimination.

Figure 9.2 (a - left) shows the Shapley values of two *Massaging* models trained respectively in (a) 2018 and (b) 2019 and (c) their differences. It can be seen that the marginal contributions, assigned to the protected class of the sensitive attribute, increase in 2019 compared to 2018. Both models ensure Demographic Parity in their year of competence, but to do so, in 2019 individual discrimination must be increased. This synthetic result clearly indicates a *decline*; the financial gaps have clearly widened without the necessity to observe the conditional distribution drifts of all the variables present in the data.

This *decline* highlighted by this result is due to macroeconomic phenomena that have affected the population. It is not possible to verify the effects of the fairness policy proposed in this paper as it was not really used to grant the loans. To provide an example of *long-term improvement* we have replicated the presented explanation techniques after injecting a *positive conditioned shock* (see Figure 9.2 (b - right)).

## 9.5 Discussion on Addressing Fairness Through Time

In this chapter, we conducted experiments on real data owned by Intesa Sanpaolo of about 800,000 personal loan granting documents dated between 2016 and 2019. The results of the experiments showed that (i) in the financial context, classical systems for ensuring DP can fail when used over time, raising the issue of instability in fairness when deployed to production. This enforces the need of monitoring fairness as stated in the sixth step of the roadmap presented in Chapter 7. We also (ii) have addressed the topic of how to carry out the retraining of the AI system assuming the continuous unavailability of the ground truth. We investigated a strategy that encodes human behaviour while retraining the model over time, favoring the convergence to the optimal situation and providing superior performance. Finally, (iii) we introduced `FairX`, a methodology that exploits Shapley values to monitor the trend of fairness. Its purpose is to highlight changes in the level of individual discrimination after the retraining of the group mitigation model. Thanks to this explanation it is possible to verify qualitatively if a chosen mitigation strategy is contributing to inducing *long-term improvement*, *stagnation*, or a *decline* in the population.

Regarding our motivating example, it's fundamental to acknowledge that the scenario presented is a strong simplification of the complex reality. Granting a loan, especially in the presence of adverse life biases, may not suffice on its own to significantly improve the financial status of a minority. In certain cases, it could potentially result in adverse outcomes by exacerbating default rates within that minority, leading to harm. Just as with any decision-making context, whether driven by AI or human judgment, the impact of a binary "yes" or "no" decision might not bring about substantial positive change without the support of more comprehensive structural social initiatives.

# Part IV

# Conclusions

This thesis represents a comprehensive exploration of bias and fairness in the context of the banking sector. The journey through the intricate landscape of fairness in artificial intelligence has revealed its vast and intricate nature, particularly when the consequences of AI decisions impact people's lives. It is evident that "fairness in AI" is a multifaceted and sensitive topic, demanding nuanced approaches.

Throughout this thesis, our mission has been to pave the way for Responsible AI decision-making, recognizing the critical importance of addressing bias. To achieve this goal, we navigated the complexities of bias with real-world applications within Intesa Sanpaolo. Our efforts have revolved around the fundamental pillars of *understanding*, *mitigating*, and *accounting for* bias, all of which are crucial for fostering fairness in AI, aligning with ethical values, and forthcoming regulations.

In the realm of understanding bias, we introduced `Bias On Demand`, a model framework capable of generating synthetic data that vividly illustrates specific bias types. We also contributed to the field by shedding light on the intricate landscape of fairness metrics, enhancing our comprehension of their interplay and nuances.

Moving to the mitigation of bias, we presented `BeFair`, an adaptable toolkit designed for the practical implementation of fairness in real-world scenarios. Additionally, we introduced `FFTree`, a fairness-aware decision tree that enhances transparency and flexibility in decision-making systems.

In the pursuit of accounting for bias, we laid out a roadmap for addressing fairness within the banking sector, emphasizing the need for interdisciplinary collaboration to comprehensively address context and societal impact. We introduced `FairView`, a novel tool that supports users in selecting ethical frameworks for addressing fairness, bridging the gap between Explainable AI and moral philosophies. Furthermore, our investigation into the dynamic nature of fairness through time led to the development of `FairX`, an XAI strategy capable of monitoring the effects of fairness over time.

By applying these contributions in real-world settings in collaboration with Intesa Sanpaolo, we have not only deepened our understanding of fairness but have also provided practical tools for the responsible implementation of AI-based decision-making systems. Upholding the principles of open-source development, we have made `Bias On Demand` and `FairView` available as Python packages.

As the field of AI continues its rapid evolution and integration into diverse sectors, the imperative of responsible AI, underpinned by fairness, transparency, and human oversight, becomes increasingly evident. This thesis serves as a step towards a future where AI-based decision-making systems are not only technically proficient but ethically robust, earning the trust of society and advancing the promise of responsible AI.